\documentclass[10pt, journal]{IEEEtran}
\usepackage{bm}
\usepackage{booktabs}
\usepackage{amsmath}
\usepackage{subfigure}
\usepackage{amssymb}
\usepackage{float}
\usepackage{caption}
\usepackage{subcaption}
\usepackage{diagbox}
\usepackage{adjustbox}
\usepackage{makecell}
\usepackage{mathrsfs}
\usepackage{multirow}
\usepackage{xcolor}
\usepackage{balance}
\usepackage{graphicx}
\usepackage{animate}
\usepackage[citecolor=blue, colorlinks]{hyperref}
\usepackage[T1]{fontenc}
\usepackage{soul}

\begin{document}

\title{Dynamic 3D Point Cloud Sequences as 2D Videos}
	
\author{Yiming Zeng, Junhui Hou, \textit{Senior Member, IEEE}, Qijian Zhang, Siyu Ren, and Wenping Wang, \textit{ Fellow, IEEE}
\thanks{This work was supported by the Hong Kong Research Grants Council under Grant 11202320, Grant 11219422, and Grant 11218121. \textit{Corresponding Author: Junhui Hou}}
\thanks{Y. Zeng, J. Hou, Q. Zhang, and S. Ren are with the Department of Computer Science, City University of Hong Kong, Hong Kong SAR. Email:jh.hou@cityu.edu.hk}
\thanks{W. Wang is with the Department of Computer Science \& Engineering, Texas A \& M University, USA. Email: wenping@tamu.edu}
}

\markboth{Manuscript submitted to IEEE TPAMI}
{}
\maketitle

\begin{abstract}
    Dynamic 3D point cloud sequences serve as one of the most common and practical representation modalities of dynamic real-world environments. However, their unstructured nature in both spatial and temporal domains poses significant challenges to effective and efficient processing. Existing deep point cloud sequence modeling approaches imitate the mature 2D video learning mechanisms by developing complex spatio-temporal point neighbor grouping and feature aggregation schemes, often resulting in methods lacking effectiveness, efficiency, and expressive power. In this paper, we propose a novel generic representation called \textit{Structured Point Cloud Videos} (SPCVs). Intuitively, by leveraging the fact that 3D geometric shapes are essentially 2D manifolds, SPCV re-organizes a point cloud sequence as a 2D video with spatial smoothness and temporal consistency, where the pixel values correspond to the 3D coordinates of points. The structured nature of our SPCV representation allows for the seamless adaptation of well-established 2D image/video techniques, enabling efficient and effective processing and analysis of 3D point cloud sequences. To achieve such re-organization, we design a self-supervised learning pipeline that is geometrically regularized and driven by self-reconstructive and deformation field learning objectives. Additionally, we construct SPCV-based frameworks for both low-level and high-level 3D point cloud sequence processing and analysis tasks, including action recognition, temporal interpolation, and compression. Extensive experiments demonstrate the versatility and superiority of the proposed SPCV, which has the potential to offer new possibilities for deep learning on unstructured 3D point cloud sequences. Code will be released at \textit{\url{https://github.com/ZENGYIMING-EAMON/SPCV}}.
\end{abstract}
	
		
\section{Introduction} \label{S_Intro}

\IEEEPARstart{A}{} dynamic 3D point cloud sequence comprises multiple frames of static 3D point clouds captured at consecutive time steps, providing a depiction of geometric changes in objects/scenes. This type of data finds extensive applications in areas such as autonomous driving, robotics navigation, virtual/augmented reality, and immersive telecommunication. As the demand for 3D data processing continues to grow, fueled by the remarkable success of deep learning in handling 2D images/videos, there is an urgent need to develop effective and efficient learning methods for processing dynamic 3D point cloud sequences.

\begin{figure}[t]
    \hsize=1\linewidth
    \centering
    \animategraphics[width=0.95\linewidth, loop, autoplay]{12}{demo_contour/gif_01/Len_}{0001}{0100}
    \caption{Visualization of structuring of a single 3D point cloud frame. Each line indicates the corresponding 2D pixel and 3D point that are displayed with an identical color. The black points correspond to the pixels located at the 2D image's boundaries. Please use \textbf{Adobe Acrobat} to display the video.}
    \label{fig_demo_contour}
    \vspace{-0.6cm}
\end{figure}

\begin{figure*}[t]
    \centering
    \includegraphics[width=0.9\textwidth]{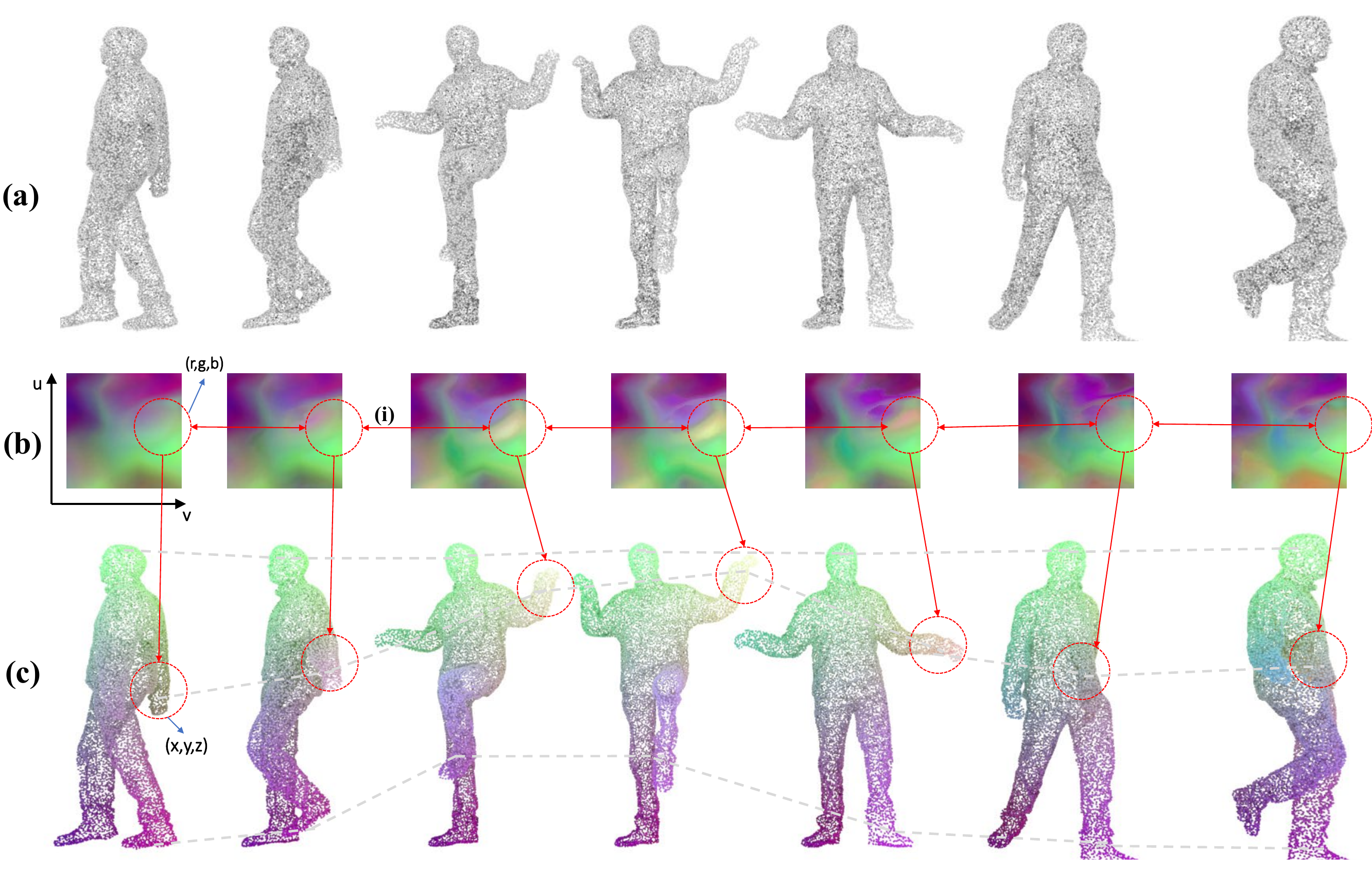}
    \caption{Illustration of the proposed \textbf{SPCV} representations for dynamic 3D point cloud sequences. (a) A dynamic 3D point cloud sequence with point-wise correspondences \textit{unknown}. (b) The SPCV representation of the above sequence. (c) The dynamic 3D point cloud sequence reshaped/reprojected from the above SPCV by taking the value of a pixel as the coordinate of a 3D point. Note that for each frame, the points are rendered in the same colors as the corresponding pixels to illustrate the correspondence between 2D and 3D domains.}
    \label{fig:concept_demo}
    \vspace{-0.3cm}
\end{figure*}

However, deep modeling of 3D point clouds, characterized by irregularity and lack of structure, is not as straightforward as applying standard 2D convolutional operators \cite{dosovitskiy2015flownet,he2016deep,ilg2017flownet,simonyan2014two} on well-structured 2D image/video signals defined on regular 2D grids. In recent years, researchers have devoted significant efforts to exploring numerous deep set architectures \cite{qi2017pointnet, qi2017pointnet++,wang2019dynamic,zhao2021point,qian2022pointnext,lin2023meta} that operate on \textit{static} 3D point cloud inputs, by developing various complicated learning mechanisms and highly specialized feature extraction operators. the processing and modeling challenges become even more pronounced when transitioning from a \textit{static} point cloud to a \textit{dynamic} sequence of consecutive point cloud frames. This is due to the irregular spatial structure of each frame and the lack of temporal correspondence across frames. Existing dynamic 3D point cloud sequence learning networks \cite{liu2019meteornet,pami_22_psttransformer_fanhehe} typically rely on extensive spatio-temporal point neighbor grouping and multi-scale feature aggregation operations, which are less expressive, memory consuming, and computationally inefficient,  particularly with an increasing number of input points. The substantial computational complexity of these time-consuming operations hinders the development of deeper or more sophisticated architectural models for 3D point cloud sequences, thereby limiting performance on downstream tasks. In addition to the issues of network structures, in various reconstruction/generation scenarios, the inefficiency of commonly-used loss functions, e.g., Chamfer Distance (CD) and Earth Mover’s Distance (EMD) \cite{fan2017point}, can also significantly degrade the overall learning effects. This stands in stark contrast to the 2D image/video domain, where pixel-wise errors can be directly computed at the same 2D grid coordinates.
	
In this paper, we address the aforementioned challenges from a novel and more fundamental perspective by structuring 3D point cloud sequences. By recognizing that 3D geometric shapes are essentially 2D manifolds, we propose to structure dynamic 3D point cloud sequences with a 2D video-like representation, namely structured point cloud video (SPCV). As depicted in Fig. \ref{fig:concept_demo}, the proposed SPCV representation exhibits the following key features: (\textit{i}) the pixel values of an SPCV correspond to the coordinates of 3D points; (\textit{ii}) the adjacent pixels within a frame generally represent neighboring points in 3D space; (\textit{iii}) the pixels with identical locations across different frames of an SPCV correspond to the consistent position of the object/scene. To achieve such a structured representation, we design a geometrically regularized self-supervised learning pipeline comprising two stages: frame-wise structurization and sequence-wise structurization. These stages are driven by the learning objectives of intra-frame self-reconstruction and inter-frame deformation fields, respectively.  Furthermore,  for a better understanding of the structurization process, we direct readers to the video demonstrations presented in Fig. \ref{fig_demo_contour} and Fig. \ref{fig_demo_contour_series}. 

\begin{figure*}[t]
    \hsize=1\textwidth 
    \centering    \animategraphics[width=\textwidth, loop, autoplay]{12}{temporal_gif/}{0001}{0100}
    \caption{Visualization of our structurization of eight frames of a point cloud sequence named \textit{Crane} from a typical viewpoint. Note that pixels of all frames are projected to 3D space following the same order to illustrate the temporal consistency across different frames. Please use \textbf{Adobe Acrobat} to view the video.}
    \label{fig_demo_contour_series}
    \vspace{-0.4cm}
\end{figure*}

Essentially, the process of structuring a dynamic 3D point cloud sequence to its SPCV representation involves explicitly learning the indexing cues of the geometric coordinates, resulting in the re-organization of original 3D spatial points into regular 2D grids. The uniform structure of SPCV relives the challenge faced by neural networks in feature learning, thereby potentially generating more expressive features. The decoupled manner can significantly enhance the efficiency of downstream applications, allowing for the integration of advanced techniques to improve performance. The SPCV representation possesses spatial smoothness and temporal consistency properties, enabling the direct application of 2D image/video processing techniques for seamless processing of dynamic 3D point cloud sequences.  By leveraging well-established 2D image/video techniques, we further construct SPCV-based frameworks for a range of high-level and low-level 3D point cloud sequence processing and analysis tasks, including action recognition, temporal interpolation, and geometry compression. Experimental results demonstrate the versatility and superiority of our approaches.
	
In summary, the main contributions of this paper are:
	
\begin{enumerate}
    \item We propose a new generic representation modality for dynamic 3D point cloud sequences, offering numerous benefits for point cloud sequence processing and analysis. 
    \item We propose a novel self-supervised learning framework that efficiently and effectively represents any point cloud sequences as SPCVs.
    \item We construct diverse SPCV-based frameworks for dynamic 3D point sequence analysis and processing, achieving state-of-the-art performances. 
\end{enumerate}

The remainder of this paper is organized as follows: Section \ref{S_R} reviews related works in the field. Section \ref{sec_proposed_method} details the methodology behind our SPCV representation. Section \ref{SPCV_applications} provides a detailed introduction to the SPCV-based learning network employed for various downstream tasks. Section \ref{S_Exp} discusses the experimental setup and benchmarks our approach against existing methods. Finally, Section \ref{S_Con} concludes the paper and suggests potential directions for future research.

\section{Related Work} \label{S_R}
	
In this section, we begin with a comprehensive summarization of representative deep set architectures for learning from static 3D point clouds. Then, we focus on the more challenging problem of learning from dynamic 3D point cloud sequences, which is directly related to the scope of our work.
	
\subsection{Deep Learning on Static 3D Point Clouds}\label{rw:dl-static-pc}
	
The recent years have witnessed the proliferation of various deep set architectures that directly operate on unstructured 3D point sets without any pre-processing procedures, as pioneered by \cite{qi2017pointnet,qi2017pointnet++}. The follow-up researches further investigated rich varieties of convolution-based \cite{li2018so,li2018pointcnn,thomas2019kpconv,liu2019relation,yan2020pointasnl,xu2021paconv,xiang2021walk,qian2022pointnext,lin2023meta}, graph-based \cite{verma2018feastnet,wang2019dynamic,xu2020grid}, and transformer-based \cite{guo2021pct,zhao2021point,park2022fast,yu2022point} backbone learning architectures.
	
More specifically, PointNet~\cite{qi2017pointnet} proposed to perform point-wise embedding and global aggregation via shared multi-layer perceptrons (MLPs) and channel max-pooling. PointNet++~\cite{qi2017pointnet++} further incorporates multi-scale hierarchical feature abstraction mechanisms with farthest point sampling (FPS) and neighborhood interpolation. SO-Net~\cite{li2018so} explicitly utilizes the spatial distribution of point clouds by building self-organizing maps. PointCNN~\cite{li2018pointcnn} applies learned transformations for adaptively weighting and permuting point features in potentially canonical orders. FeaStNet~\cite{verma2018feastnet} dynamically builds correspondences between filter weights and graph nodes. KPConv~\cite{thomas2019kpconv} learns to continuously locate convolution weights in Euclidean space through kernel points and further explores spatially deformable convolution operators. RS-CNN~\cite{liu2019relation} focuses on capturing the high-level relations among local point sets via learning from predefined geometric priors. DGCNN~\cite{wang2019dynamic} queries neighbors globally in the feature space for constructing graph edges and performing dynamic feature aggregation. PointASNL~\cite{yan2020pointasnl} adopts an adaptive sampling scheme together with local-nonlocal cells to improve its model robustness when dealing with noises and outliers. CurveNet~\cite{xiang2021walk} aggregates hypothetical curves initially grouped through guided walks to augment point-wise features. PointMLP~\cite{ma2022rethinking} builds a pure residual MLP network equipped with lightweight geometric affine modules, which can perform competitively without any sophisticated learning mechanisms. PointNeXt~\cite{qian2022pointnext} systematically revisits the classic PointNet++ processing pipeline and then improves model training and scaling strategies. PT~\cite{zhao2021point} investigates highly-expressive self-attention layers for point clouds. FastPT~\cite{park2022fast} incorporates lightweight local self-attention with voxel hashing architectures to encode continuous 3D positional information to greatly enhance computational efficiency. Conclusively, these approaches above are specialized for learning from a single 3D point cloud input to extract geometric feature representations in the spatial domain. The adaptation to dynamic point cloud learning for achieving sequence processing and analysis is highly non-trivial, due to the complexity of joint spatio-temporal modeling.
	
	Besides, another line of work resorts to a different perspective for overcoming the irregularity and unstructuredness of 3D point clouds by creating regular 2D geometry image (GI \cite{gu2002geometry}) or GI-like representation structures, on which numerous mature 2D modeling architectures (e.g., convolutional neural networks (CNNs)) can be seamlessly applied, or adapted with minimal modifications, to achieve various point cloud learning tasks. Representatively, \cite{sinha2016deep,sinha2017surfnet,maron2017convolutional,haim2019surface} exploit traditional mesh parameterization techniques to implement the process of surface-to-plane mapping, while \cite{zhang2022reggeonet,zhang2023flattening} propose to directly learn deep 2D regular representations from 3D point clouds in an unsupervised manner. 
	These works are also related to the previous folding-based methods~\cite{yang2018foldingnet,groueix2018papier,TearingNet}.
	However, these approaches are still limited to static geometric representations, and are thus inapplicable to dynamic point cloud sequences with requirements of spatio-temporal structurization.	
	
\subsection{Deep Learning on 3D Point Cloud Sequences} \label{rw:dl-dynamic-pc}
	
DPMix~\cite{zhang2023dpmix} employs 2D depth video methods to enhance the understanding of point cloud videos.
Other methods explore the temporal interpolation of the 3D point cloud sequences~\cite{lu2020pointinet, zeng2022idea, zheng2023neuralpci}. These advancements demonstrate the growing interest and diversification in approaches to effectively model the dynamic nature of 3D point cloud sequences.
	
Specifically, PSTNet~\cite{fan2021pstnet} introduces the point spatio-temporal (PST) convolution, leveraging a `point tube' concept to aggregate local spatio-temporal features from 3D point cloud sequences. Building upon this, PSTNet2~\cite{pami_21_pstnet2_fanhehe} merges spatial features into a unified spatio-temporal representation. And MaST-Pre~\cite{shen2023masked} advances this concept with a masked point tube for pre-training on point cloud videos, incorporating the spatiotemporal masked autoencoder idea from \cite{feichtenhofer2022masked}. P4Transformer~\cite{cvpr_21_p4d_fanhehe} boosts the performance of PSTNet by employing transformers to bypass the need for point tracking while capturing spatio-temporal correlations across point cloud sequences. Similarly, $\text{PST}^2$~\cite{wei2022spatial} introduces a spatio-temporal self-attention (STSA) module to grasp contextual information across adjacent frames. This concept is further enhanced by PST-Transformer~\cite{pami_22_psttransformer_fanhehe} with spatio-temporal encoding. These architectures primarily follow an autoencoder-like framework to process dynamic point cloud sequences, akin to the role of PointNet++~\cite{qi2017pointnet++} in handling static point clouds. While these methods introduce specially designed spatio-temporal feature aggregation structures to adapt to dynamic point cloud learning challenges, they do not fundamentally address the inherent irregularities of spatio-temporal characteristics in 3D point cloud sequences. In essence, the irregular data modality of 3D point cloud sequences remains unaltered.
	
Another line of research explores implicit representation methods. OFlow~\cite{niemeyer2019occupancy} adopts implicit strategies for 4D reconstruction, utilizing ground truth occupancy and trajectory information. CaSPR~\cite{rempe2020caspr} introduces spatio-temporal representations for objects within normalized coordinate space. Similarly, CaDeX~\cite{Lei2022CaDeX} tackles inter-frame deformations by employing continuous bijective canonical maps, focusing on a canonical shape. Notably, these methodologies primarily utilize mesh, voxel, or occupancy data, diverging from pure point cloud sequences. Our work, in contrast, concentrates on self-supervised explicit representations. Specifically, we structure the irregular 3D point cloud sequences as structured point cloud videos, aiming to achieve a compact representation that effectively minimizes both distortion and storage requirements.
	
A variety of methods have been developed for specific 3D point cloud sequence processing tasks, as well as related technologies. These technologies include flow-based~\cite{wei2021pv,li2021neural,he2022learning}, depth-based~\cite{wang20203dv}, GAN-based~\cite{li2021tpu}, kinematics-inspired~\cite{zhong2022no}, contrastive learning~\cite{sheng2023point}, and correspondence approaches~\cite{eisenberger2021neuromorph, zeng2021corrnet3d}. There are also some task-specific sequence processing methods, notably compression~\cite{quach2020folding,graziosi2020overview} and interpolation~\cite{lu2020pointinet, zeng2022idea, zheng2023neuralpci}.
	
For example, PV-RAFT~\cite{wei2021pv} proposes a point-voxel recurrent all-pairs field transforms method to estimate scene flow from point clouds. NSFP~\cite{li2021neural} revisits the scene flow problem, emphasizing runtime optimization and regularization of the scene flow. SSFE~\cite{he2022learning} aims to predict 3D scene flow for all pairs of point clouds in a given sequence. 3DV~\cite{wang20203dv} utilizes the dynamic 3D voxel for depth-based 3D action recognition. TPU-GAN~\cite{li2021tpu} proposes a super-resolution generative adversarial network for upsampling dynamic point cloud sequences. Kinet~\cite{zhong2022no} explores kinematics-inspired neural networks to capture 3D motions without explicit tracking. These methods employed various approaches, such as utilizing voxel and depth information for feature extraction, utilizing GAN-based methods to avoid the need for explicit point correspondence annotation, or relying on explicit scene flow annotation datasets. Overall, these methods did not directly address the underlying irregularity issues of 3D point cloud data. Furthermore, they lacked powerful insights into the explicit geometric aspects of 3D point clouds.

\section{Proposed Method} \label{sec_proposed_method}
	
\subsection{Problem Statement} \label{sec_prob_state}
	
As mentioned earlier, learning from irregular and unordered point sets presents a significant challenge compared to feature extraction from 2D images or videos organized in a regular grid. Essentially, an RGB image of sizes $H\times W\times 3$ can be regarded as a set of 5-dimensional vectors denoted as $(r, g, b, h, w)$, where the color information (or pixel value) $(r, g, b)$ to be processed is indexed by the 2D coordinate $(h, w)$ uniformly distributed on a regular 2D grid of sizes $H\times W$, while for point cloud data, the spatial coordinates $(x,y,z)$ serve as both the geometry information to be processed and the indirect indexing cues to be used for determining inter-point relationships such as proximity and correspondence. Consequently, existing point-based deep learning architectures often fall short in efficiency when processing large-scale point data due to their involvement in complex learning mechanisms, including cumbersome pre-processing, extensive neighbor querying, and costly sampling, matching, and multi-scale abstraction processes. The significant computational complexity of the aforementioned time-consuming operations in turn hinders the development of deeper or more sophisticated architectural models for 3D point cloud sequences, thus limiting performance on downstream tasks. Additionally, these operations increase memory consumption, particularly when processing denser 3D point cloud sequences. Furthermore, measuring the discrepancy between two point clouds poses a significant challenge due to their irregular and unordered nature, making standard methods like CD and EMD either ineffective or inefficient \cite{nguyen2021point}. 

The shift from static point cloud to 3D point cloud sequence modeling introduces additional challenges, exacerbated by the inherent lack of structure in both spatial and temporal dimensions. Current approaches rely heavily on intensive spatial neighborhood searches and the establishment of temporal correspondences, necessitating complex spatio-temporal feature processing. This intricate methodology is not just laborious and costly—it also substantially obstructs the creation of streamlined and efficient learning frameworks for 3D point cloud sequences. 

\subsection{Our Objective} \label{sec_objective}

Denote by $\mathcal{P}=\{\mathbf{P}_t \in \mathbb{R}^{N_t\times 3}\}_{t=0}^{T-1}$ a dynamic 3D point cloud sequence with $T$ point cloud frames, where $N_t$ refers to the number of points contained in the $t$-th frame\footnote{Without losing generality, we assign a fixed number of points to all frames in practice.}. Note that the point-wise correspondence across the frames of $\mathcal{P}$ is \textit{unknown}. Inspired by the above-mentioned structure of 2D images/videos and the fact that 3D geometric shapes are essentially 2D manifolds, we aim to structure dynamic 3D point cloud sequences to address the aforementioned challenges.

Generally, for each point of a typical point cloud frame $\mathbf{P}_t$, we plan to obtain a unique 2D coordinate $(u,v)$, thereby forming a 5-dimensional signal $(x, y, z, u, v)$. Meanwhile, points that are nearest neighbors in 3D space should have close 2D coordinates. When the $(u, v)$ coordinates are uniformly distributed across the regular 2D grid, the structure effectively resembles that of an RGB image. In other words, we can fill $[x,~y,~z]$ into the corresponding $(u, v)$ location, constructing  an RGB image-like representation of dimensions $U\times V\times 3$ ($U\times V=N_t$). After processing all point cloud frames of $\mathcal{P}$ with this manner, we can derive a 2D video-like representation of $\mathcal{P}$, denoted as $\mathcal{G}=\{\mathbf{G}_t \in \mathbb{R}^{U\times V\times 3}\}_{t=0}^{T-1}$, named \textit{Structured Point Cloud Video} (SPCV).  Moreover, we constrain that the encoded points at an identical location across different frames of $\mathcal{G}$ correspond to the same or approximately the same position of the 3D object/scene, which is more advanced than traditional RGB videos. In the following, we also call an element of $\mathbf{G}_t$ a pixel for convenience. Note that $\mathbf{G}_t$ can be directly reshaped/reprojected back to a point cloud without any additional computation, i.e., the value of a pixel is taken as a 3D point.

\begin{figure}[t]
    \centering
    \includegraphics[width=0.5\textwidth]{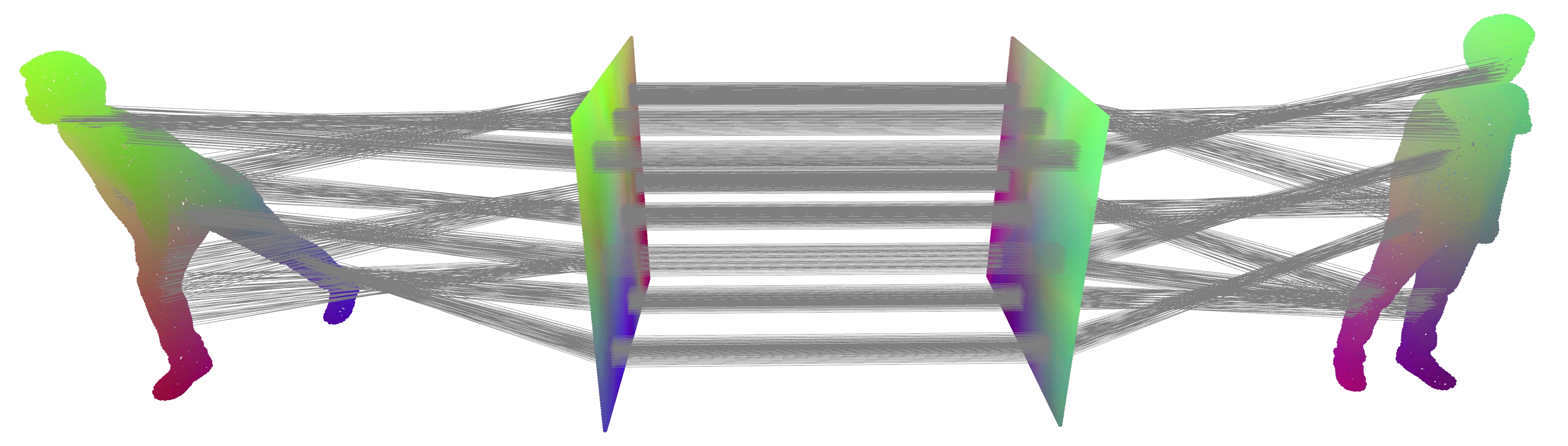}
    \caption{Illustration of the local spatial smoothness and temporal consistency properties of our proposed representation, named SPCV. Here, we take two frames of a dynamic point cloud sequence as an example.}
    \vspace{-0.6cm}
    \label{fig_local_smth}
\end{figure}

In summary, we anticipate that the resulting SPCV should exhibit the following two characteristics, as illustrated in Fig.~\ref{fig_local_smth}:

\begin{enumerate}
    \item \textit{Spatial Smoothness}: the pixels in a local patch of a typical frame of an SPCV correspond to a set of neighboring points in 3D space; 
    \item \textit{Temporal Consistency}: the pixels with identical locations across the $T$ frames of an SPCV correspond to the consistent position of the object/scene.
\end{enumerate}

The proposed representation modality is able to offer the following advantages:

\begin{enumerate}
    \item \textit{Improved Efficiency and Performance of Downstream Applications.} 
    The SPCV framework streamlines downstream processing and analysis tasks by replacing complex and time-consuming operations like FPS and K-NN aggregation with simpler and highly efficient pixel indexing and sliding window mechanisms. Meanwhile, the uniform structure of SPCV relives the challenge faced by neural networks in feature learning, thereby potentially generating more expressive features. Moreover, SPCV can seamlessly integrate with established learning techniques used in 2D image/video network designs, enhancing overall performance. 
    \item \textit{Reduction of Memory Consumption.}  SPCV efficiently reduces memory usage by indexing frames using variable pixel locations, a contrast to traditional methods that necessitate storing multiple scales of point clouds for feature aggregation. This advantage becomes particularly evident when processing dense or large-scale point clouds.
    \item \textit{Simplified Design and Implementation.} Adopting SPCV simplifies the design and implementation complexities in point cloud applications. For instance, encoders with specialized spatio-temporal feature extraction can be replaced with encoders that are already commonly used in the 2D image/video community.
    \item \textit{Free of the Limitations of CD and EMD.}  Traditional models often struggle with inefficiencies from point cloud loss functions like CD and Earth Mover's Distance (EMD). The SPCV framework, however, enables more efficient and straightforward optimization using grid-based losses like the Frobenius norm or \(\mathcal{L}_1\), thanks to its inherent structure.
\end{enumerate}
	
\begin{figure*}
    \centering
    \includegraphics[width=0.95\textwidth]{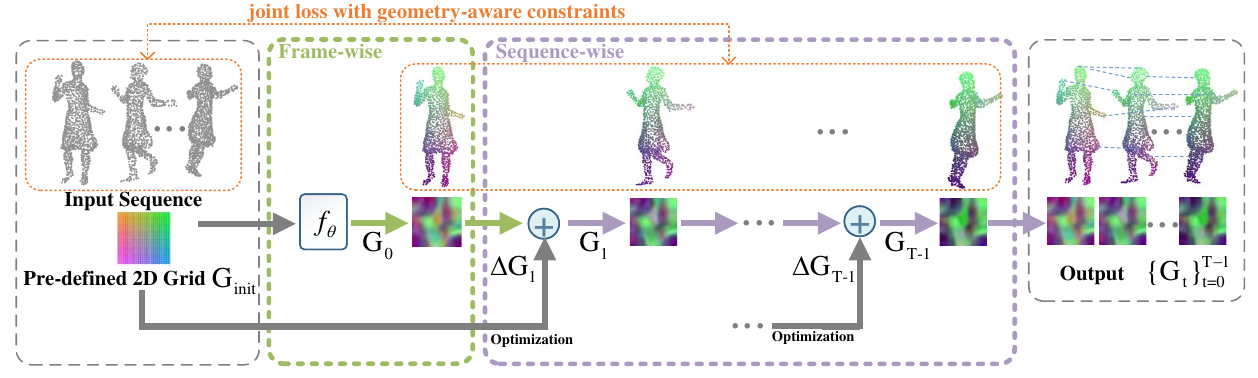}
    \caption{Flowchart of the SPCV representation for 3D point cloud sequences.}
    \vspace{-0.5cm}
    \label{fig:SPCV_twostages}
\end{figure*}

\subsection{Our Pipeline} \label{sec_tech_overview}  
 
\noindent \textbf{Technical Intuition}. Intuitively, to achieve the objective of re-organizing $\mathcal{P}$ into $\mathcal{G}$ through a learning-based method, we can construct a neural network, which takes $\{(x,y,z)\}$ as input and outputs $\{(u,v)\}$, and train it by using $(x,y,z)$ and $(u,v)$ paired training data. However, adopting a supervised learning approach presents significant challenges. Even for static point clouds exhibiting arbitrary and non-manifold geometric structures, acquiring the ground-truth $(u,v)$ poses a formidable task due to the complex parameterization problem \cite{sheffer2007mesh}. This challenge becomes even more pronounced when considering dynamic point cloud sequences, as obtaining additional accurate point-wise correspondences across frames as ground-truth remains a persistent and challenging problem.  

To overcome this challenge, we consider constructing a \textit{self-supervised} pipeline without relying on ground-truth 2D locations and temporal correspondences. Specifically, for a single point cloud, we first pre-define a set of 2D locations uniformly distributed on a regular 2D grid of sizes $U\times V$, each location storing/encoding $(u, v, 0)$. We then feed $\{(u, v, 0)\}$ into a learnable neural network to generate $\{(\hat{x},\hat{y},\hat{z})\}$ further assigned to the corresponding locations of their inputs. We can utilize the original point cloud as supervision to drive the learning of the network, ensuring that the generated $\{(\hat{x},\hat{y},\hat{z})\}$ closely resemble $\{(x,y,z)\}$. Moreover, owing to the inherent spectral bias property of neural networks \cite{rahaman2019spectral}, the generated data from spatially smooth inputs are expected to preserve spatial smoothness.

Furthermore, observing that a dynamic 3D point cloud sequence recording the motion of an object can be essentially thought of as a set of 3D points with a fixed topology deforming/evolving over time, we can adopt the following procedure to structure it. First, we represent $\mathbf{P}_0$ through the above-mentioned self-supervised learning approach for single point clouds, producing $\mathbf{G}_0$. We then obtain the representations of the remaining $T-1$ point cloud frames $\{\mathbf{P}_t\}_{t=1}^{T-1}$ by learning the deformation fields that deform $\mathbf{G}_{t-1}$ to $\mathbf{G}_{t}$ ($t\in[1,T-1]$) in a recursive fashion, i.e., $\mathbf{G}_t=\mathbf{G}_{t-1}+\bm{\Delta}_{\mathbf{G}_t}$, where $\bm{\Delta}_{\mathbf{G}_t}\in\mathbb{R}^{U\times V\times 3}$ stands for the deformation field for $\mathbf{G}_t$. To be specific, we can minimize the discrepancy between $\mathbf{P}_t$ and $\mathbf{G}_t$\footnote{Note that we reshape $\mathbf{G}_t$ into a point cloud and compute the discrepancy in 3D space.} to estimate $\bm{\Delta}_{\mathbf{G}_t}$. Such a procedure aligns the resulting structures of $\{\mathbf{G}_t\}_{t=1}^{T-1}$ with that of $\mathbf{G}_0$, leading to the spatial smoothness of all frames. Also, the deformation mechanism establishes a fixed topology across all frames, thus naturally preserving temporal consistency across all frames. 

\vspace{1em}
\noindent \textbf{Technical Details}. Based on the above analyses, we construct the self-supervised learning pipeline shown in Fig.~\ref{fig:SPCV_twostages} for representing $\mathcal{P}$ into $\mathcal{G}$. It consists of two main stages: \textbf{(\textbf{1})} frame-wise structurization and \textbf{(\textbf{2})} sequence-wise structurization, which are detailed as follows.
	
\subsubsection{Frame-wise Structurization} \label{frame_wise_stage}  
	
This stage only overfits the first point cloud frame of $\mathcal{P}$ (i.e., $\mathbf{P}_0$) into the first frame of $\mathcal{G}$ (i.e., $\mathbf{G}_0$) through a \textit{self-supervised} neural network regularized with additional geometrically meaningful constraints. Specifically, we first construct a pre-defined 2D grid $\mathbf{G}_{\rm init} \in \mathbb{R}^{U\times V\times 3}$, with each pixel filled with $(u, v, 0)$ where $u$ and $v$ is sampled regularly from $[0,U-1]$ and $[0,V-1]$, respectively. We then feed $\mathbf{G}_{\rm init}$ into a network composed of 2D convolutional (Conv2D) layers, denoted as $f_{\bm{\theta}}(\cdot)$ with $\bm{\theta}$ being network parameters. Owing to the inherent spectral bias property of neural networks \cite{rahaman2019spectral}, the generated image $\mathbf{G}_0 = f_{\bm{\theta}}(\mathbf{G}_{\rm init}) \in \mathbb{R}^{U\times V\times 3}$ is expected to be spatially smooth. Let $\widehat{\mathbf{P}}_0$ be the point cloud form reprojected from $\mathbf{G}_0$. We minimize the discrepancy between $\widehat{\mathbf{P}}_0$ and  ${\mathbf{P}}_0$ to enforce the shape represented by $\widehat{\mathbf{P}}_0$ to approximate $\mathbf{P}_0$. Thus, we drive the learning of $f_{\bm{\theta}}(\cdot)$ by optimizing the following loss function: 
\begin{equation}
    \min_{\bm{\theta}}\texttt{D}(\widehat{\mathbf{P}}_0, \mathbf{P}_{0}) + \texttt{R}_{\rm geo}(\mathbf{G}_0),
\end{equation}
where $\texttt{D}(\cdot, \cdot)$ stands for a typical distance metric for 3D point clouds \cite{ren2023unleash}, and $\texttt{R}_{\rm geo}(\cdot)$ is a geometry-aware regularization term further promoting the spatial smoothness characteristic of $\mathbf{G}_0$. Specifically, we explicitly regularize the generated pixel values of $\mathbf{G}_0$ to promote its spatial smoothness, i.e., the value of a typical pixel of $\mathbf{G}_0$ should be close to the average of those of its neighboring pixels. In addition, based on the fact that the normal of a typical point in a locally smooth region of a 3D surface is generally close to the average of the normals of its neighboring points,  we also regularize the geometry attribute of the generated pixels of $\mathbf{G}_0$, i.e., the normal of a typical pixel of $\mathbf{G}_0$ should be close to the average of those of its neighboring pixels. As $f_{\bm{\theta}}(\cdot)$ essentially models the process of 2D-to-3D mapping, we can compute the normal of a pixel located at $(u,v)$ through the cross product of $\mathbf{G}_0$'s partial derivatives with respect to $(u,~v)$:
\begin{equation}
    \begin{aligned}
    \mathbf{V}_{0}^{(u,v)}&= \frac{\partial\mathbf{G}_0^{(u,v)}}{\partial u}\times \frac{\partial\mathbf{G}_0^{(u,v)}}{\partial v},
    \end{aligned}
\end{equation}
where the partial derivatives can be derived through the back-propagation of $f_{\bm{\theta}}(\cdot)$. In other words, such regularization regularizes the network behavior. In all, we explicitly written $\texttt{R}_{\rm geo}(\mathbf{G}_0)$ as 
\begin{equation}
    \begin{aligned}
    \texttt{R}_{\rm geo}(\mathbf{G}_0)=&\sum_{(u,v)}\Big[\lambda_{\rm s}\Big(\mathbf{G}_{0}^{(u,v)}-\frac{1}{L^2}\sum_{(u',v')\in\mathcal{N}_{L}(u,v)}\mathbf{G}_{0}^{(u',v')}\Big)^2 \Big. \\
    &\Big.+\lambda_{\rm n} \Big(\mathbf{V}_0^{(u,v)}-\frac{1}{L^2}\sum_{(u',~v')\in\mathcal{N}_L(u,~v)}\mathbf{V}_{0}^{(u',~v')}\Big)^2\Big],
    \end{aligned}
    \label{equ:frame_reg}
\end{equation}
where $\lambda_{\rm s}$ and $\lambda_{\rm n}$ are two weights balancing the spatial and normal regularization terms, and $\mathcal{N}_{L}(\cdot,\cdot)$ represents the $L\times L$ window around the center pixel.

\begin{figure*}
    \centering
    \includegraphics[width=0.9\textwidth]{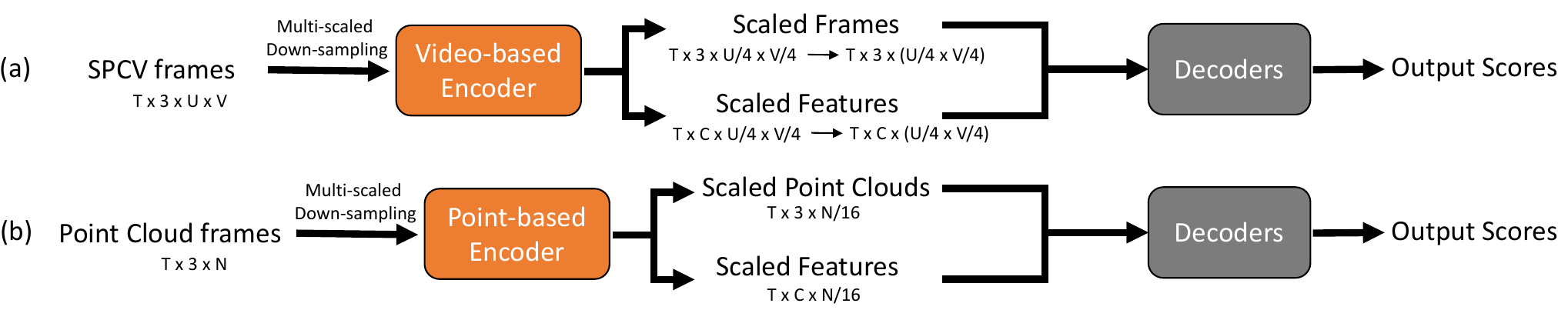}
    \caption{Flowchart of the action recognition model based on (a) SPCV and (b) PST-Transformer~\cite{pami_22_psttransformer_fanhehe}. (a) and (b) adopt decoders with the same architecture.}
    \vspace{-0.5cm}
    \label{fig:SPCV_act_recog_model}
\end{figure*}

\subsubsection{Sequence-wise Structurization} \label{sequence_wise_stage}  
	
This stage represents the remaining $(T-1)$ point cloud frames of $\mathcal{P}$ (i.e., $\{\mathbf{P}_t\}_{t=1}^{T-1}$) as the frames of $\mathcal{G}$  (i.e., \{$\mathbf{G}_t$\}$_{t=1}^{T-1}$) through an optimization scheme, which predicts the deformation field between two adjacent frames to deform $\mathbf{G}_{t-1}$ to $\mathbf{G}_{t}$ recursively, based on $\mathbf{G}_0$, i.e., $\mathbf{G}_t = \mathbf{G}_{t-1} + \bm{\Delta}_{\mathbf{G}_t}$ ($t\in[1,~T-1]$), where $\bm{\Delta}_{\mathbf{G}_t}\in\mathbb{R}^{U\times V\times 3}$ stands for the deformation fields at the $t$-th step. By establishing inter-frame relationships within $\mathcal{G}$ through deformation fields, we can achieve temporal consistency effectively. To be specific, we formulate the following optimization problem To obtain $\bm{\Delta}_{\mathbf{G}_t}$:
\begin{equation} \label{sec_stage_eq}
    \min_{\bm{\Delta}_{\mathbf{P}_t}} \texttt{D}(\widehat{\mathbf{P}}_{t-1}+\bm{\Delta}_{\mathbf{P}_t},\mathbf{P}_t)+\lambda\texttt{R}_{\rm smooth}(\bm{\Delta}_{\mathbf{P}_t}),
\end{equation}
where $\widehat{\mathbf{P}}_{t-1}\in\mathbb{R}^{UV\times 3}$ and $\bm{\Delta}_{\mathbf{P}_t}\in\mathbb{R}^{UV\times 3}$ are the point cloud forms, respectively reshaped from $\mathbf{G}_{t-1}$ and $\bm{\Delta}_{\mathbf{G}_t}$,  $\lambda>0$ balances the two terms, and $\texttt{R}_{\rm smooth}(\cdot)$ regularizes $\bm{\Delta}_{\mathbf{P}_t}$ (or $\bm{\Delta}_{\mathbf{G}_t}$) to be locally smooth/constant, based on the fact that neighboring points on $\widehat{\mathbf{P}}_{t-1}$ exhibit similar deformation. Such regularization also propagates the spatial smoothness of $\mathbf{G}_{t-1}$ to $\mathbf{G}_t$, defined as
\begin{equation}
    \texttt{R}_{\rm smooth}(\bm{\Delta}_{\mathbf{P}_t})=\frac{1}{N}\sum_{i=1}^{N}\frac{\sum_{j\in\mathcal{N}_K(i)}w(i,j)\|\bm{\Delta}_{\mathbf{P}_t}^{(i)}-\bm{\Delta}_{\mathbf{P}_t}^{(j)}\|^2}{\sum_{j\in\mathcal{N}_K(i)}w(i,j)},
    \label{equ:seq_reg}
\end{equation}
where $\mathcal{N}_K(\cdot)$ returns the index of $K$-NN points, and $w(i,j)=1/\|\mathbf{P}_{t-1}^{(i)}-\mathbf{P}_{t-1}^{(j)}\|^2$. We solve Eq. \eqref{sec_stage_eq} using the Adam optimizer \cite{kingma2014adam}, where $\bm{\Delta}_{\mathbf{P}t}$ is initialized with the values from the point cloud form of $\mathbf{G}_{\rm init}$.

It is also worth noting that as a trade-off between efficiency and effectiveness, we achieve the second stage with an optimization-based approach, and it is straightforward to substitute it with a self-supervised learning-based approach, where a neural network is trained to automatically learn and predict the deformation fields for each subsequent frame, for pursuing high performance.

\subsection{Discussion}

\begin{figure}[htbp]
    \centering
    \includegraphics[width=0.3\linewidth]{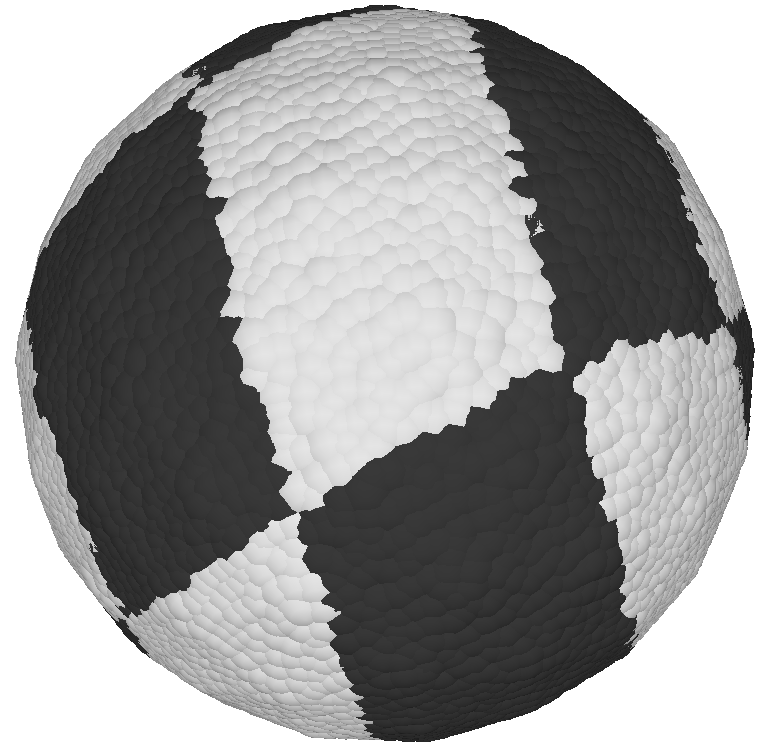}
    \hspace{1cm}
    \includegraphics[width=0.3\linewidth]{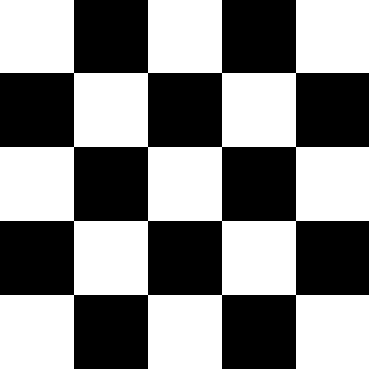}
    \caption{Example of representing a sphere point cloud (left) as an SPCV frame (right) with the checkerboard texture.}
    \label{fig:sphere}
\end{figure}

Generally, the proposed SPCV shares a similar concept with previous parameterization-based geometry image/video representation for 3D triangle meshes/dynamic mesh sequences \cite{gu2002geometry,briceno2003geometry,xia2010modeling,quynh2011modeling}, i.e., representing 3D geometry data with a 2D RGB image/video-like structure. However, our work significantly distinguishes itself from previous methods in terms of motivation, technical implementation, and application scenarios. First, previous methods map 3D vertices to the 2D domain through traditional parameterization algorithms optimizing the area and/or angle distortions of triangles. Due to the fundamental challenges of parameterization, these methods impose high requirements on the topological structures of 3D meshes to be processed, along with the requirement of correspondence across frames to model temporal information. Given that 3D point cloud sequences lack connectivity information and temporal correspondence and may have arbitrarily complex topological structures, these methods are largely inapplicable to our scenario. Second, driven by the objective of separating the indexing cues from the geometry information for improving efficiency, our pipeline re-organizes neighboring 3D points determined in the sense of the Euclidean distance into adjacent 2D locations on a regular 2D grid. The manner of determining a neighborhood is consistent with that of mainstream deep point-based architectures. In contrast, the parameterization-based methods essentially achieve the mapping in a geodesic distance-aware fashion. Accordingly, the theorems they employed are not suitable for our method. Finally, it is also worth noting that our method mimics the behavior of traditional parameterization when handling 3D objects with low curvatures, as demonstrated in Fig. \ref{fig:sphere}.

In summary, unlike the previous representation mechanism pursuing accurate parameterization, the proposed learning-based pipeline leverages the impressive capability of deep learning to offer a practical and promising solution for real-world applications. The advantages and effectiveness of our proposed SPCV are extensively demonstrated through experiments on various downstream tasks in Section \ref{SPCV_applications}.

\section{SPCV-based Applications} \label{SPCV_applications}

To verify the practical advantages of representing 3D point cloud sequences as SPCVs, we construct task-specific processing frameworks that directly consume SPCVs as inputs to achieve the corresponding dynamic point cloud applications. As aforementioned, the 2D video-like structure of our SPCV representation enables the adoption of various existing well-established 2D image/video deep learning techniques to construct task-specific frameworks.
	
Essentially, SPCV is designed to be a \textit{generic} representation structure, meaning that there are no restrictions on the downstream task scenarios. In principle, the downstream tasks should be sensitive to spatial smoothness and temporal consistency of input SPCVs, such that the resulting task performances can better indicate the representation quality of SPCVs. And the downstream tasks should cover both high-level semantic understanding and low-level geometry processing applications. Additionally, both learning-based and non-learning-based tasks are necessary to demonstrate the versatility of our SPCV representations. According to the above principles, we consider three targeted downstream tasks: 1) learning-based 3D action recognition; 2) learning-based temporal interpolation of point cloud sequences; and 3) non-learning-based 3D point cloud data compression. Specifically, action recognition evaluates the discriminative and expressive capabilities of high-level features learned through the SPCV representation. The effectiveness of temporal interpolation and compression serves as a measure of SPCV's representational quality in both spatial and temporal domains.

\begin{figure}
    \centering
    \includegraphics[width=0.5\textwidth]{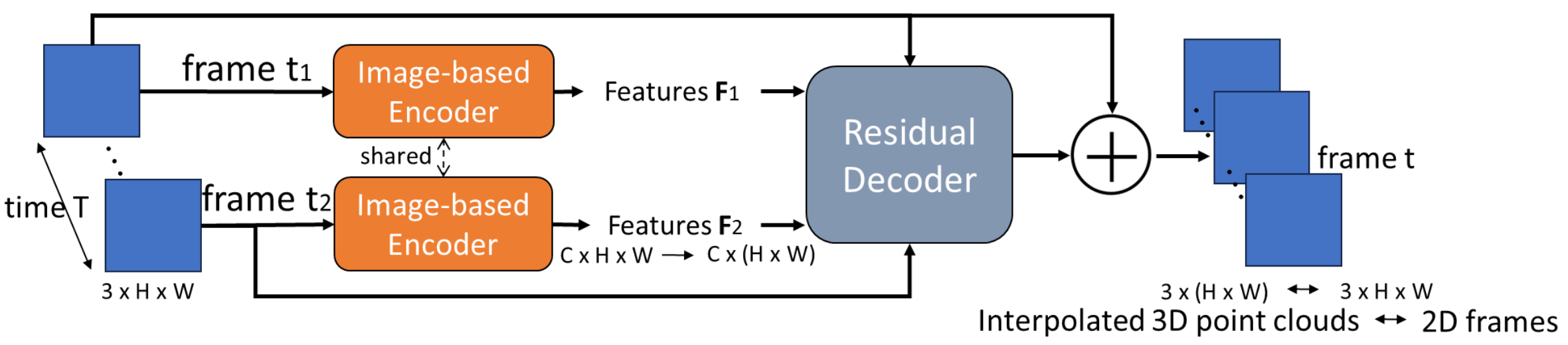}
    \caption{Flowchart of the proposed SPCV-based temporal interpolation for 3D point cloud sequences.}
    \vspace{-0.6cm}
    \label{fig:SPCV_intpl_model}
\end{figure}

\subsection{3D Action Recognition} \label{sec_app_act_rec}
	
As a common sequence-level task in point cloud sequence processing, this task evaluates the model's ability to extract discriminative features from each 3D point cloud sequence, thereby determining the most probable label for the sequence. Existing methods \cite{iclr_21_pstnet_fanhehe,pami_21_pstnet2_fanhehe,cvpr_21_p4d_fanhehe,pami_22_psttransformer_fanhehe} typically employ point-based spatio-temporal convolution operations, e.g., PST convolution \cite{iclr_21_pstnet_fanhehe,pami_21_pstnet2_fanhehe}, for learning feature encodings across spatio-temporal dimensions, or transformers \cite{cvpr_21_p4d_fanhehe,pami_22_psttransformer_fanhehe} for spatio-temporal context interaction. However, due to the insufficient design to cope with the unstructured characteristic of 3D point cloud data, their capability of extracting discriminative features efficiently and effectively is still relatively weak, thus limiting their accuracy. With the video-like representation, our approach enables the use of mature 2D convolutional operations to obtain more discriminative and expressive features. We modify the feature embedding process, common in approaches like PST-Transformer~\cite{pami_22_psttransformer_fanhehe}, by employing an efficient convolutional encoder. This modification efficiently generates more expressive feature tokens for the transformer.

Specifically, as illustrated in Fig. \ref{fig:SPCV_act_recog_model} (\textcolor{red}{a}), the proposed SPCV-based framework first spatially downsamples an input SPCV to a quarter of its original scale before being passed into a video-based encoder, consisting of an 18-layer `UNet3D' architecture, styled after ResNet \cite{he2016deep}, a design also used in conventional video-based methods \cite{tran2018closer,kalluri2021flavr}. In the network's bottleneck, the scaled SPCV frames, along with features scaled down to a quarter of their original size, are converted into sequences of point clouds and their corresponding features. These sequences are then fed into a decoder with transformer to enhance feature space interactions. Inside the decoder, the transformer's output features undergo two layers of max-pooling to produce batches of feature vectors. These vectors are subsequently processed through an MLP head layer to generate the final scores, indicating the probabilities of various classes for the input sequence.

\subsection{Temporal Interpolation of 3D Point Cloud Sequences}

This task involves reconstructing a high temporal resolution 3D point cloud sequence from a low-resolution one, a counterpart of the 2D video frame interpolation, which targets to increase the frame rate of a video by generating the in-between frames. The unstructured nature presented within individual frames and lack of correspondence across frames compound the complexity of this task, especially for sequences with large non-rigid deformation. PointINet~\cite{lu2020pointinet} addresses this task by adopting a scene flow estimator for interpolation. IDEA-Net~\cite{zeng2022idea} estimates point-wise trajectories for coarse interpolation and trajectory compensation. These methods suffer from inefficient and ineffective feature extraction and rely on distribution-based loss functions to interpolate point cloud frames with restricted fidelity. To address these issues, we develop an interpolation model based on an image encoder. Leveraging the structured nature of our representation, the mature image encoder efficiently generates more effective feature embedding tokens, which are then enhanced through downstream transformers in both temporal and spatial domains for feature interaction. And our model can be driven by common 2D loss functions like the Frobenius norm, enabling interpolating higher-quality point cloud frames efficiently.

Fig. \ref{fig:SPCV_intpl_model} illustrates the overall learning pipeline of our proposed SPCV-based point cloud frame interpolation. Specifically, for an input pair of consecutive frames at time step $t_1$ and $t_2$, we consider interpolating a new frame at time step $t$, where $t_1 < t < t_2$. To implement this, we feed the two point cloud frames into a shared 2D image-based encoder consisting of $4$ Conv2D layers to produce the corresponding feature maps $\mathbf{F}_1$ and $\mathbf{F}_2$. Then we can deduce an interpolated feature map, i.e., $\frac{t_2 - t_1 - t}{t_2 - t_1} \mathbf{F}_1 + \frac{t}{t_2 - t_1} \mathbf{F}_2$, which is further passed into a residual decoder with $5$ Conv1D layers. Finally, we add the generated residuals to the input frame at time step $t_1$ to obtain the desired interpolation result.

\subsection{3D Point Cloud Data Compression}

With the rapid development of 3D sensing technologies and the broad deployment of consumer-level devices (e.g., depth camera, LiDAR, etc.), there is an urgent need to develop efficient 3D point cloud codecs to compress the corresponding huge data volume and effectively save transmission bandwidth and storage space. However, unlike its mature 2D counterpart, i.e., image/video compression, which has been investigated and developed for a long time, 3D point cloud compression is still a relatively young topic whose actual coding performance is far from satisfactory. Therefore, there exists an obvious contradiction between ever-growing 3D point cloud data volume and immature compression techniques. Compared to 2D image/video compression, 3D point cloud data compression\footnote{Note that we consider compression of the geometry information of 3D point clouds, rather than the attributes.} poses a greater challenge due to the unstructured nature of spatial and temporal domains, i.e., the points of each frame are irregularly distributed, the number of points may change from one frame to another, and there is no consistent point-to-point correspondence between successive frames.

The Moving Picture Experts Group (MPEG) introduced two point cloud compression (PCC) standards: Geometry-Based PCC (G-PCC) and Video-Based PCC (V-PCC) for compressing static and dynamic point cloud data, respectively. We refer the readers to \cite{schwarz2018emerging,liu2019comprehensive,graziosi2020overview} for detailed introductions of these methods and \cite{quach2022survey} for a survey on deep learning-based point cloud compression.

Like \cite{hou2014compressing,hou2014highly}, the proposed SPCV naturally bridges the gap between dynamic 3D point cloud sequences and mature 2D image/video compression techniques, paving the way for efficient and effective compression pipelines. Specifically, given a 3D point cloud sequence, we represent it as an SPCV, which is further fed into the Versatile Video Coding (VVC) Test Model, the latest H.266/VVC video compression standard \cite{ITU2021H266}, to achieve compression. This pipeline is fully compatible with existing infrastructures for 2D image/video compression and transmission.

\begin{figure} 
    \centering
    \includegraphics[width=1.3cm]{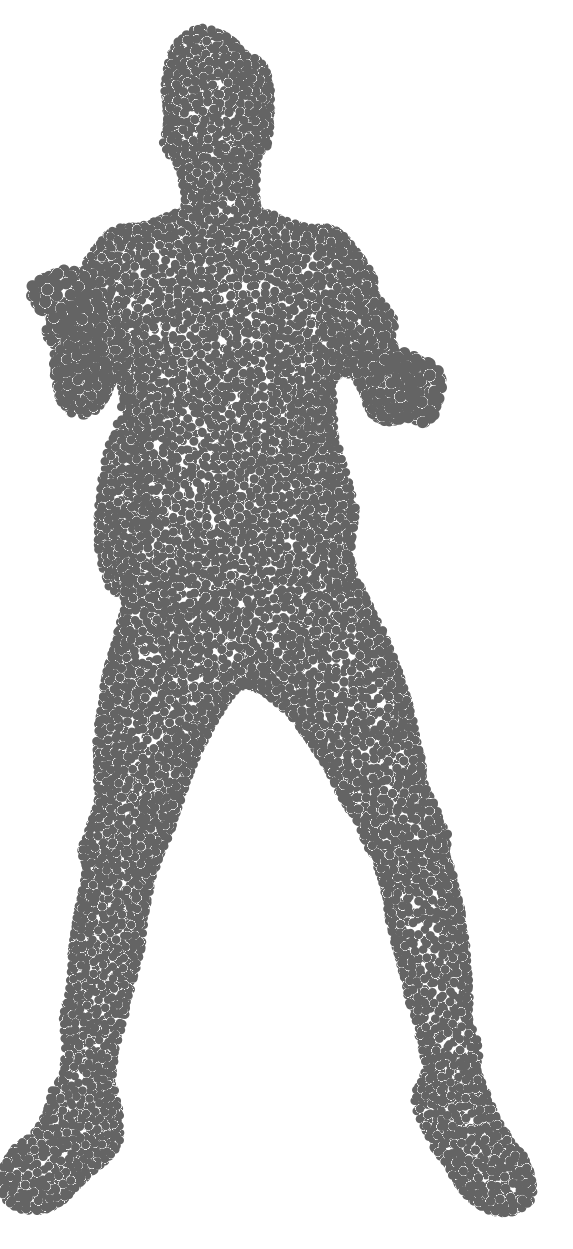}
    \includegraphics[width=1.3cm]{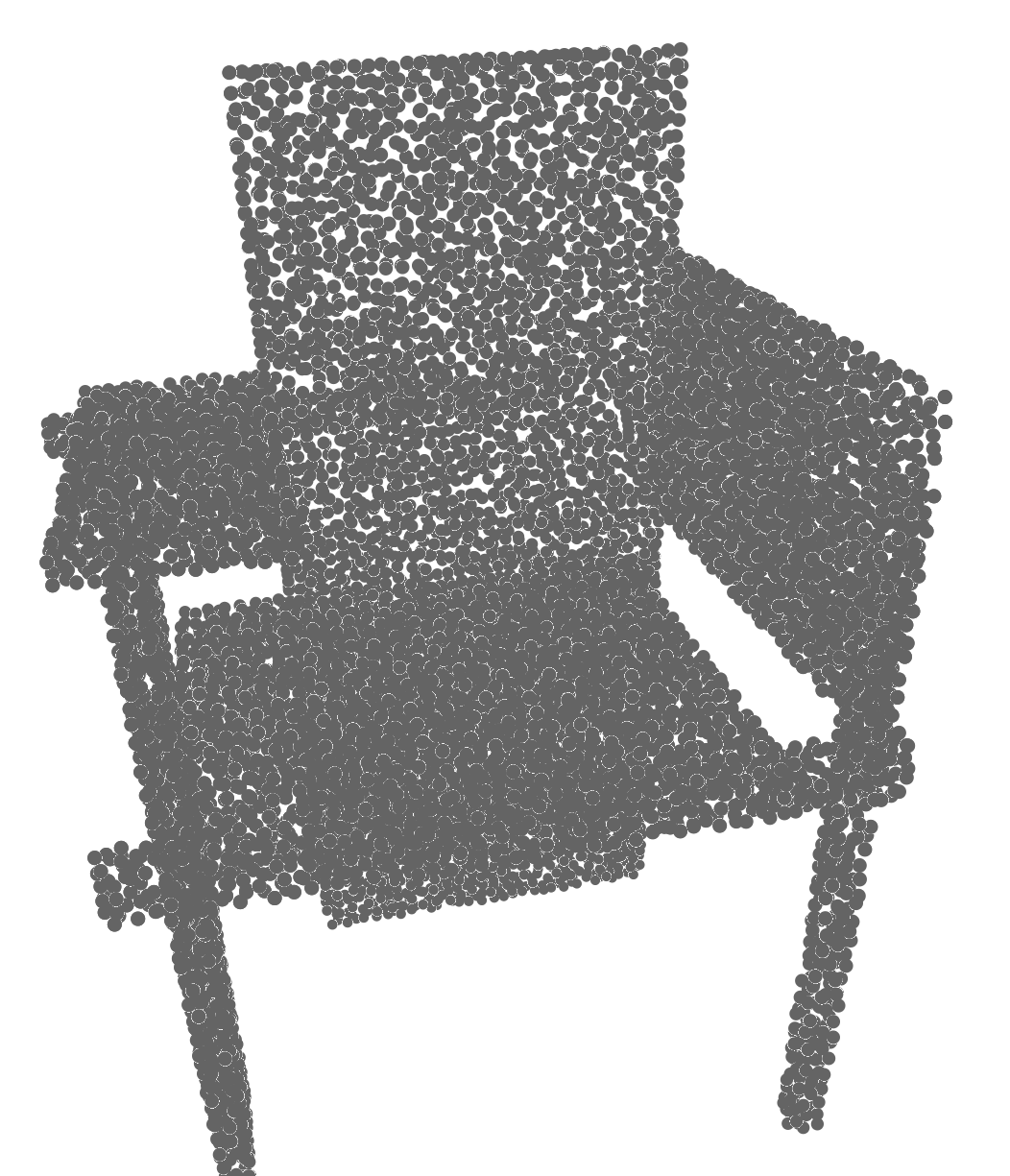}
    \includegraphics[width=1.3cm]{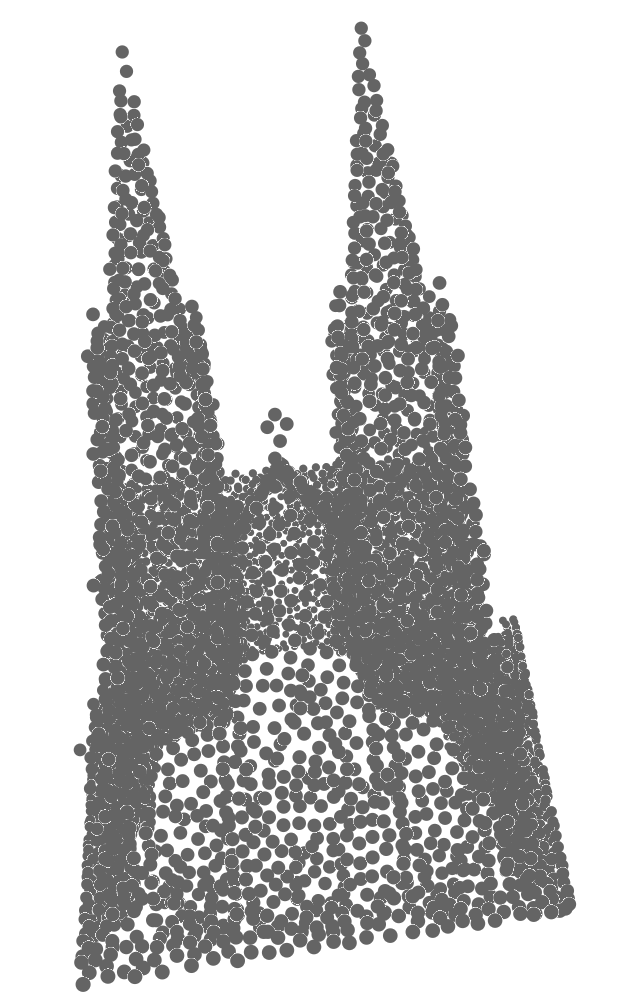}
    \includegraphics[width=1.3cm]{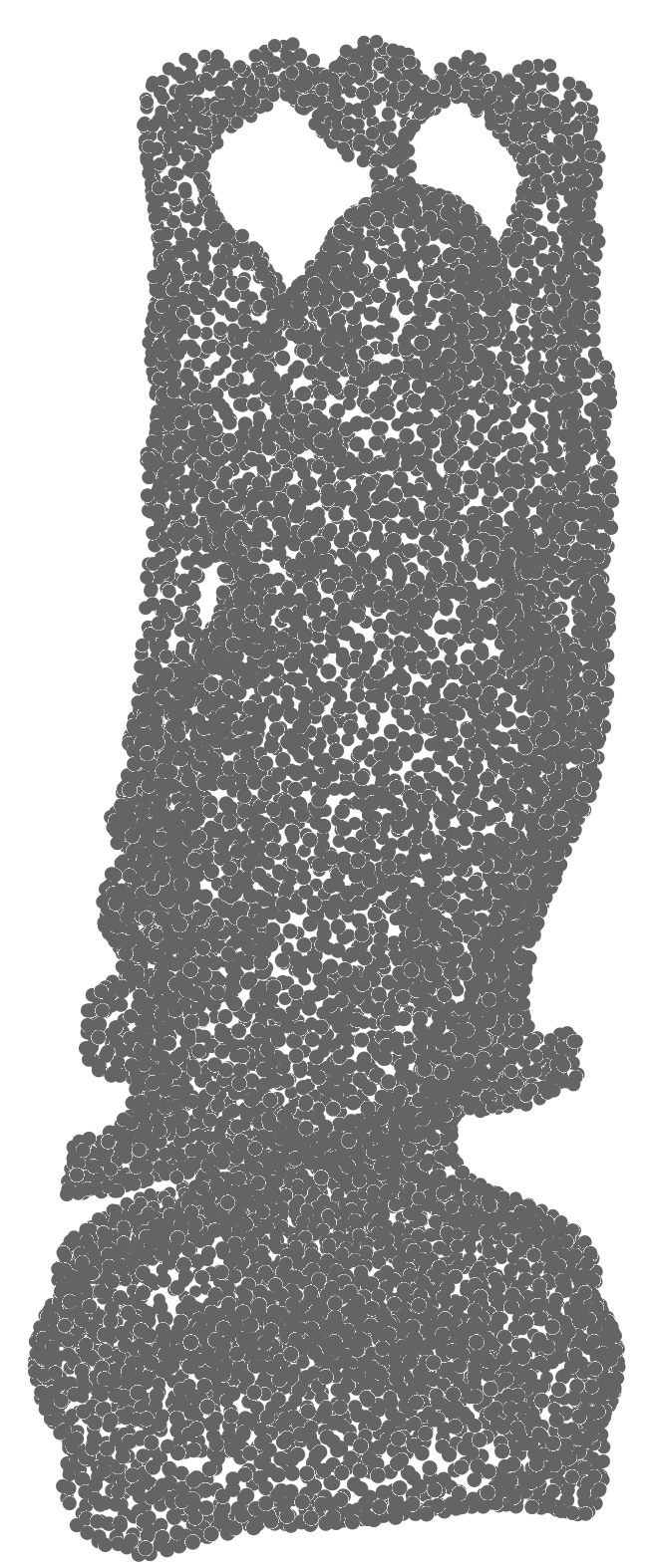}
    \includegraphics[width=1.3cm]{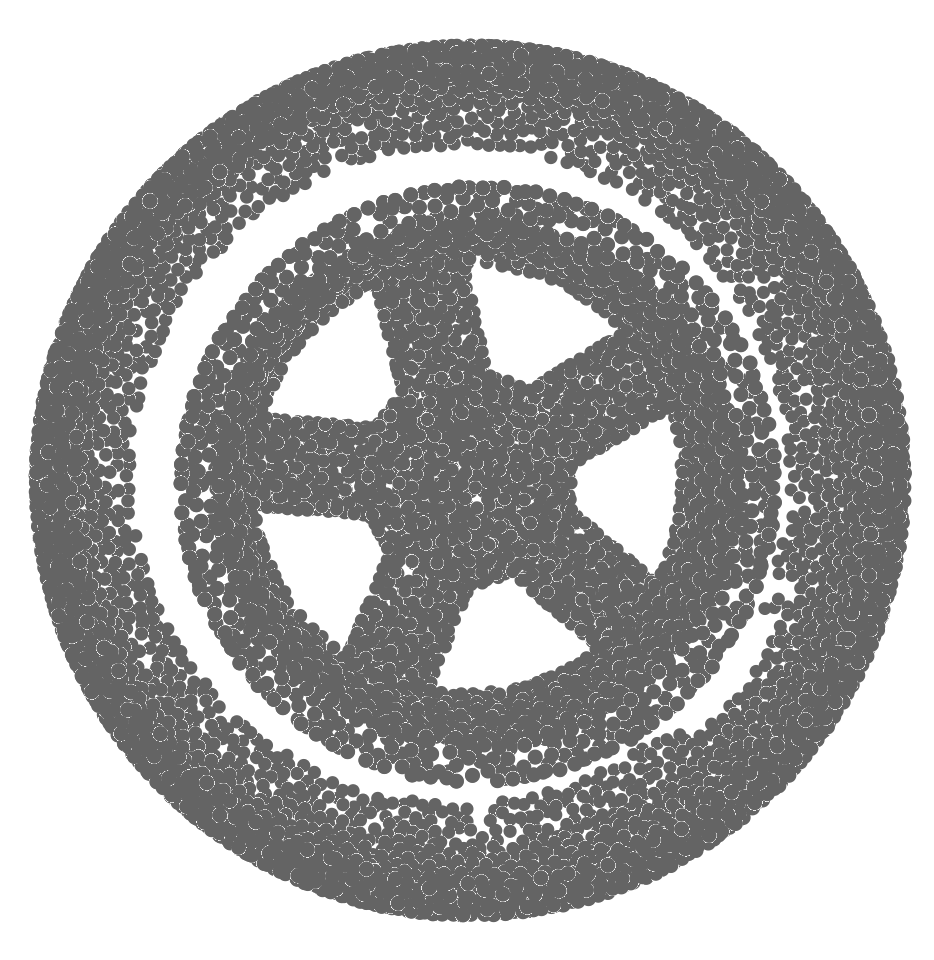}
    \includegraphics[width=1.3cm]{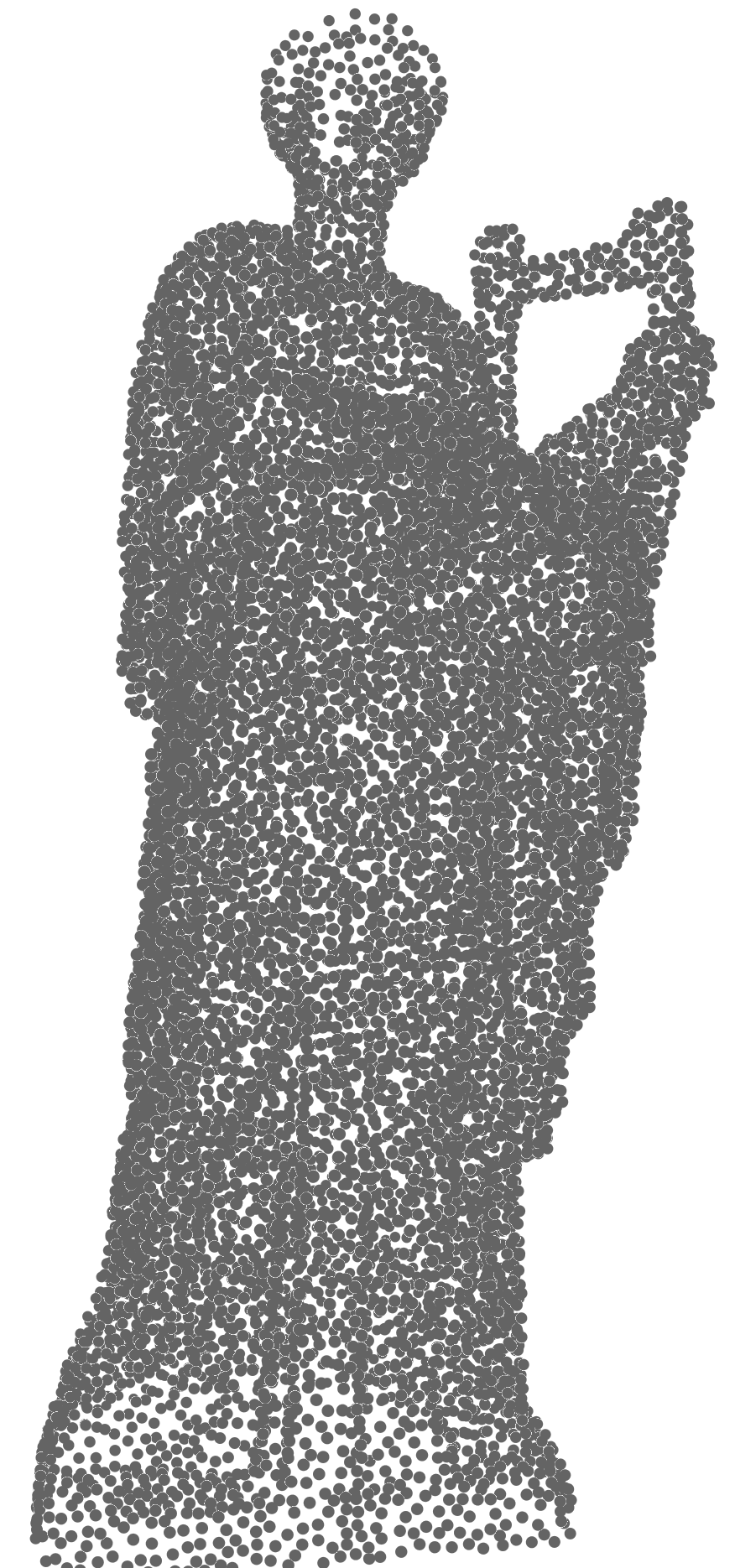}
    \vspace{0.1cm}
    
    \includegraphics[width=1.3cm]{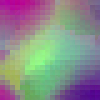}
    \includegraphics[width=1.3cm]{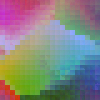}
    \includegraphics[width=1.3cm]{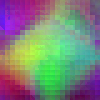}
    \includegraphics[width=1.3cm]{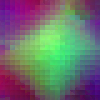}
    \includegraphics[width=1.3cm]{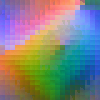}
    \includegraphics[width=1.3cm]{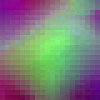}
    \vspace{0.1cm}
    \includegraphics[width=1.3cm]{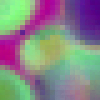}
    \includegraphics[width=1.3cm]{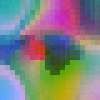}
    \includegraphics[width=1.3cm]{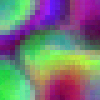}
    \includegraphics[width=1.3cm]{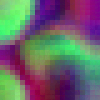}
    \includegraphics[width=1.3cm]{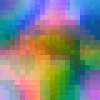}
    \includegraphics[width=1.3cm]{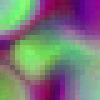}
    
    \includegraphics[width=1.3cm]{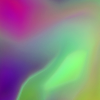}
    \includegraphics[width=1.3cm]{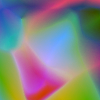}
    \includegraphics[width=1.3cm]{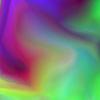}
    \includegraphics[width=1.3cm]{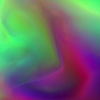}
    \includegraphics[width=1.3cm]{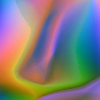}
    \includegraphics[width=1.3cm]{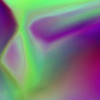}
    
    \includegraphics[width=1.3cm]{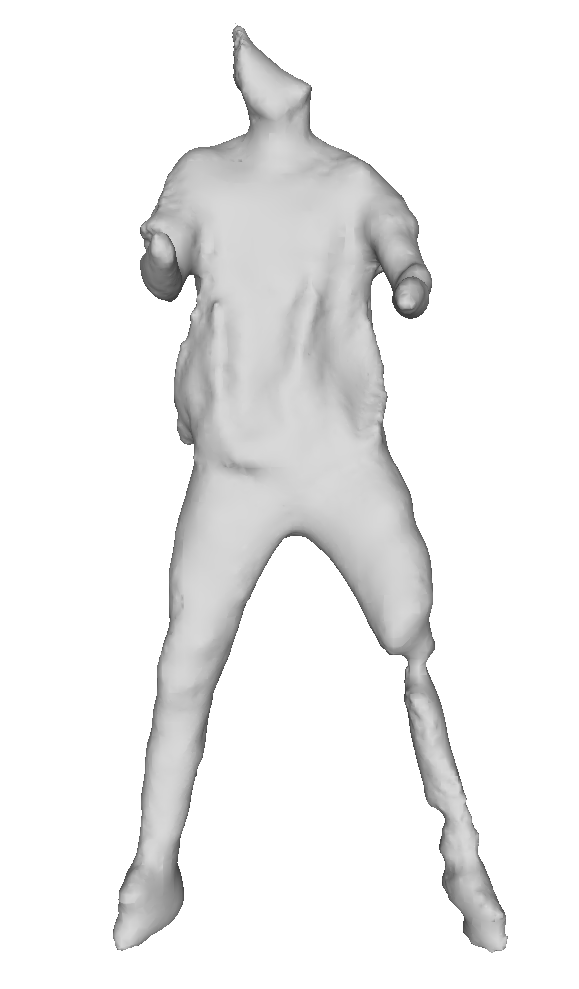}
    \includegraphics[width=1.3cm]{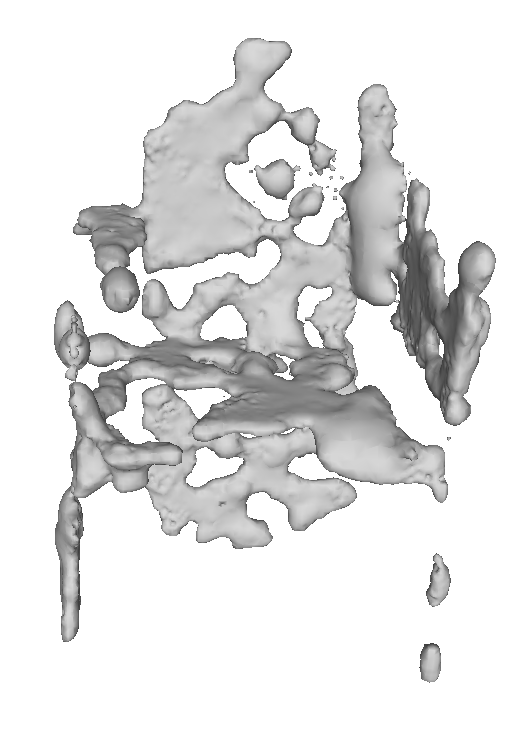}
    \includegraphics[width=1.3cm]{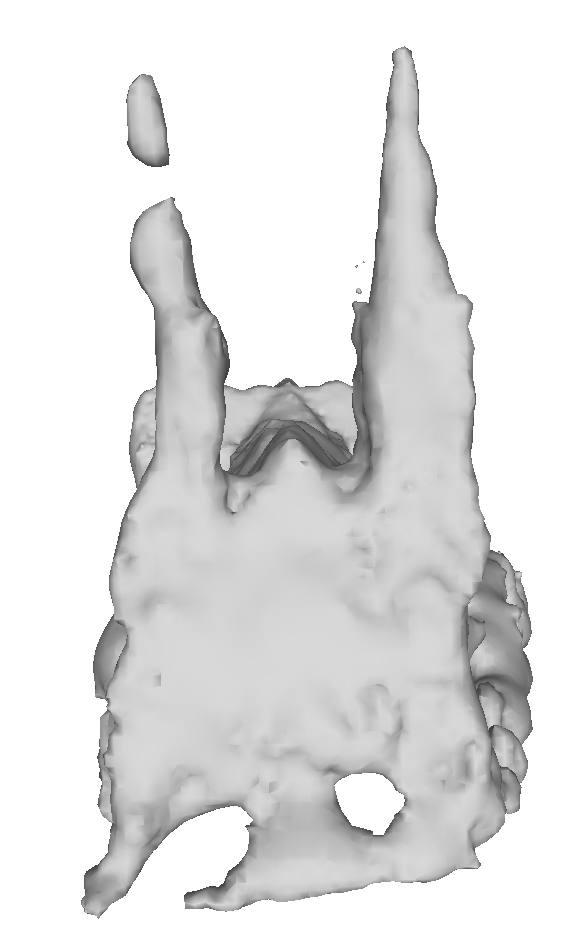}
    \includegraphics[width=1.3cm]{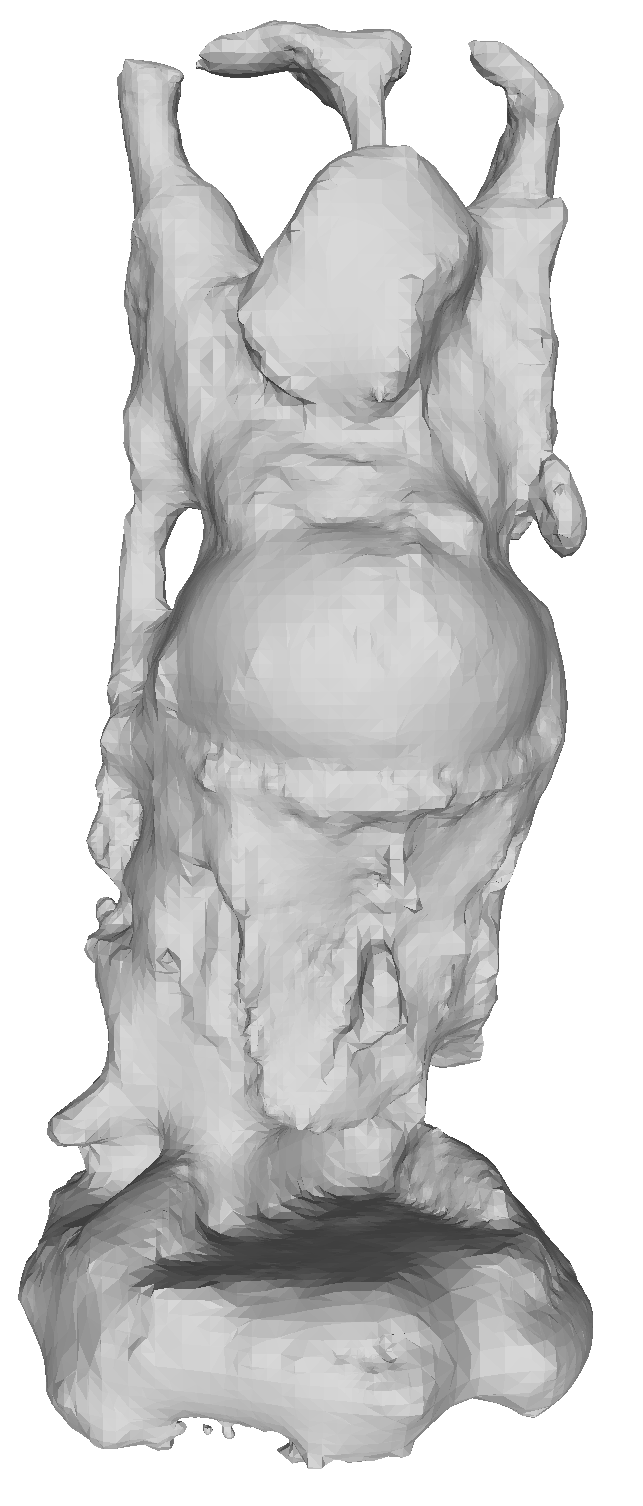}
    \includegraphics[width=1.3cm]{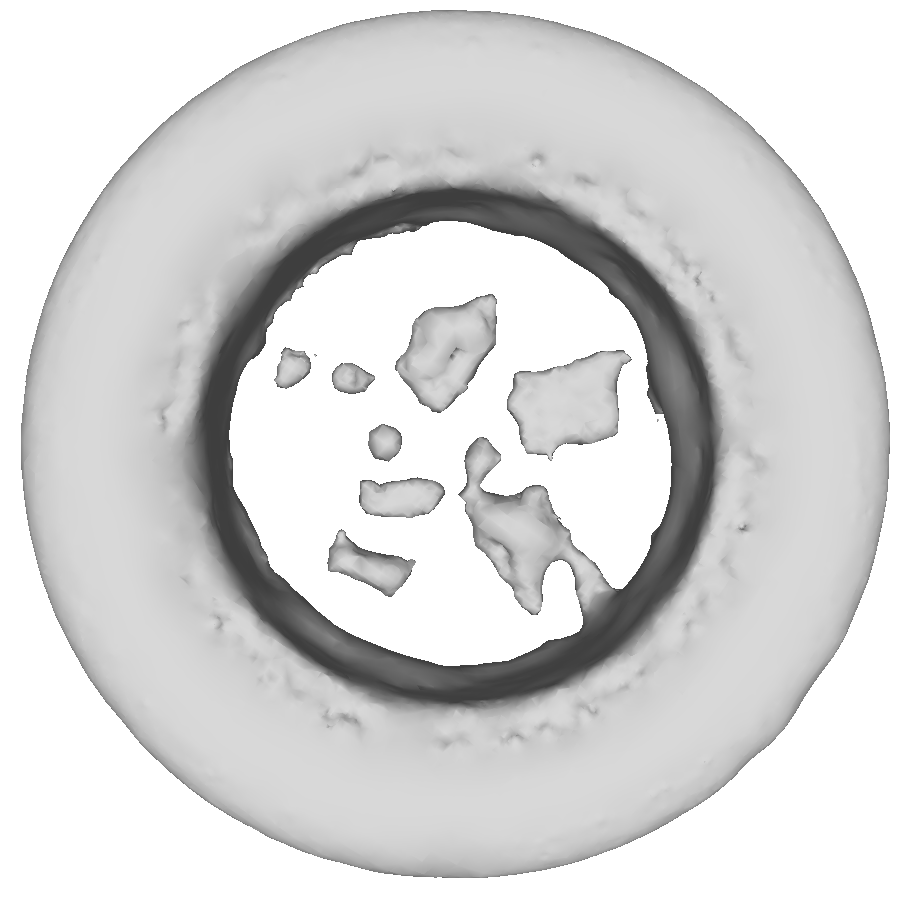}
    \includegraphics[width=1.3cm]{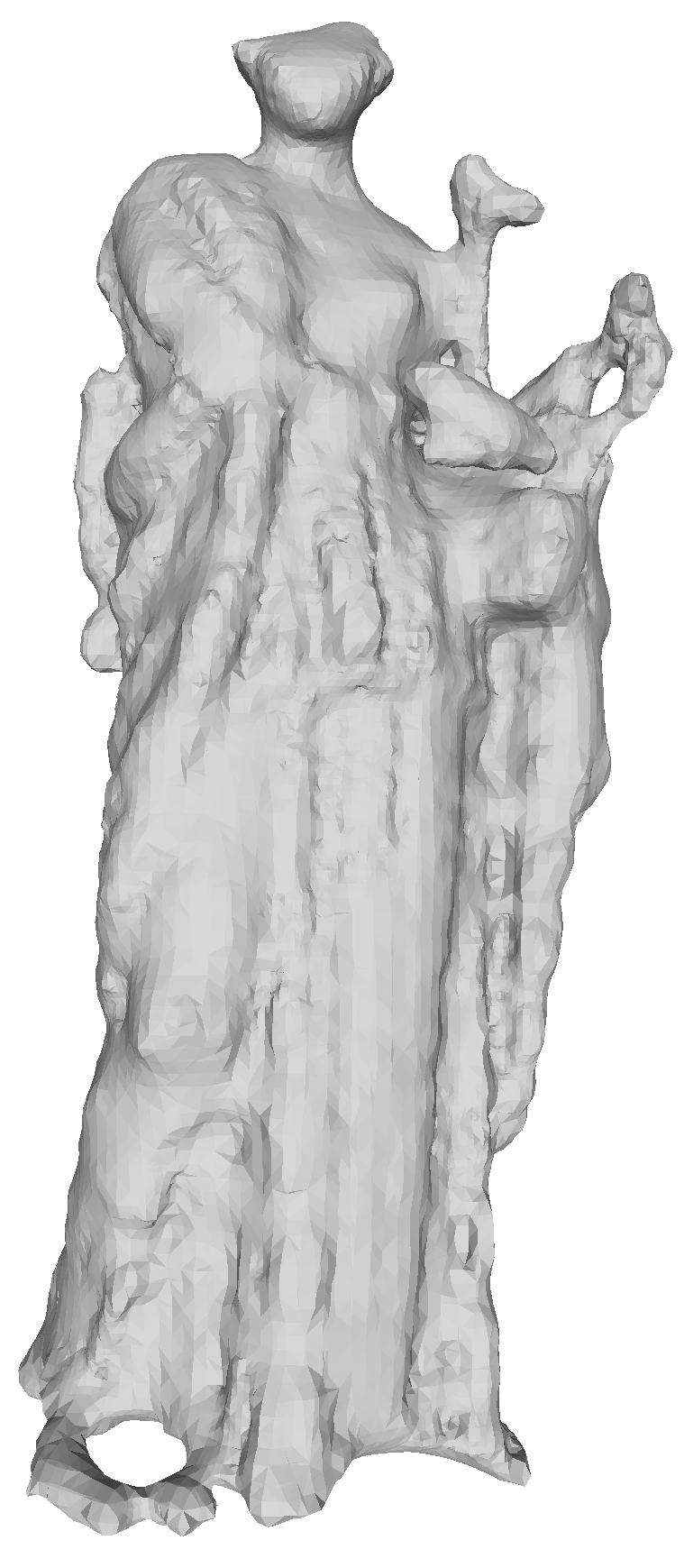}
    
    \includegraphics[width=1.3cm]{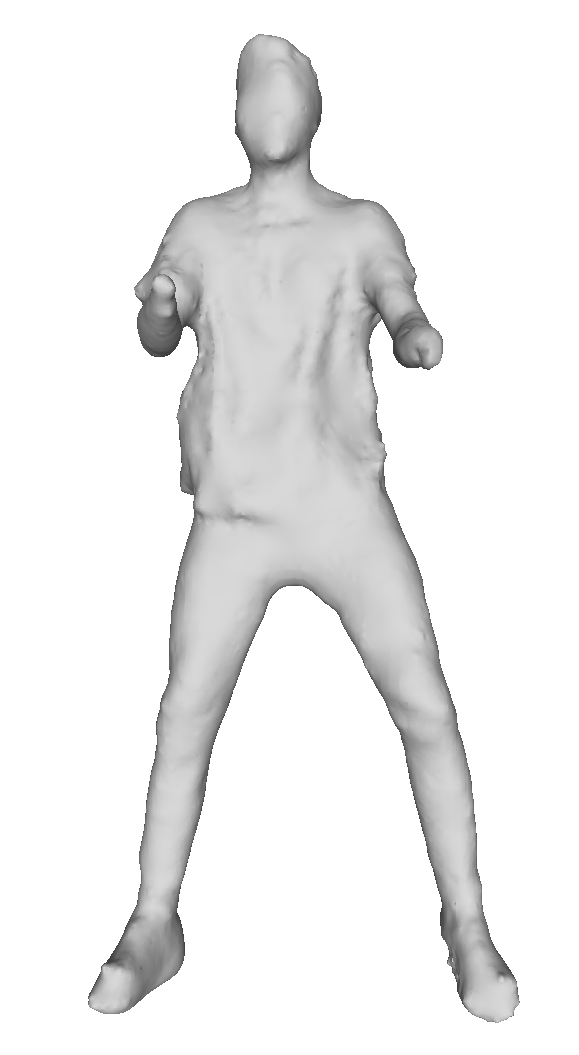}
    \includegraphics[width=1.3cm]{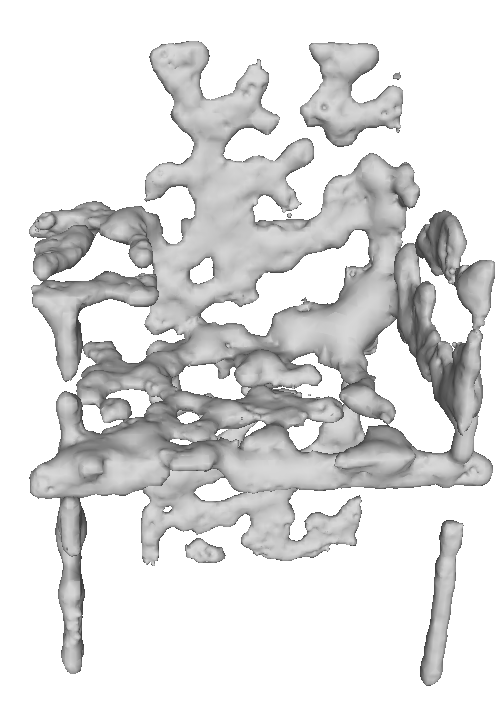}
    \includegraphics[width=1.3cm]{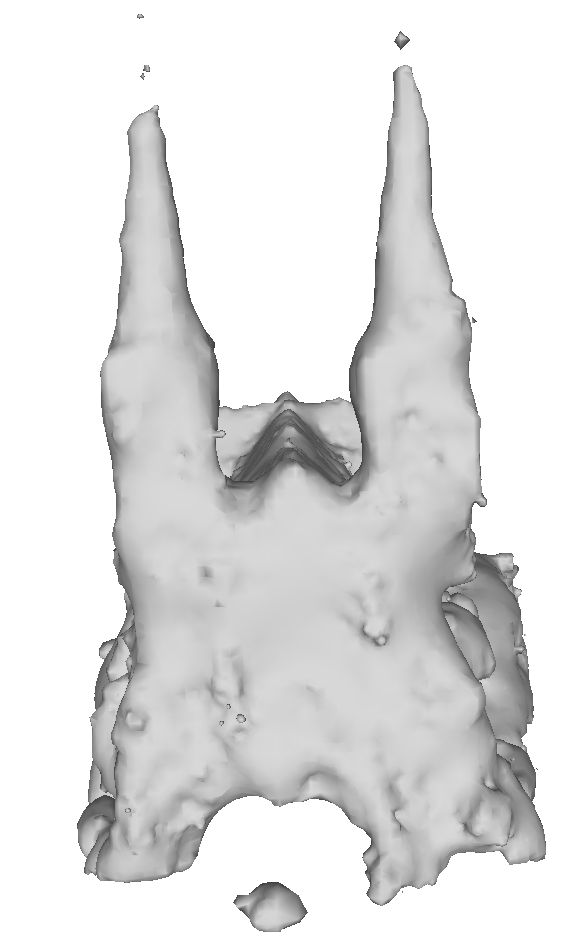}
    \includegraphics[width=1.3cm]{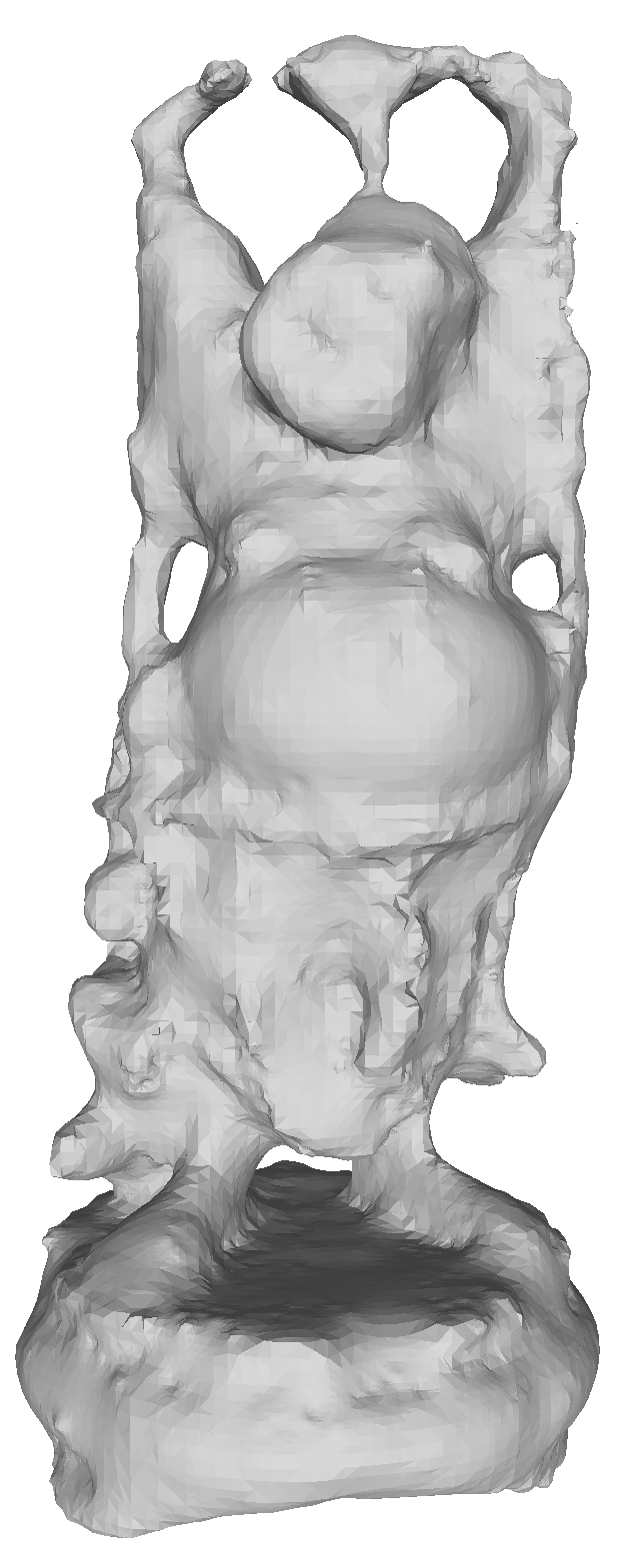}
    \includegraphics[width=1.3cm]{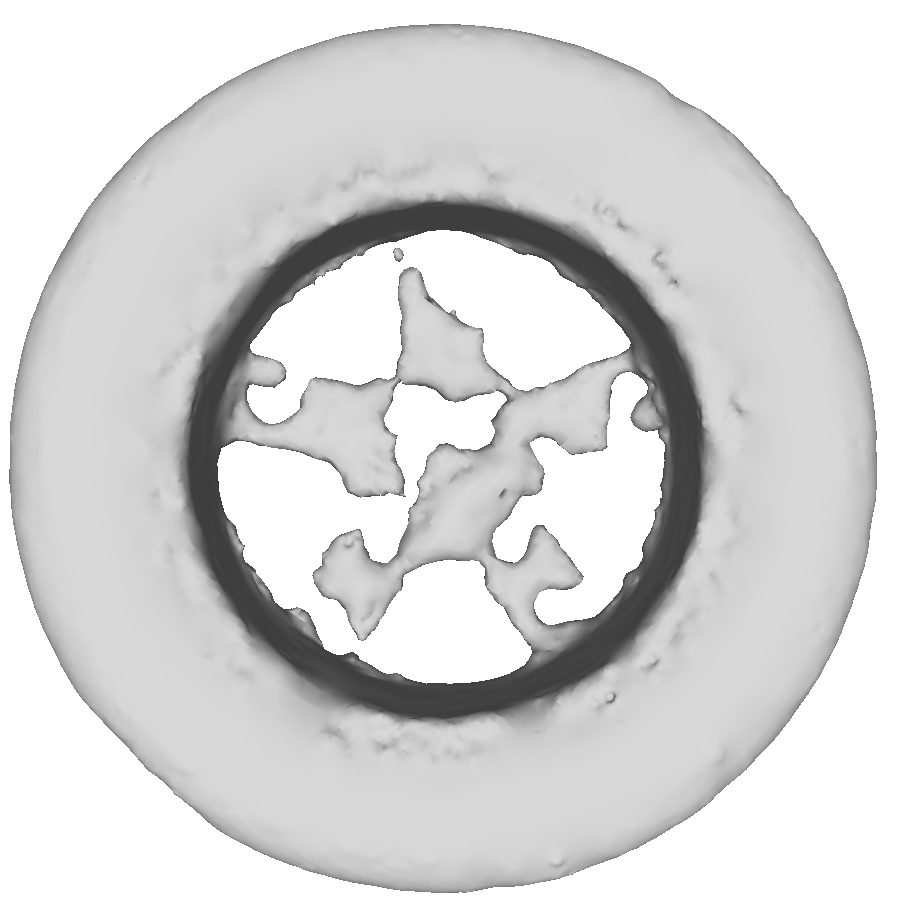}
    \includegraphics[width=1.3cm]{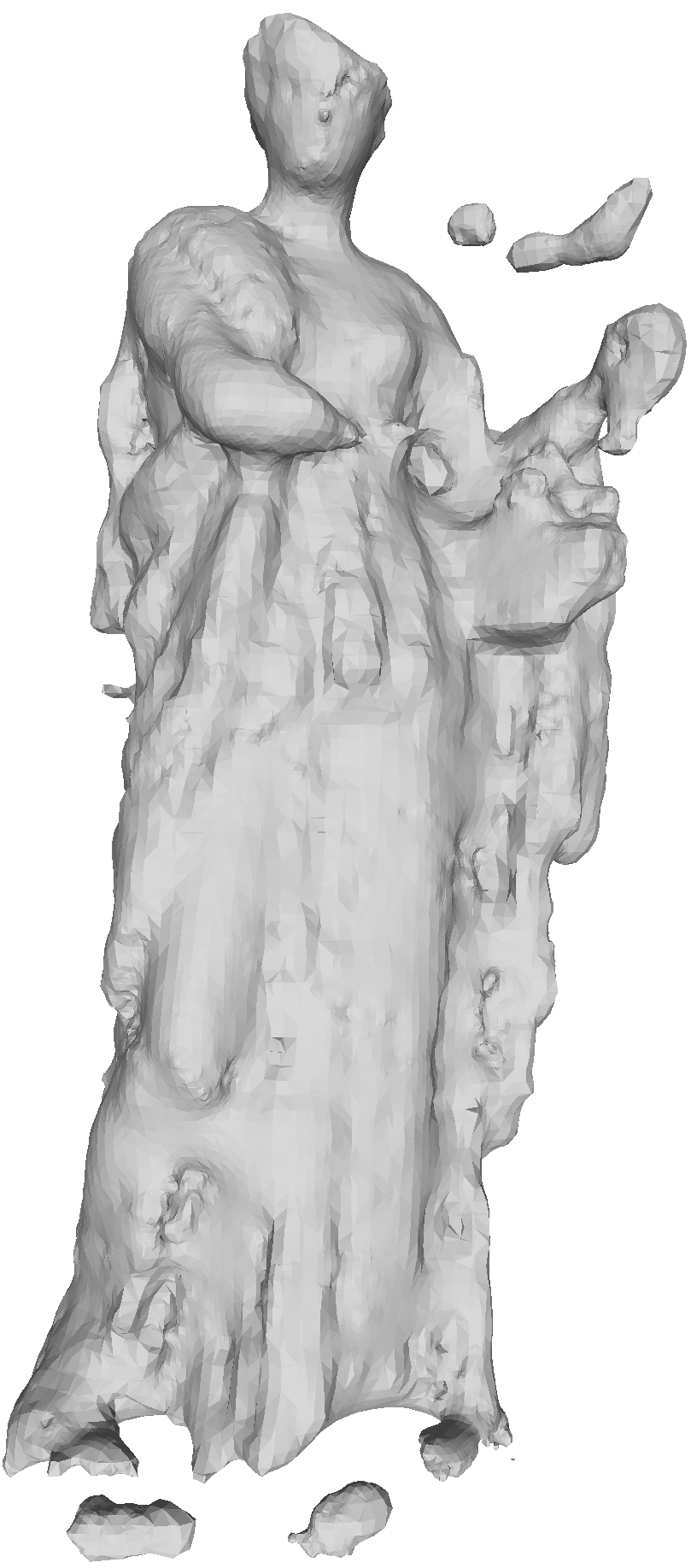}
    
    \includegraphics[width=1.3cm]{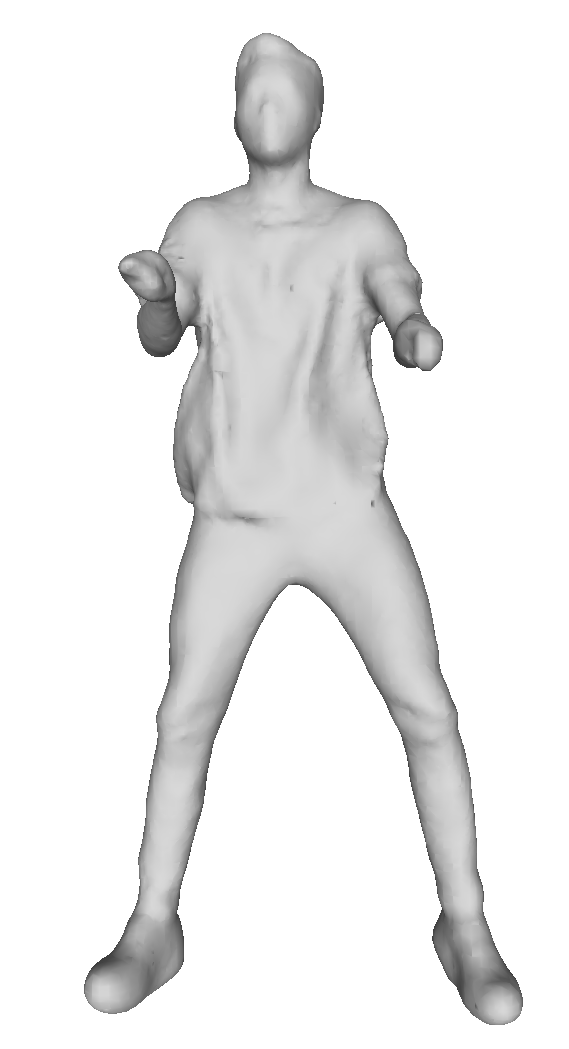}
    \includegraphics[width=1.3cm]{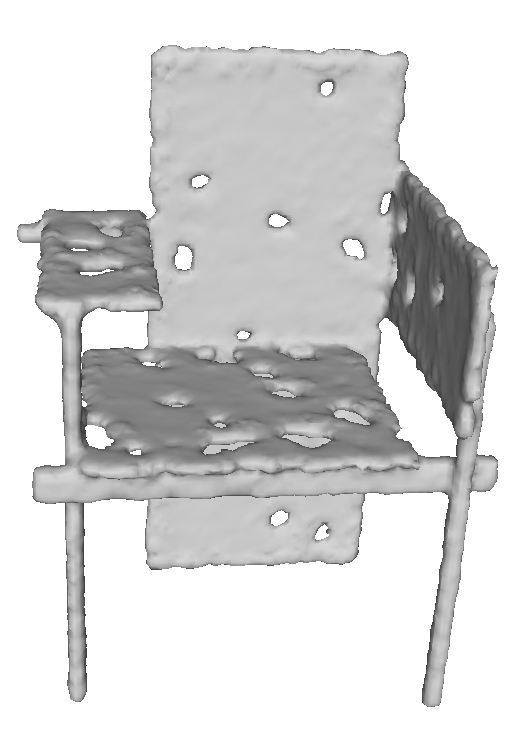}
    \includegraphics[width=1.3cm]{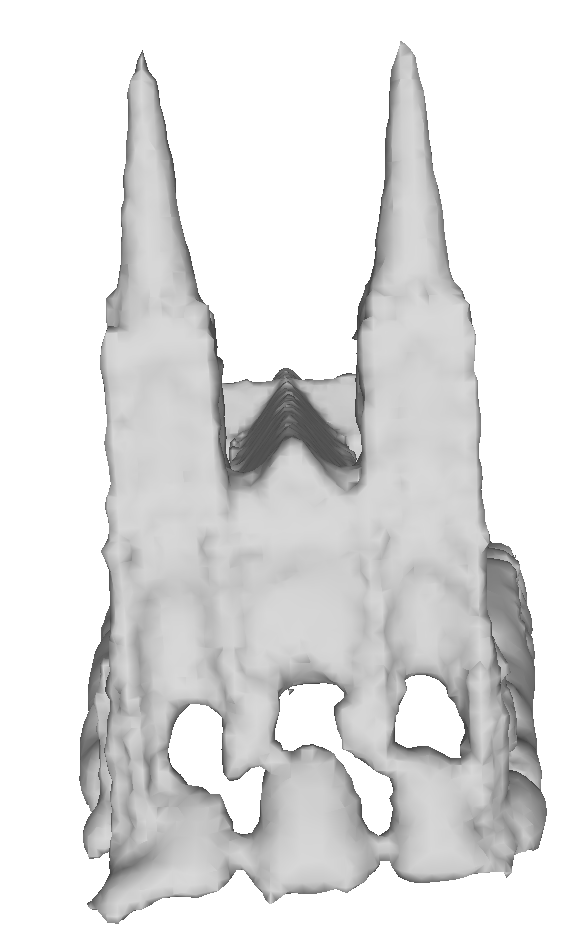}
    \includegraphics[width=1.3cm]{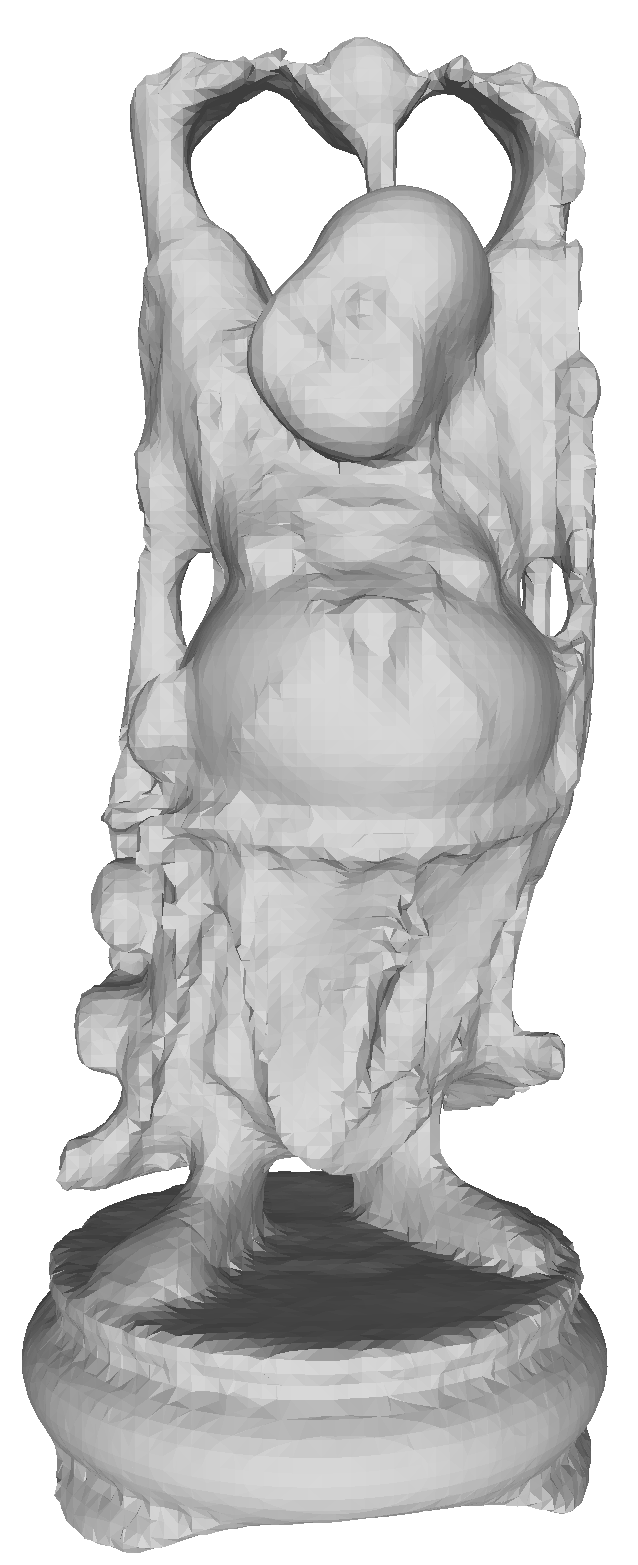}
    \includegraphics[width=1.3cm]{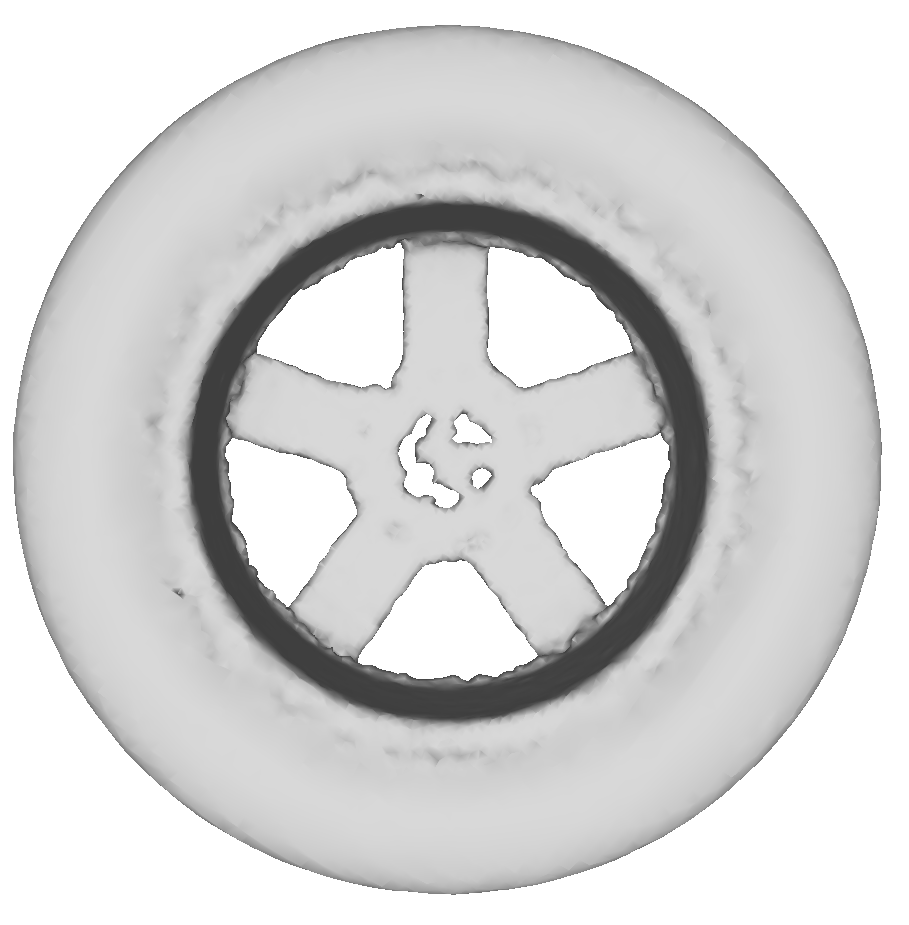}
    \includegraphics[width=1.3cm]{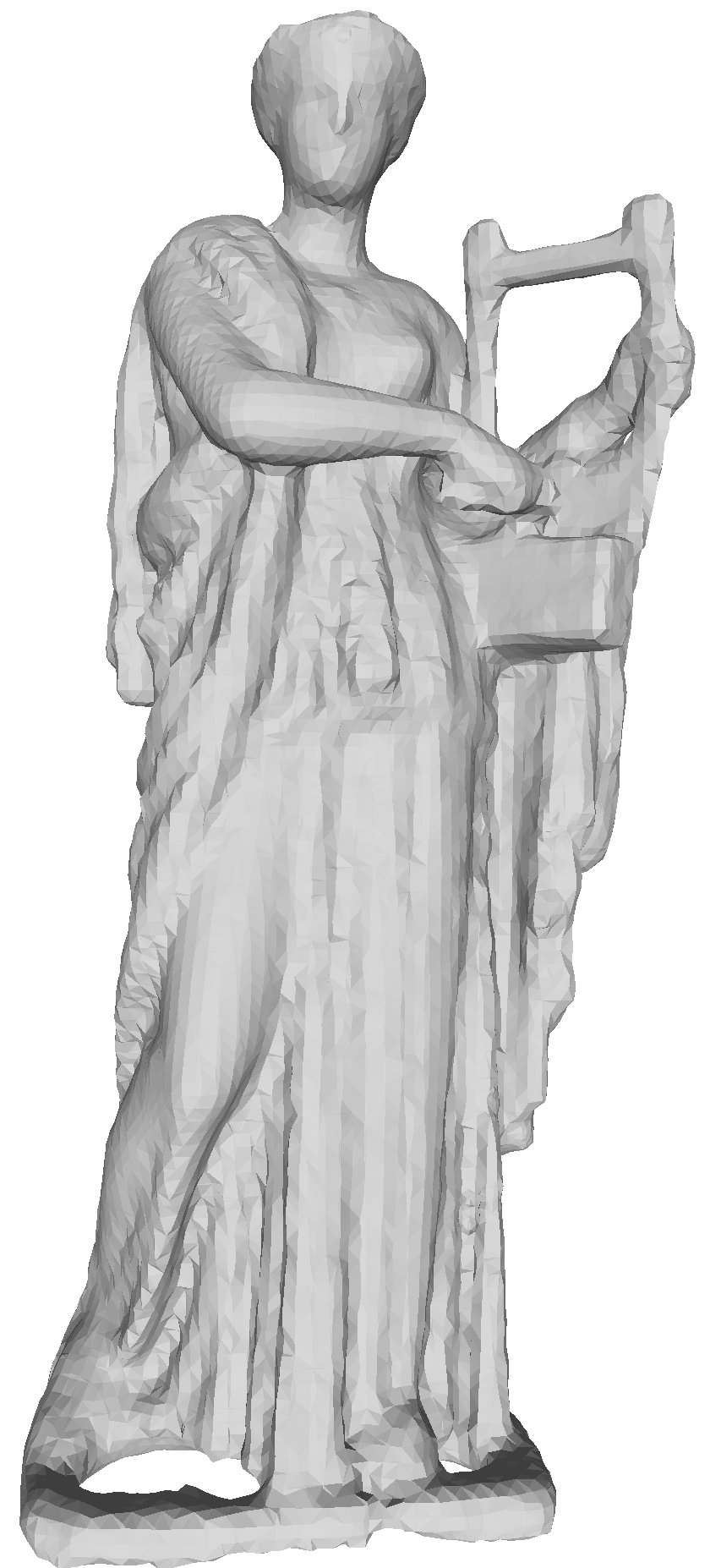}
    
    \includegraphics[width=1.3cm]{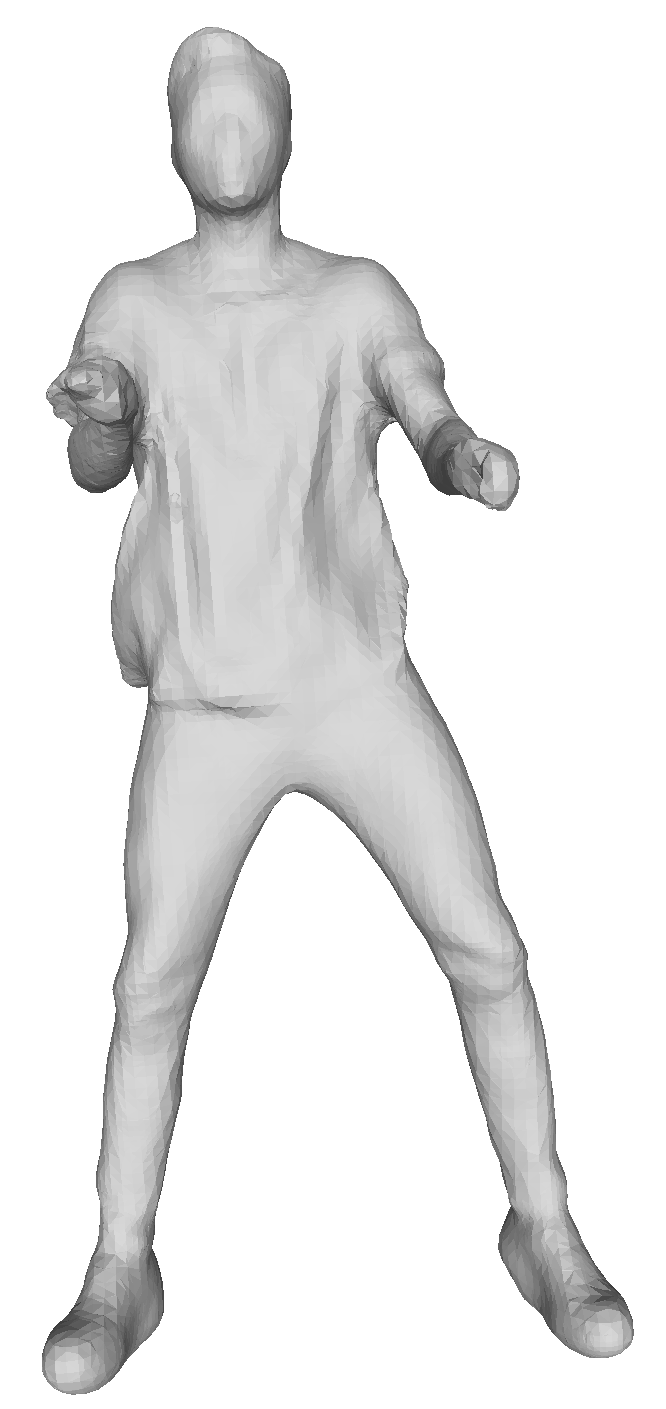}
    \includegraphics[width=1.3cm]{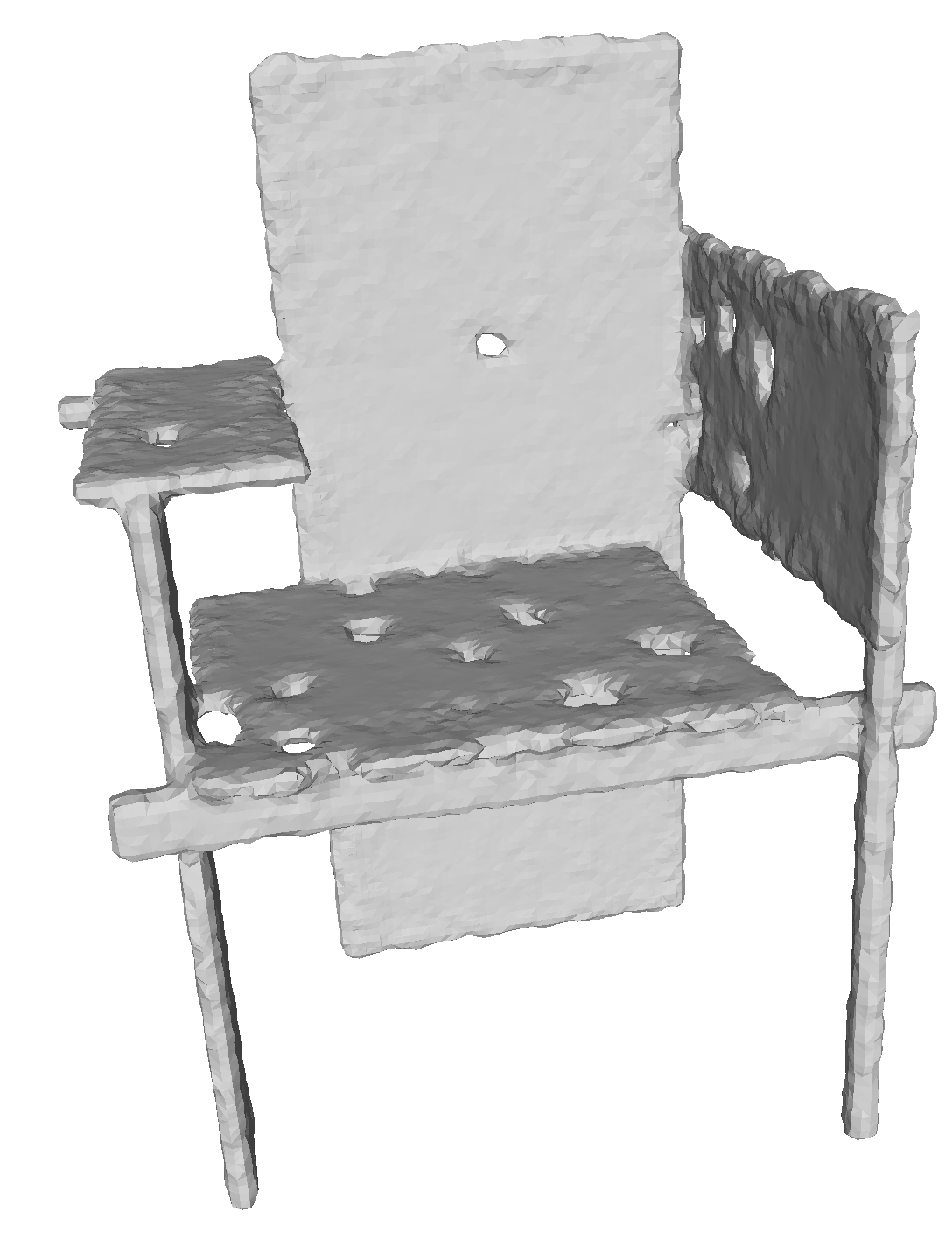}
    \includegraphics[width=1.3cm]{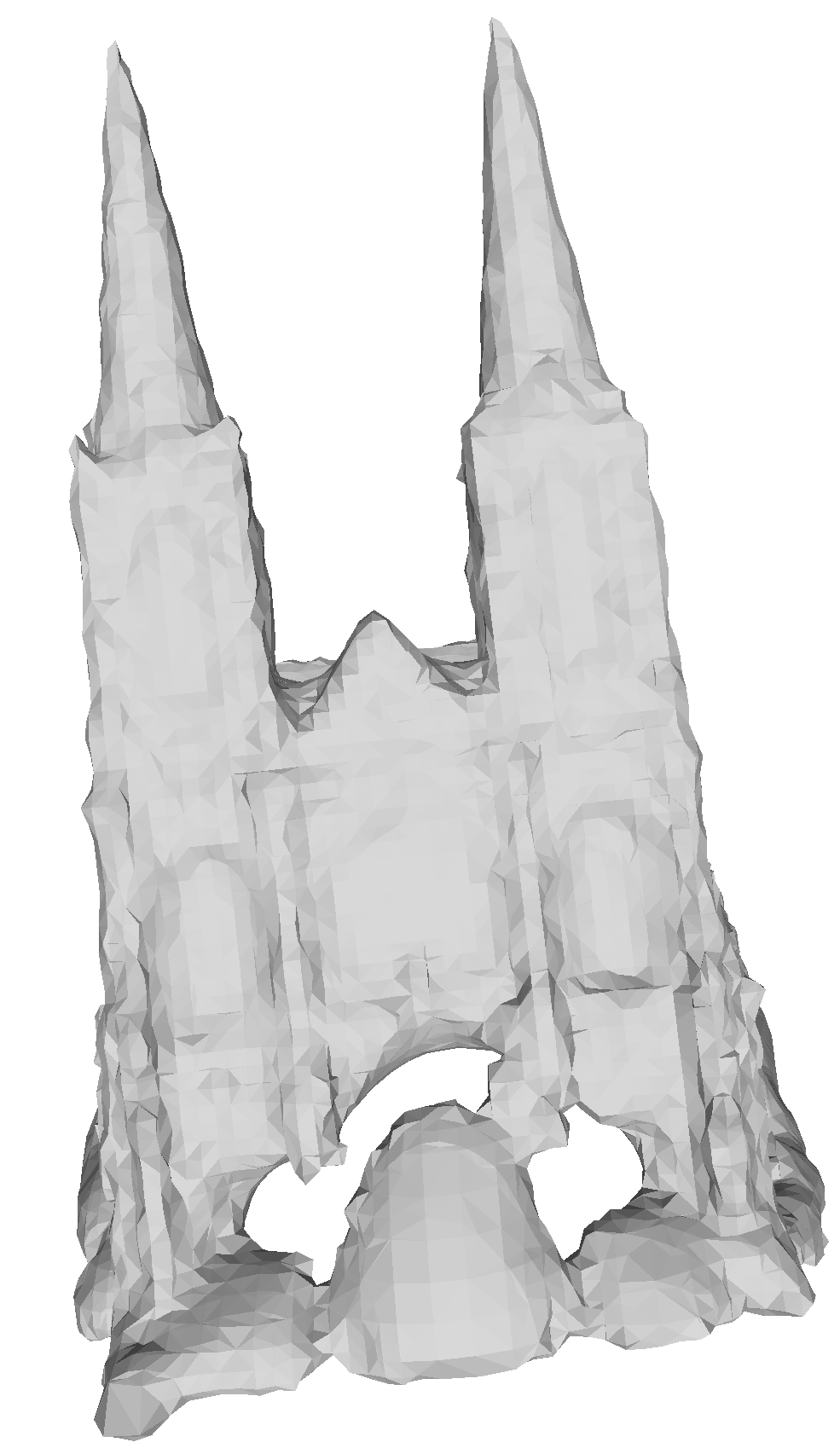}
    \includegraphics[width=1.3cm]{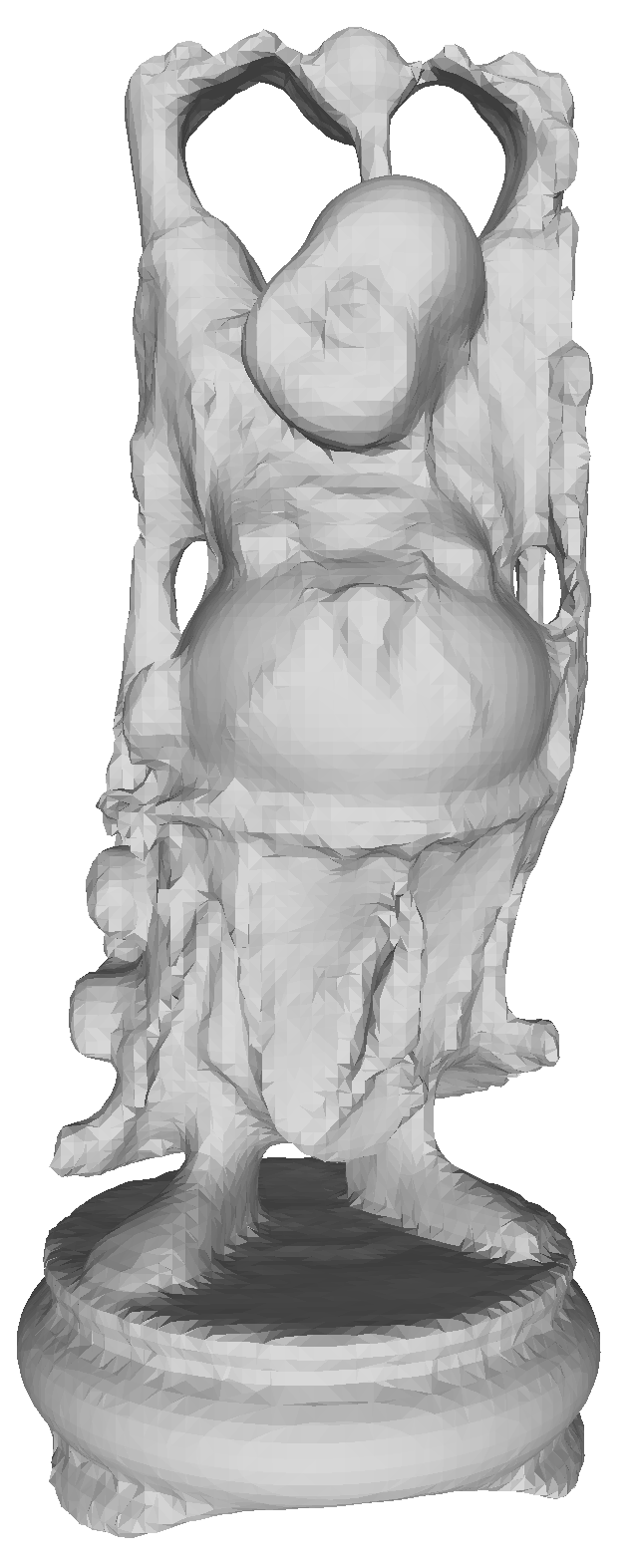}
    \includegraphics[width=1.3cm]{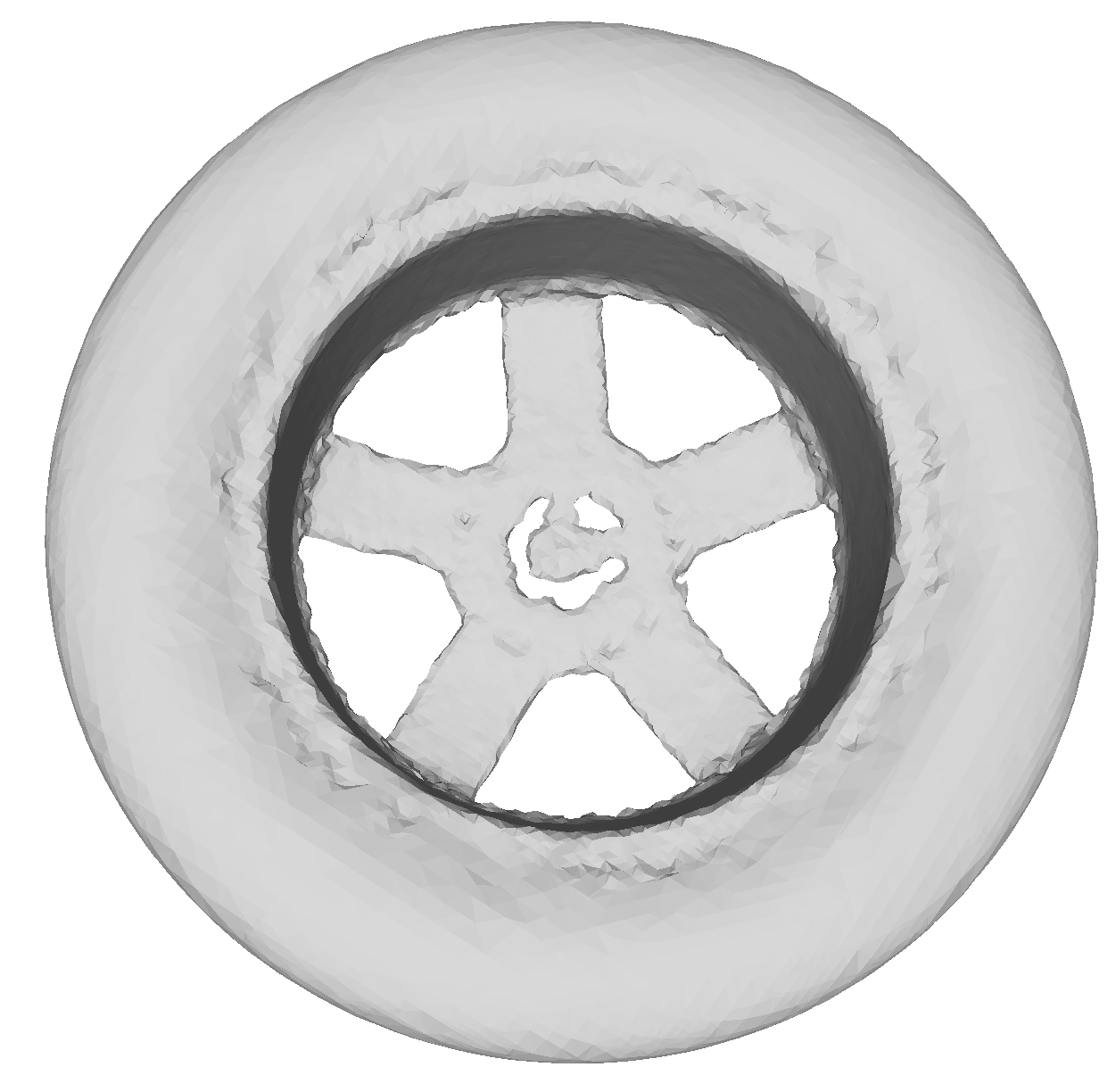}
    \includegraphics[width=1.3cm]{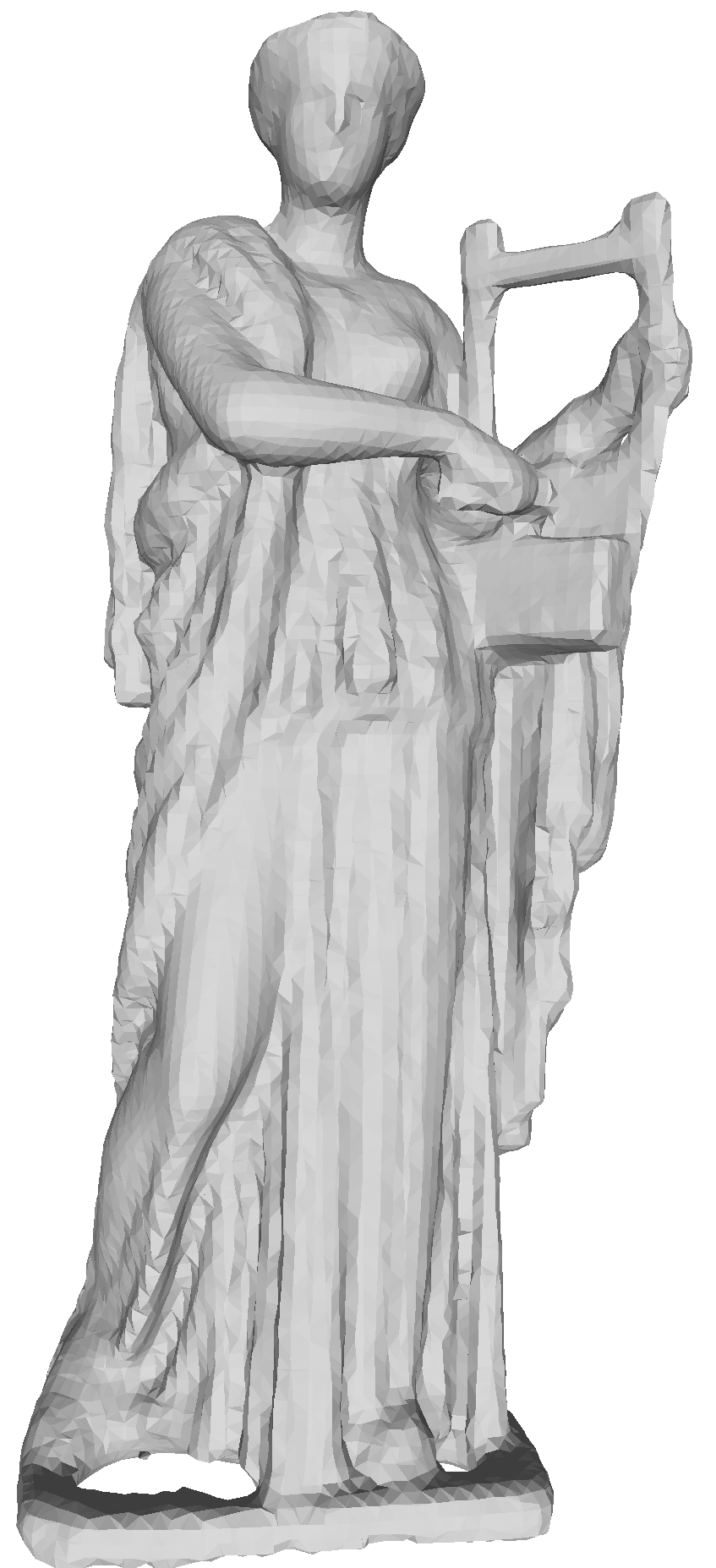}
    \caption{Row 1: Original static 3D point clouds. Rows 2-4:  2D representations of the structured point clouds through Flattening-Net~\cite{zhang2023flattening}, RegGeoNet~\cite{zhang2022reggeonet}, and our method, respectively. Rows 5-8: Surfaces reconstructed from the structured point clouds by Flattening-Net~\cite{zhang2023flattening}, RegGeoNet~\cite{zhang2022reggeonet}, and our method, and the original points, respectively.}
    \label{fig:quality_static}
\end{figure}

\begin{figure}[t]
    \centering
    \subfigure[]{\includegraphics[width=0.32\linewidth]{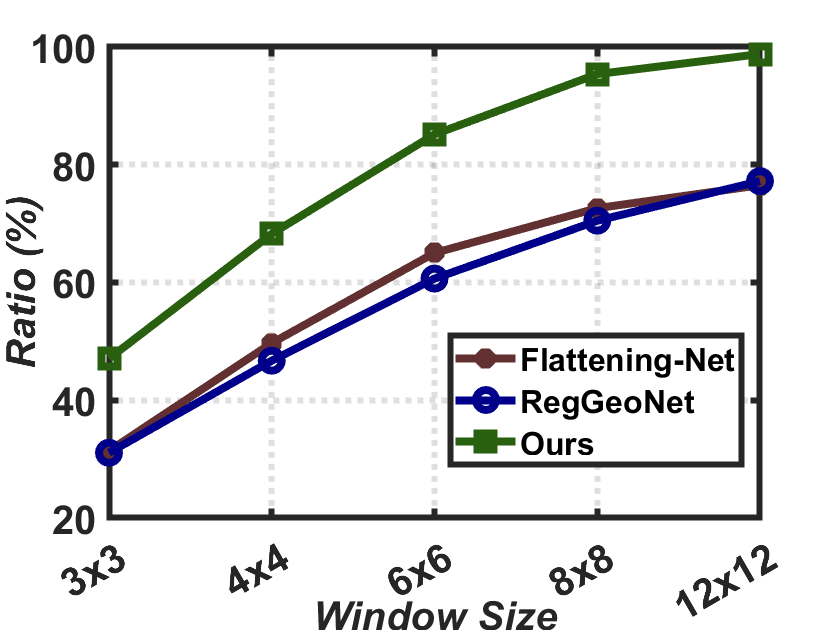}}\label{fig_dancer}
    \subfigure[]{\includegraphics[width=0.32\linewidth]{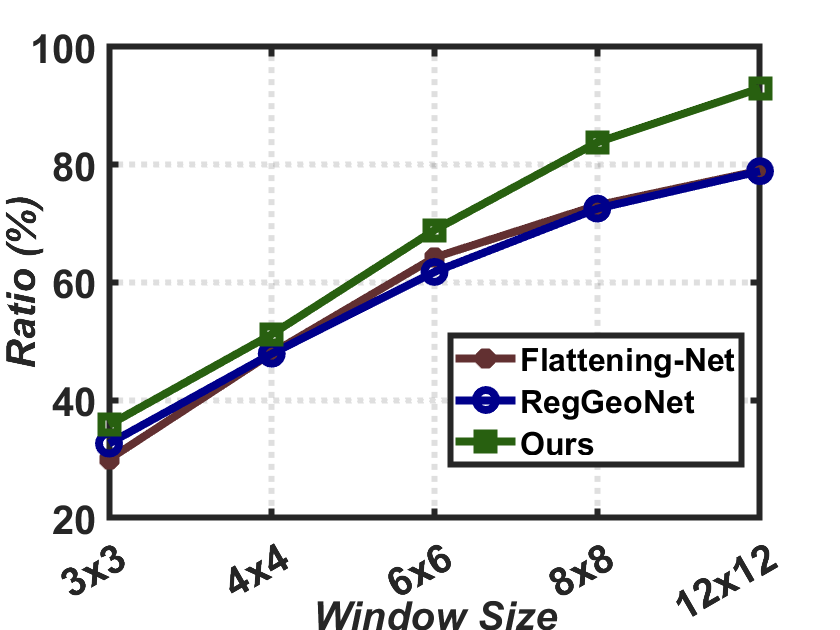}}\label{fig_chair} \vspace{-0.3cm}
    \subfigure[]{\includegraphics[width=0.32\linewidth]{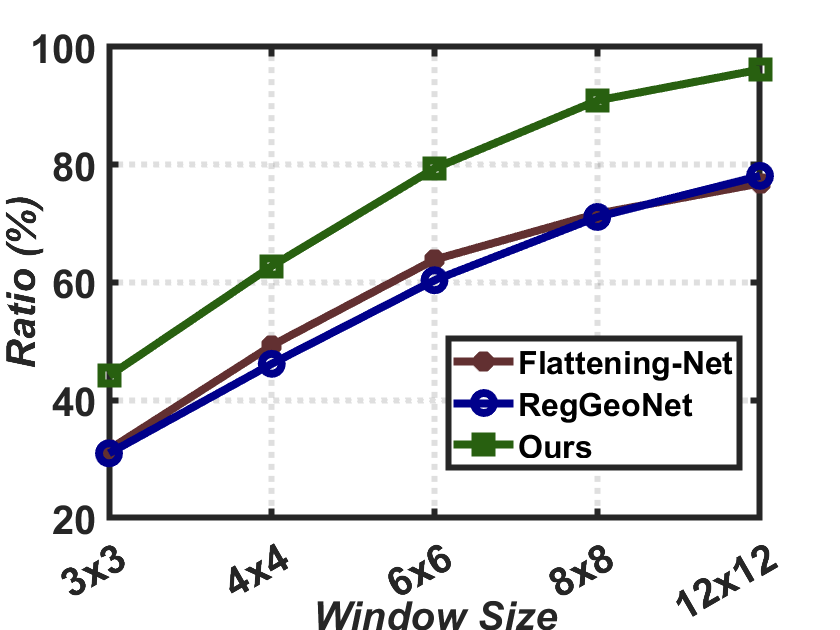}}\label{fig_church}
    
    \subfigure[]{\includegraphics[width=0.32\linewidth]{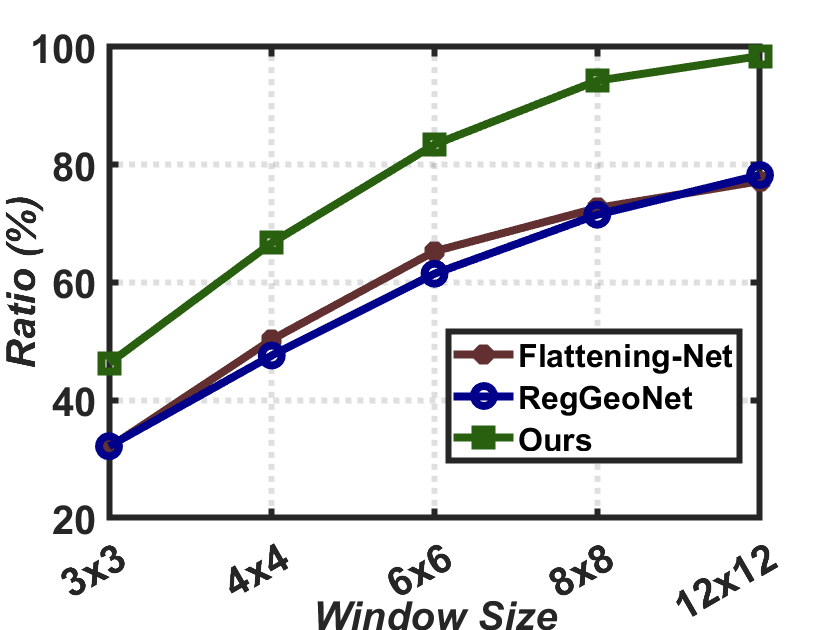}}\label{fig_buddha}
    \subfigure[]{\includegraphics[width=0.32\linewidth]{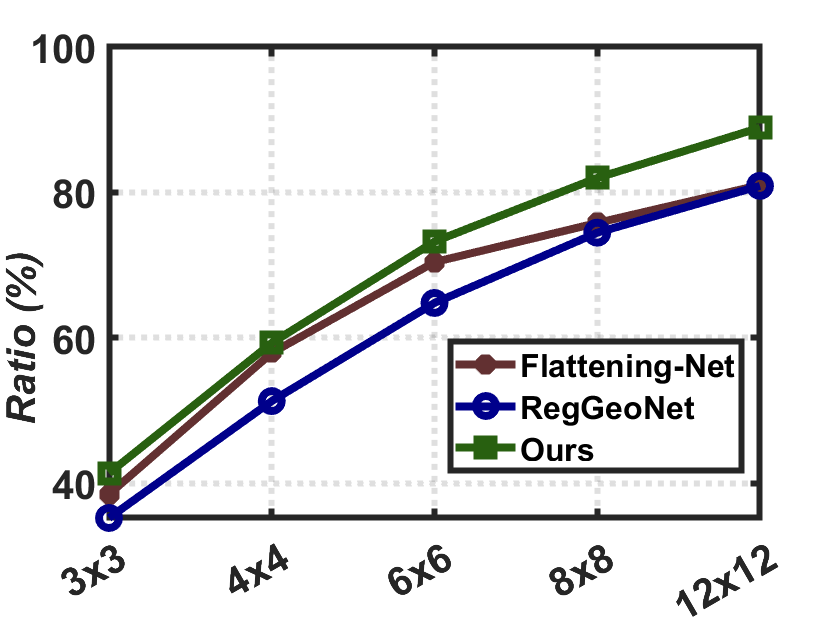}}\label{fig_wheel}
    \subfigure[]{\includegraphics[width=0.32\linewidth]{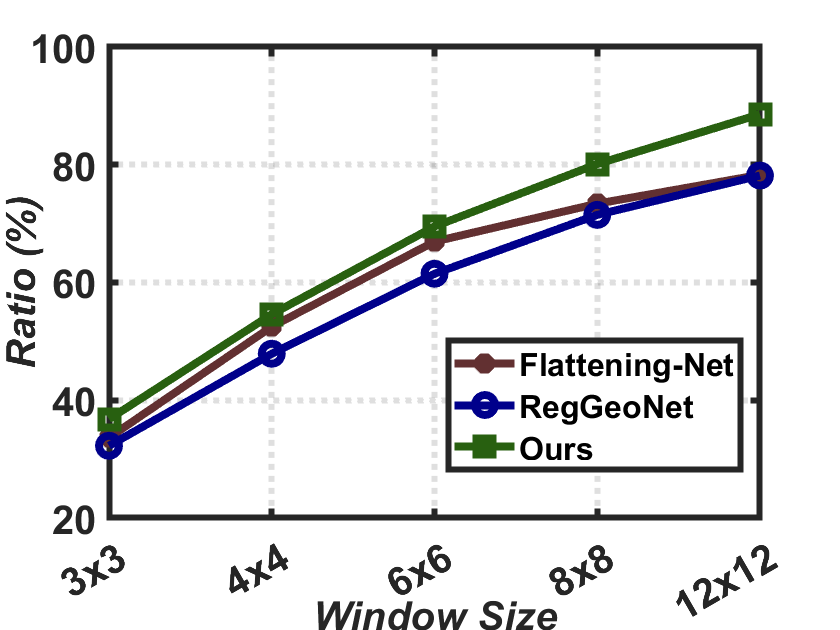}}\label{fig_erato}
    \vspace{-0.3cm}
    \caption{Quantitative comparisons of the spatial smoothness of the structured 3D point clouds by our method, RegGeoNet~\cite{zhang2022reggeonet}, and Flattening-Net~\cite{zhang2023flattening}. (a) Dancer (b) Chair (c) Church (d) Buddha (e) Wheel (f) Erato.}
    \vspace{-0.5cm}
    \label{reconstruction_smoothness}
\end{figure}

\section{Experiments} \label{S_Exp}

In Section \ref{sec_rep_qua}, we quantitatively and qualitatively evaluated the representation capability of our framework by quantifying the spatial smoothness, temporal consistency, and geometric fidelity of SPCVs. Sections \ref{sec_app_act} to \ref{sec_app_compre} evaluate the three SPCV-based point cloud analysis and processing tasks. Section \ref{sec_abl_stu} presents in-depth ablation studies focusing on key components within our SPCV representation framework.

\begin{table*}[t]
    \centering
    \caption{Quantitative comparisons of geometric fidelity of the structured static point clouds by different methods in terms of CD ($\times 10^{-4}$), HD ($\times 10^{-2}$), and mNUC ($\times 10^{-3}$). For all three metrics, the smaller, the better.}
    \label{tab_reconstruction_result_representation}
    \renewcommand{\arraystretch}{1}
    \begin{tabular}{l|c|c|c|c|c|c|c|c|c}
        \toprule[1.2pt]
        Method & 
        \multicolumn{3}{c|}{Dancer} & 
        \multicolumn{3}{c|}{Chair} & 
        \multicolumn{3}{c}{Church} \\
        \cline{2-10}
        & CD & HD & mNUC & CD & HD & mNUC & CD & HD & mNUC \\
        \hline
        Flattening-Net~\cite{zhang2023flattening} & 1.790 & 6.56 & 2.9 & 2.881 & 8.49 & 2.9 & 1.513 & 6.49 & 3.0 \\
        RegGeoNet~\cite{zhang2022reggeonet} & 0.659 & 6.02 & 3.0 & 1.043 & 4.56 & 2.7 & 1.103 & 4.85 & 3.0 \\
        Ours & 0.034 & 0.65 & 1.0 & 0.086 & 1.13 & 0.7 & 0.090 & 0.97 & 1.1\\
        \bottomrule[0.6pt]
    \end{tabular}
    \vspace{1em}
    \begin{tabular}{l|c|c|c|c|c|c|c|c|c}
        \toprule[0.6pt]
        Method & 
        \multicolumn{3}{c|}{Buddha} &
        \multicolumn{3}{c|}{Wheel} &
        \multicolumn{3}{c}{Erato} \\
        \cline{2-10}
        & CD & HD & mNUC & CD & HD & mNUC & CD & HD & mNUC \\
        \hline
        Flattening-Net~\cite{zhang2023flattening} & 1.493 & 4.90 & 3.1 & 1.934 & 6.94 & 2.5 & 1.528 & 7.88 & 2.9 \\
        RegGeoNet~\cite{zhang2022reggeonet} & 0.895 & 4.87 & 3.2 & 1.595 & 6.26 & 2.9 & 0.858 & 5.04 & 3.0 \\
        Ours & 0.064 & 1.03 & 1.4 & 0.160 & 1.42 & 1.3 & 0.063 &  0.88 & 1.2 \\
        \bottomrule[1.2pt]
    \end{tabular}
    \vspace{-0.4cm}
\end{table*}

\subsection{Evaluation of Representation Quality}\label{sec_rep_qua}

\begin{figure}[htbp]
    \centering
    \subfigure[]{\includegraphics[width=0.45\linewidth]{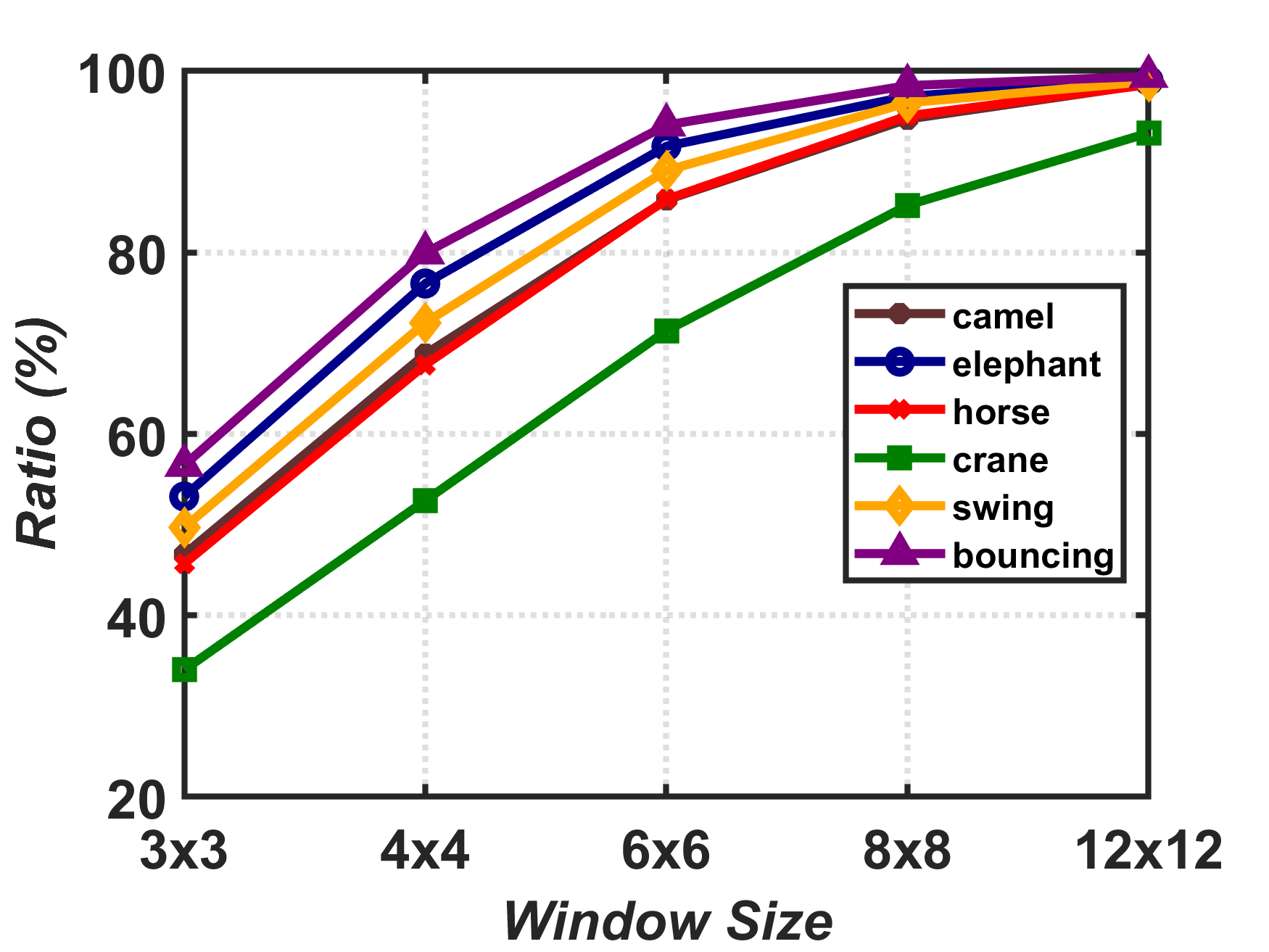}}	
    \subfigure[]{\includegraphics[width=0.45\linewidth]{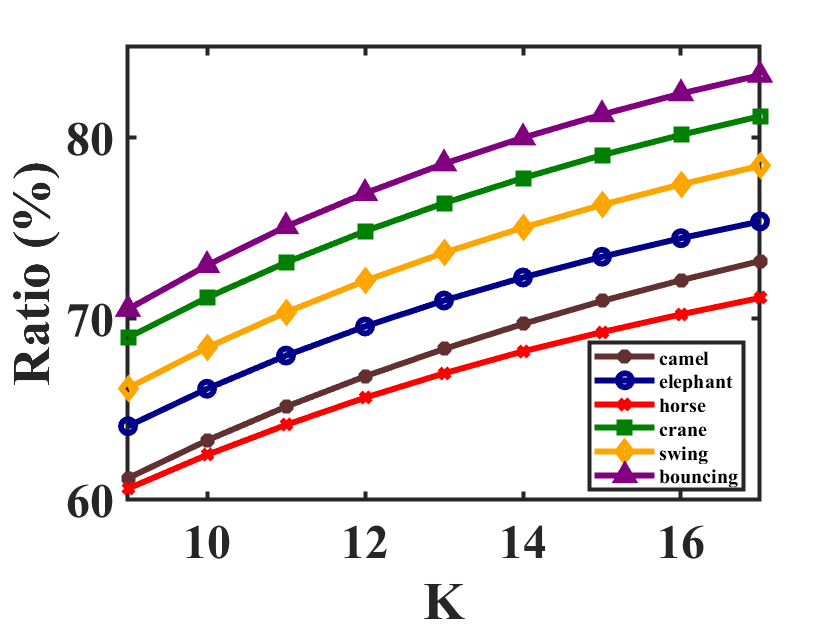}}
    \vspace{-0.3cm}
    \caption{Quantitative evaluation of our SPCVs in terms of (a) spatial smoothness and (b) temporal consistency.}
    \vspace{-0.5cm}
    \label{quant_temporal_consistency_smooth_regular}
\end{figure}

\begin{figure}[t]
    \centering
    \subfigure[]{\includegraphics[width=\linewidth]{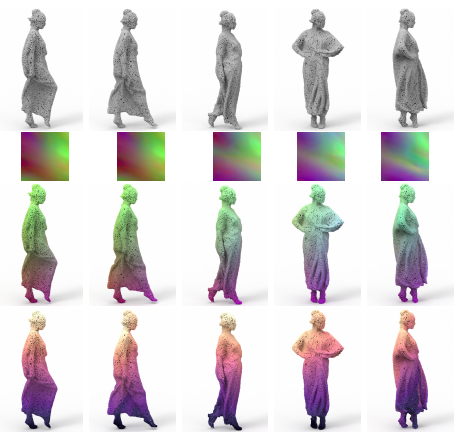}}
    
    \subfigure[]{\includegraphics[width=\linewidth]{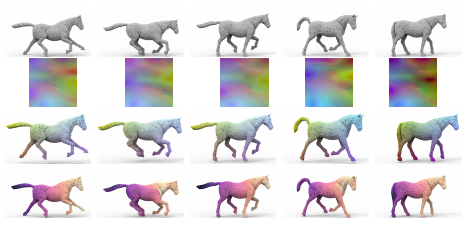}}
        \vspace{-0.4cm}
    \caption{ 
        Visual results of representing the (a) \textit{longdress} and (b) \textit{horse}  sequences as SPCVs.  
        Row 1: Five 3D point cloud frames of a sequence with point-wise correspondences unknown. 
        Row 2: The SPCV representation of the above sequence.
        Row 3: The point cloud frames reprojected from the SPCV, where for each frame, points are rendered with the same colors as corresponding pixels to show 2D-3D correspondence.
        Row 4: The 3D point cloud frames reprojected from the SPCV, where the points with an identical 2D location across different frames are rendered with an identical color to illustrate temporal consistency.
        We also refer readers to \textcolor{magenta}{\textbf{video demo}} included in the \textit{Supplementary Material} for more visual results.}
    \vspace{-0.4cm}
    \label{hum_anim_seq_vis_ours}
\end{figure}

\begin{figure}[htbp]
	\centering
\includegraphics[width=1.0\linewidth]{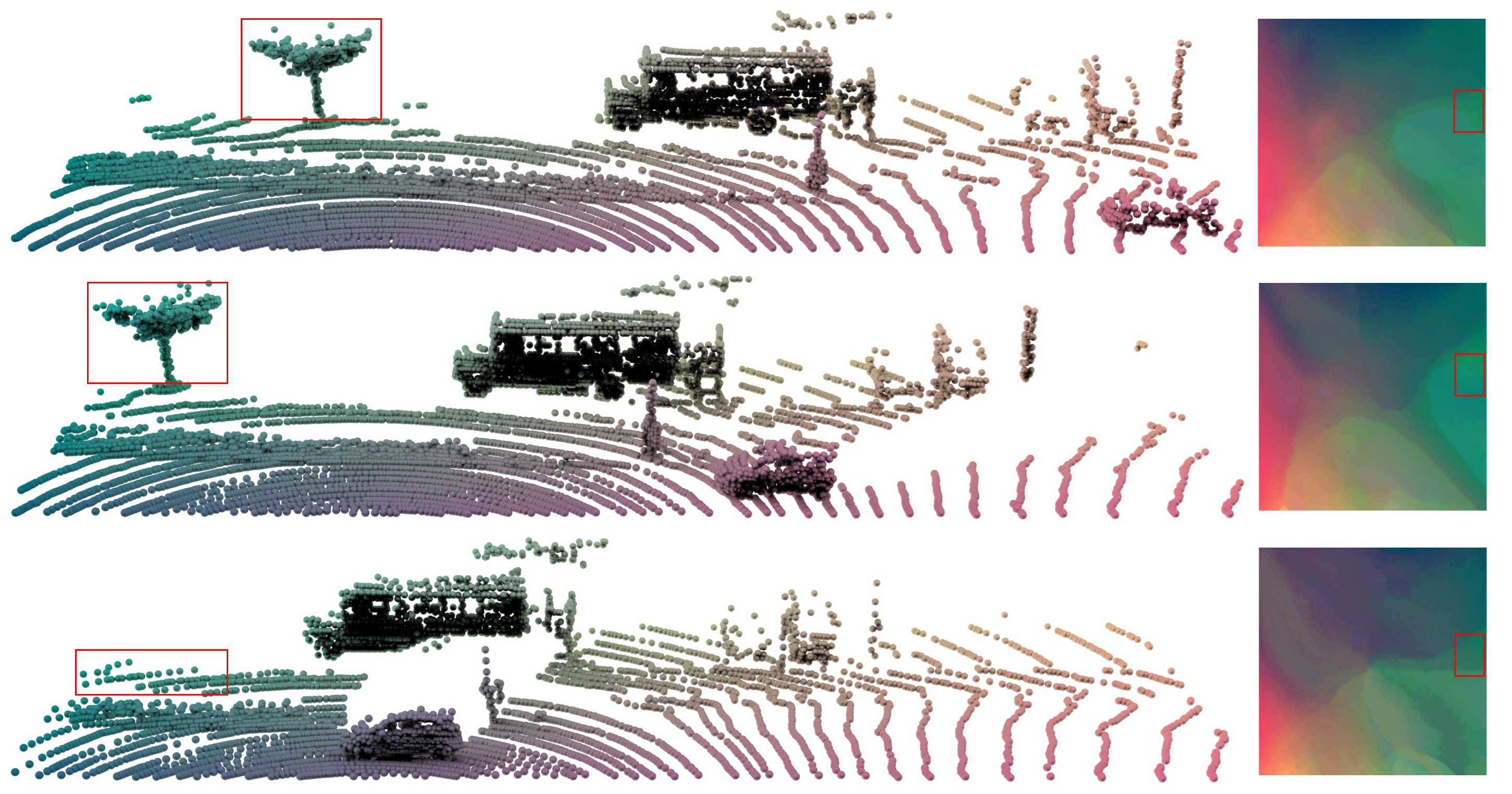}
	\caption{
 Visual results of representing a sequence from the \textit{KITTI} dataset as SPCV.
 }
 \vspace{-0.4cm}
 \label{kitti_3of10_crop_frames_1wpts}
\end{figure}

\begin{figure}[htbp]
    \centering
    \subfigure[]{\includegraphics[width=0.31\linewidth]{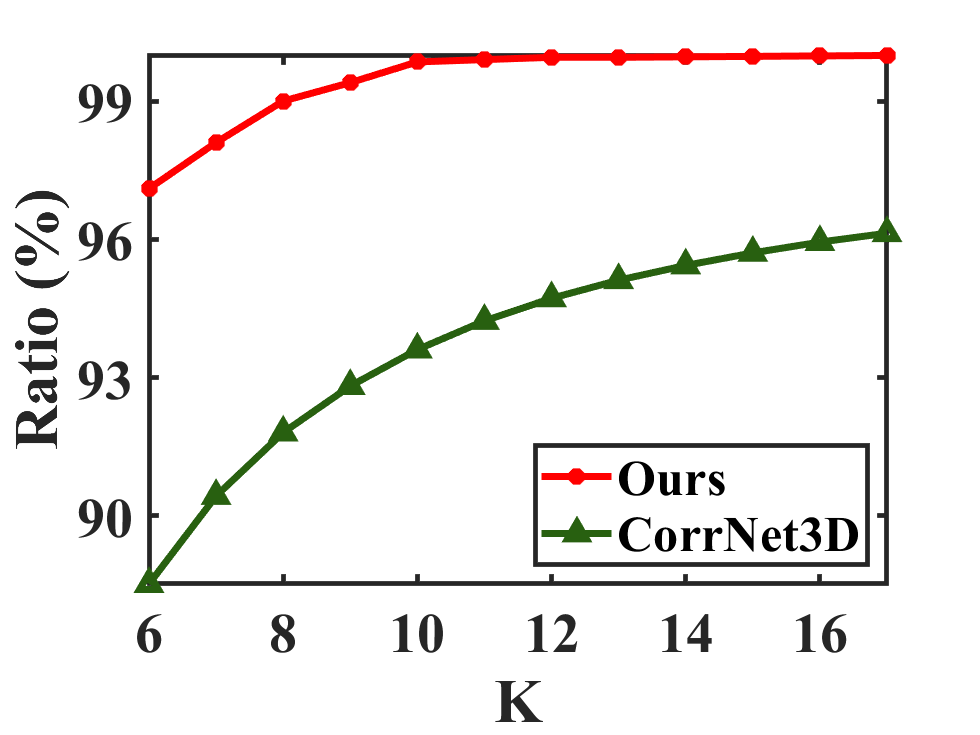}}
    \subfigure[]{\includegraphics[width=0.31\linewidth]{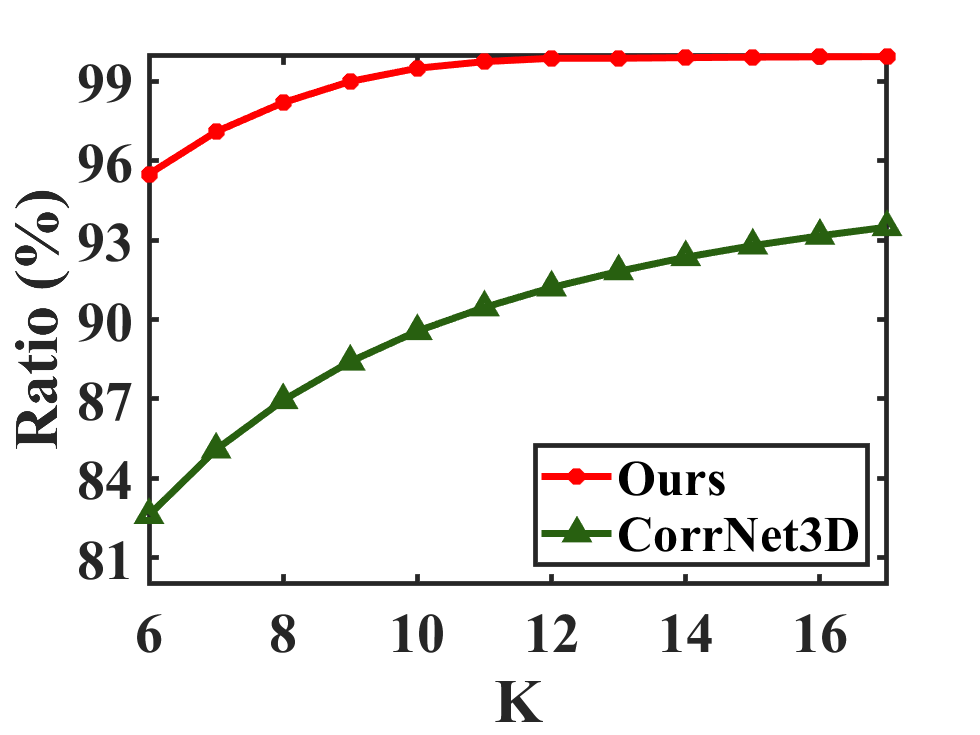}}
    \subfigure[]{\includegraphics[width=0.31\linewidth]{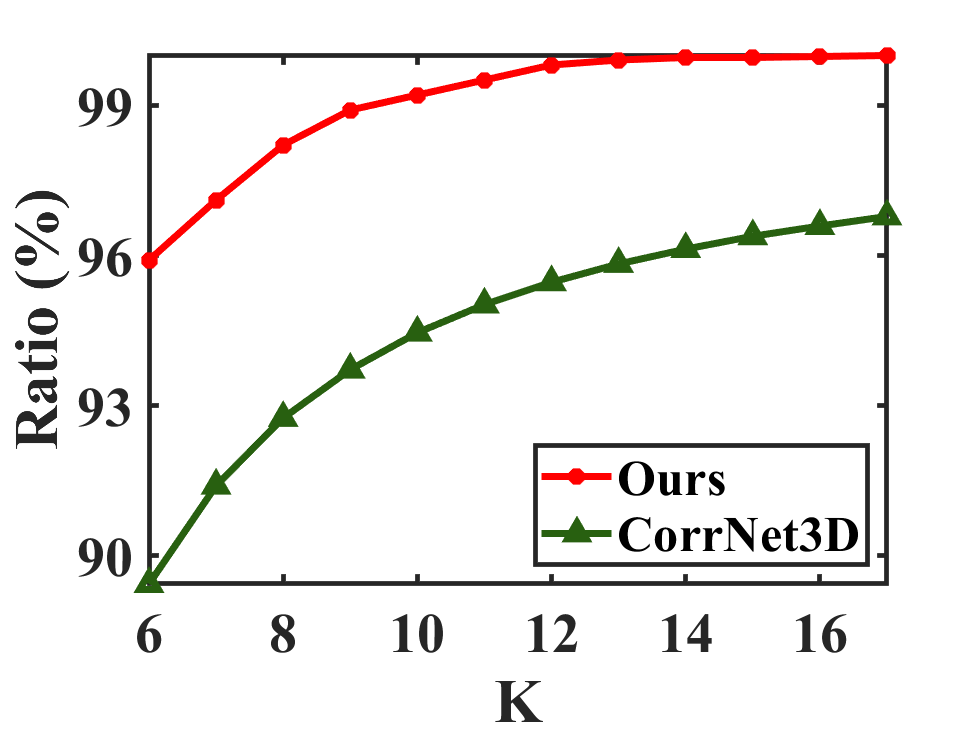}} \vspace{-0.3cm}
    \caption{Quantitative comparisons of temporal consistency obtained by our method with CorrNet3D ~\cite{zeng2021corrnet3d}, an unsupervised method primarily designed for estimating dense correspondence. (a) \textit{Swing}, (b) \textit{Longdress} and (c) \textit{Squat\_2}.}
    \vspace{-0.8cm}
    \label{quant_temporal_consistency_diff_methods}.
\end{figure}
\begin{figure}[t]
    \centering
    \subfigure[]{
	\includegraphics[width=1.9cm]{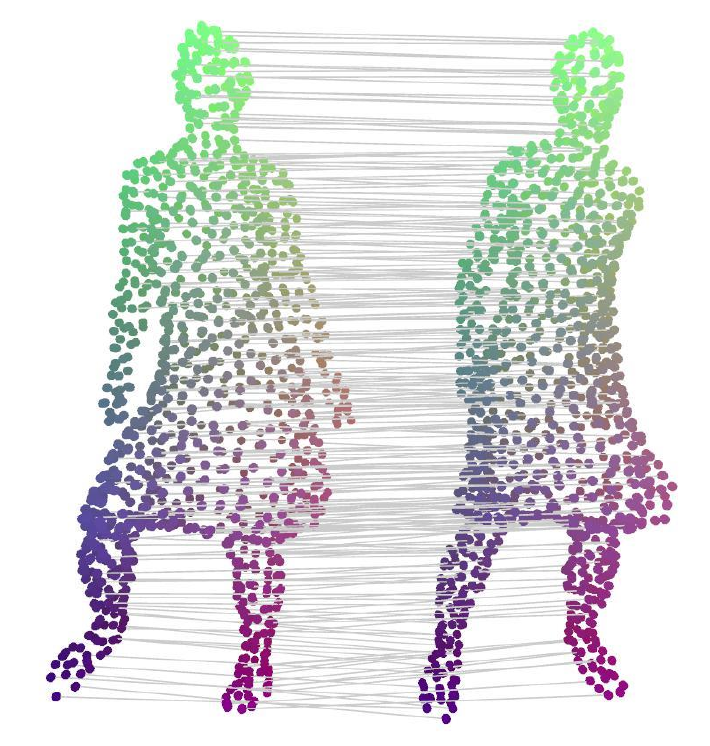}
	\includegraphics[width=2.1cm]{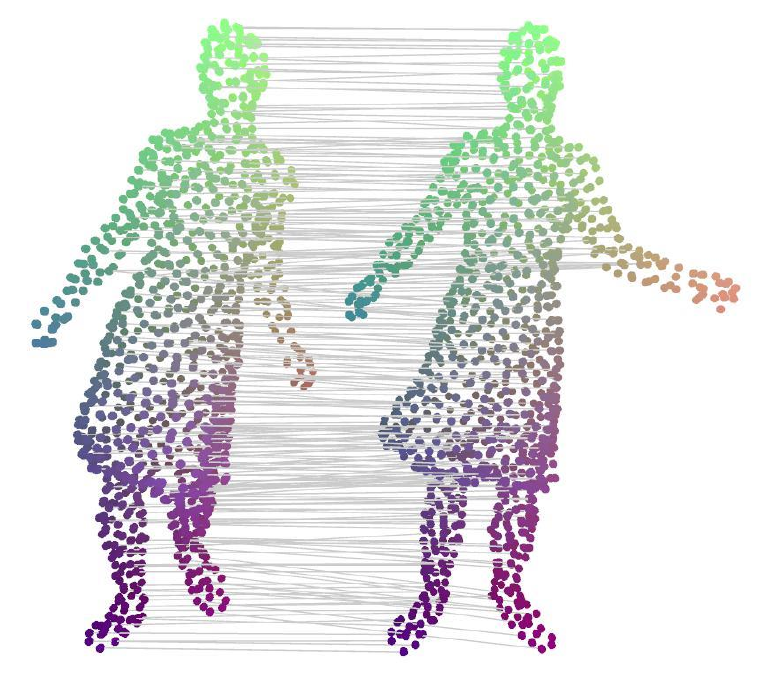}
	\includegraphics[width=1.9cm]{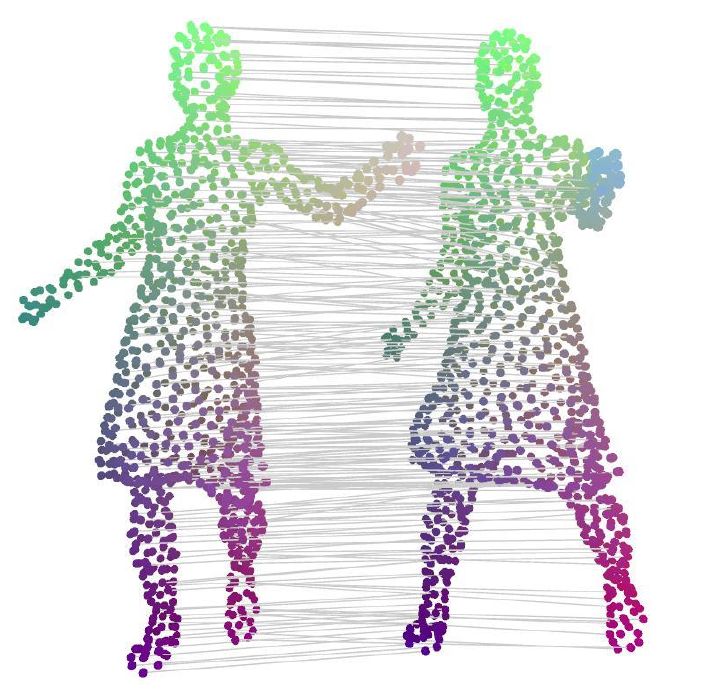}
	\includegraphics[width=1.9cm]{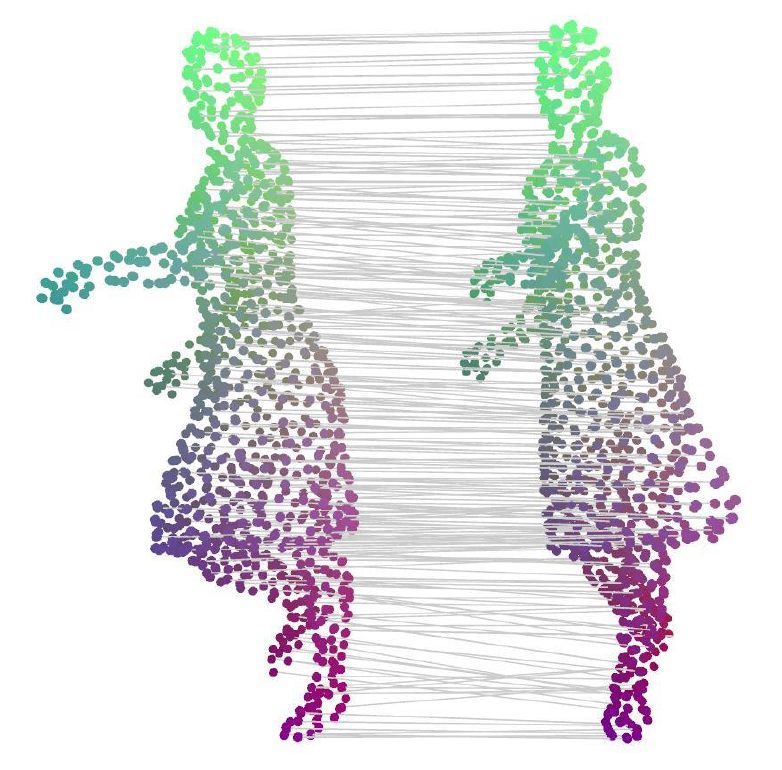}
    }\vspace{-0.4cm}
    \subfigure[]{
	\includegraphics[width=1.9cm]{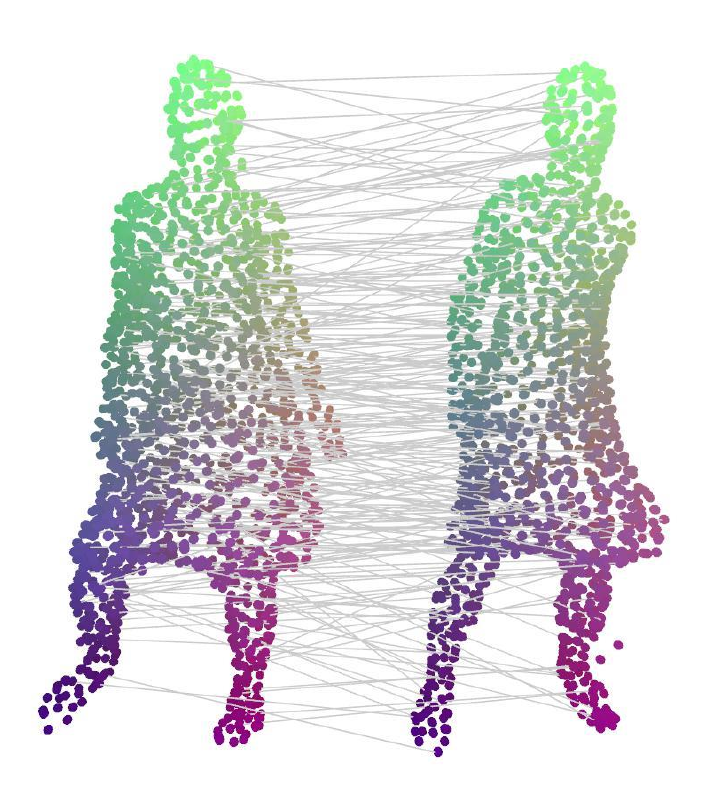}
	\includegraphics[width=2.1cm]{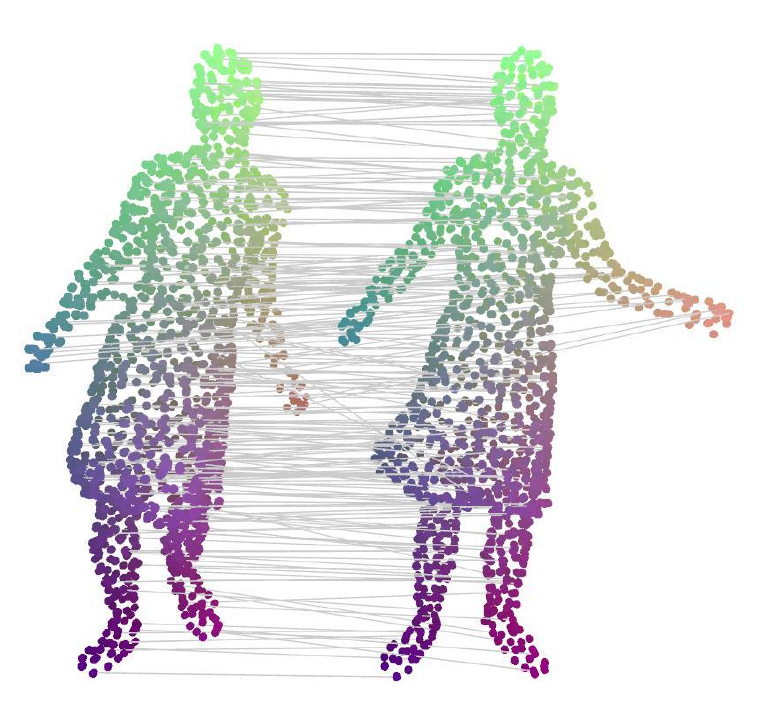}
	\includegraphics[width=1.9cm]{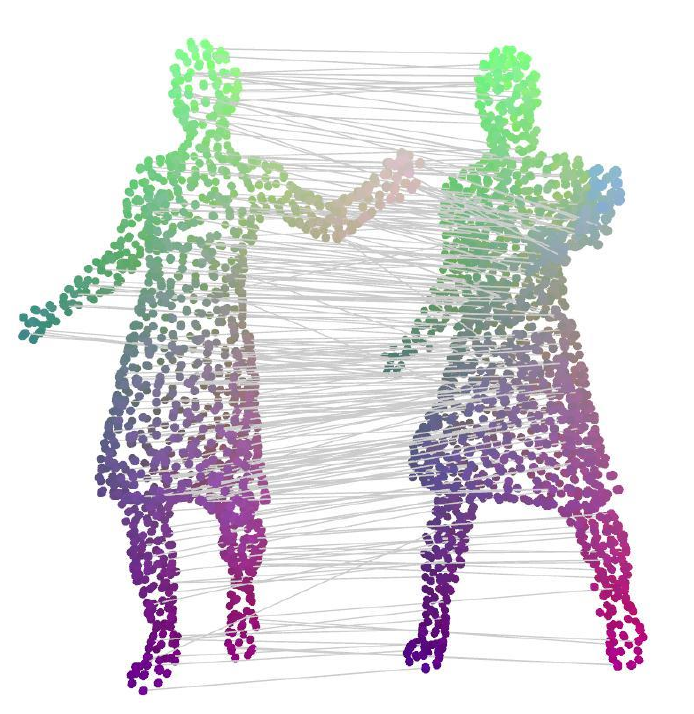}
	\includegraphics[width=1.9cm]{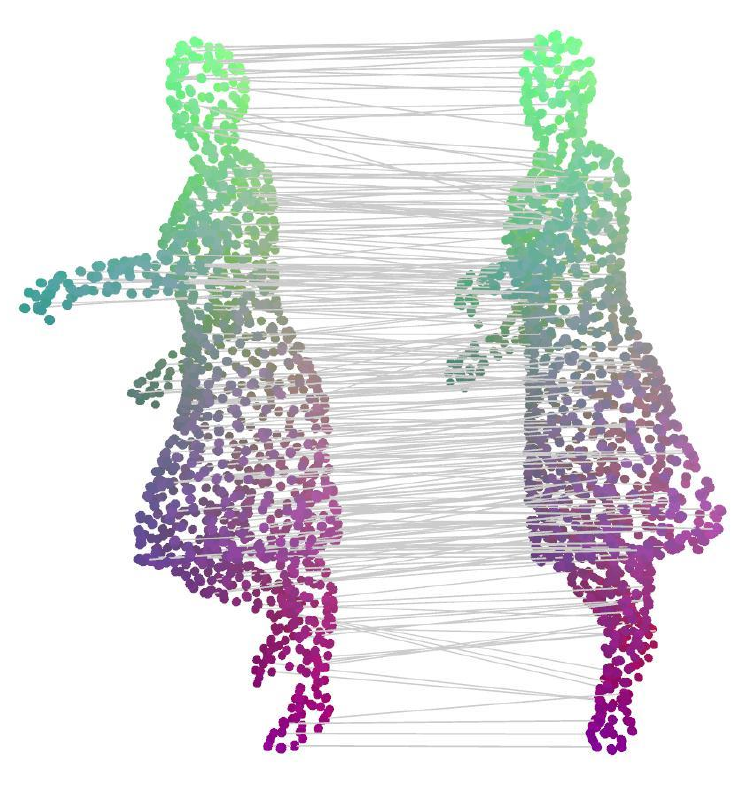}
    }\vspace{-0.4cm}
    \caption{Visual comparisons of temporal consistency (or dense correspondence) across frames of sequence \textit{Swing} learned by (a) our representation framework and (b) CorrNet3D~\cite{zeng2021corrnet3d}, an unsupervised method primarily designed for estimating dense correspondence. Each pair refers to two adjacent point cloud frames of a sequence, and the lines indicate correspondence. Zoom in for details.}
    \label{correspondence_results_representation}
\end{figure}

\begin{figure*}
    \centering
    \subfigure[]{\includegraphics[width=0.22\linewidth]{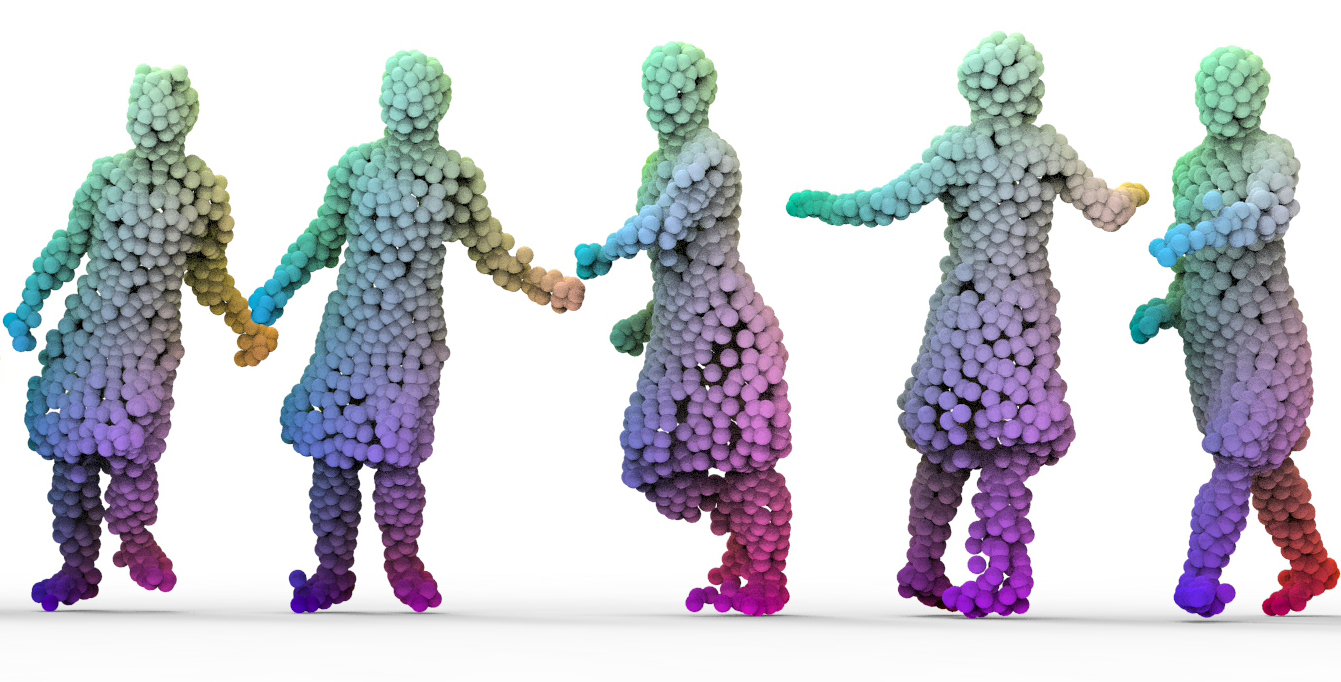}}
    \subfigure[]{\includegraphics[width=0.21\linewidth]{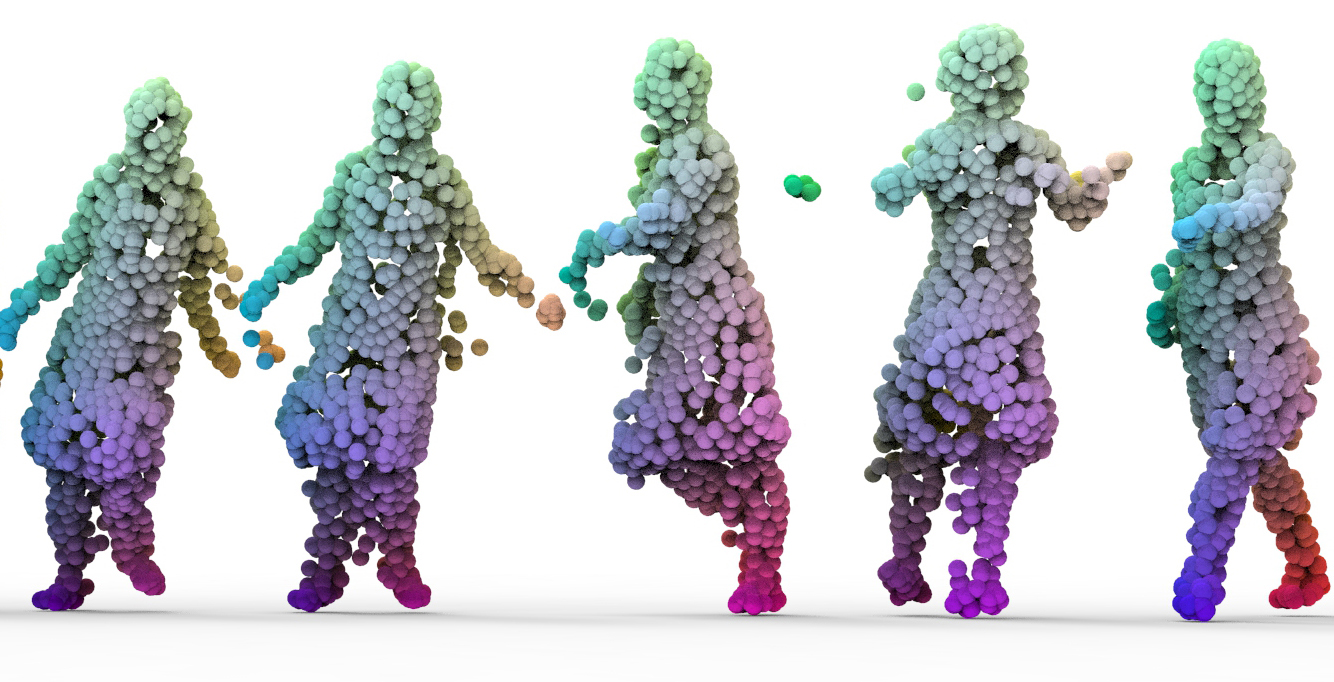}}
    \subfigure[]{\includegraphics[width=0.24\linewidth]{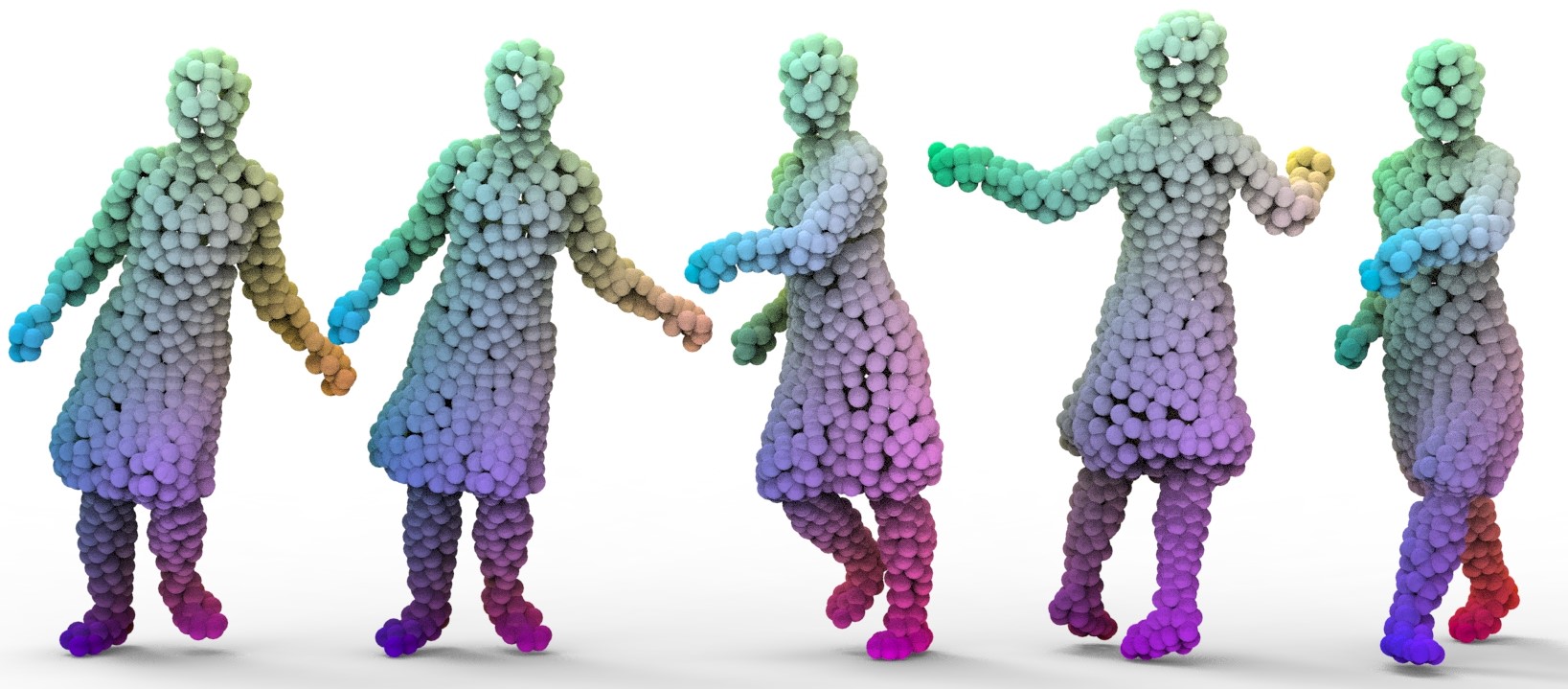}}
    \subfigure[]{\includegraphics[width=0.22\linewidth]{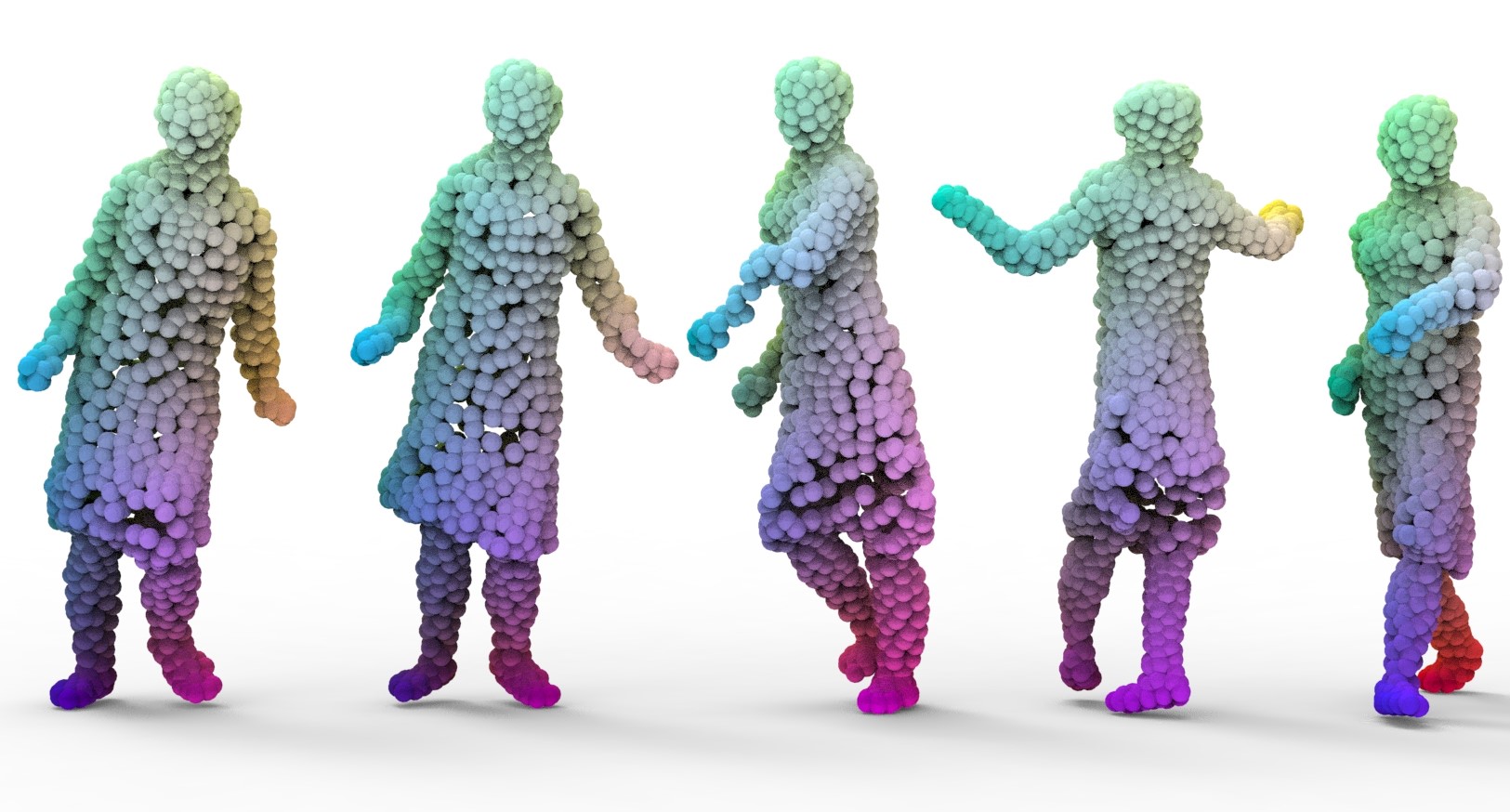}}
    \subfigure[]{\includegraphics[width=0.22\linewidth]{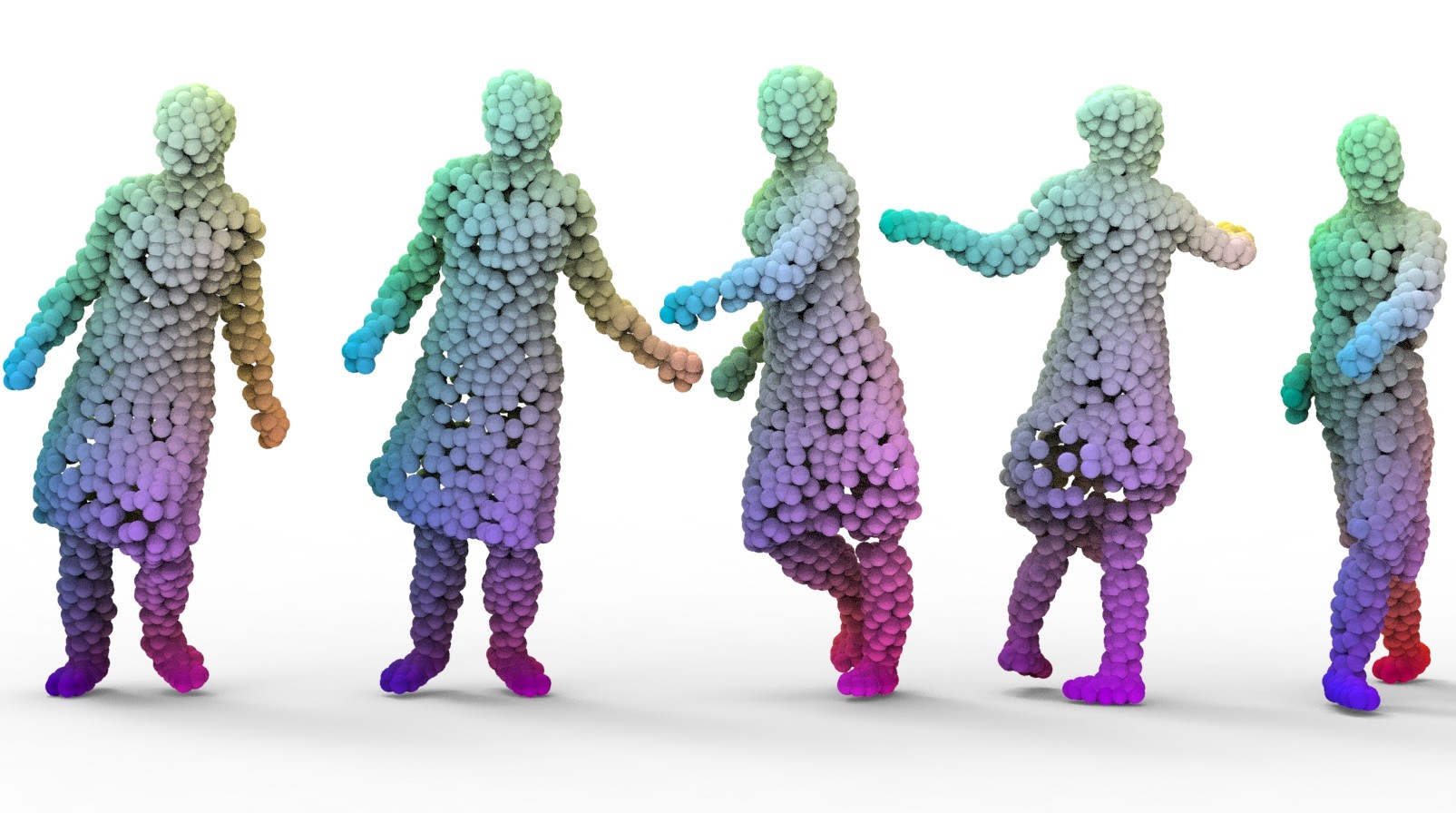}}
    \subfigure[]{\includegraphics[width=0.22\linewidth]{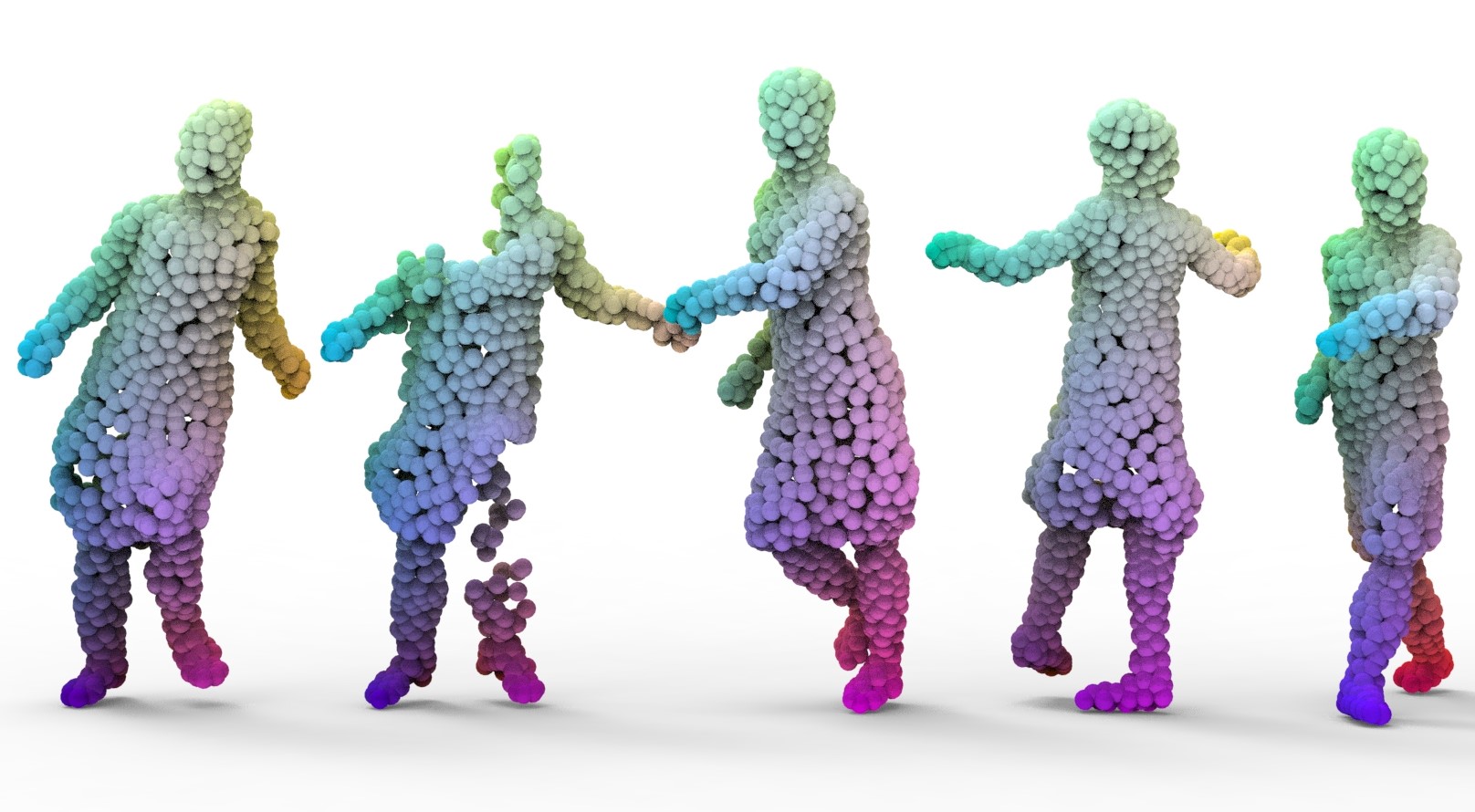}}
    \subfigure[]{\includegraphics[width=0.22\linewidth]{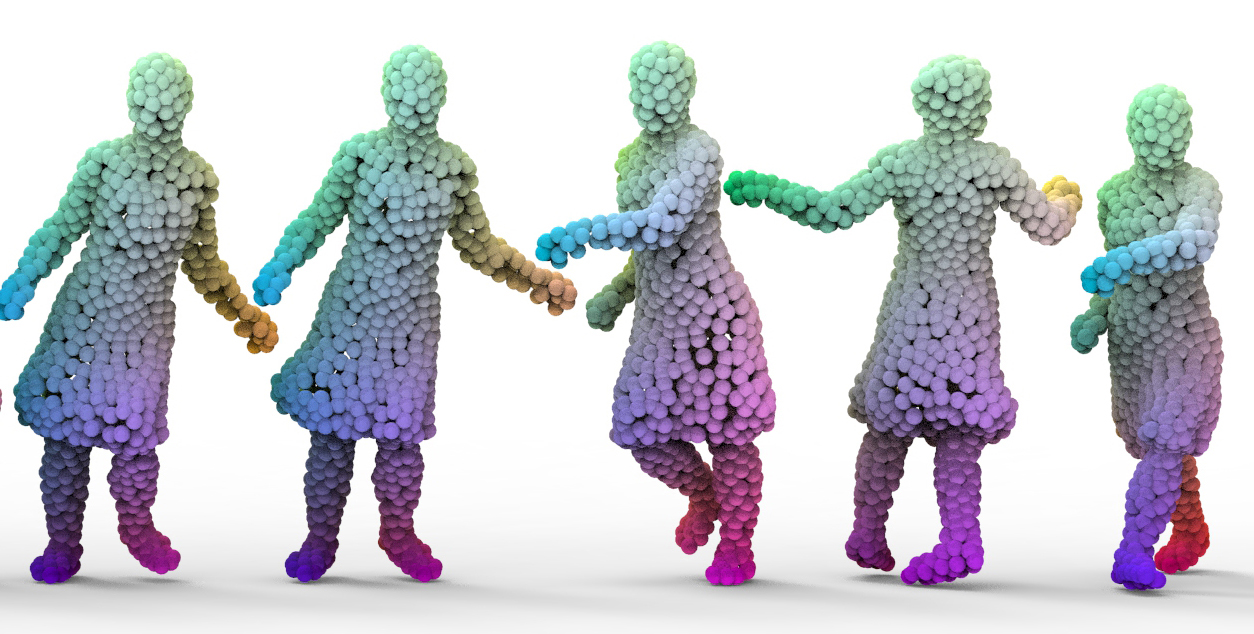}}\vspace{-0.3cm}
    \caption{Visual comparisons of the in-between frames interpolated by (a) IDEA-Net \cite{Zeng_2022_CVPR}, (b) PointINet \cite{lu2020pointinet}, (c) Ours, (d) P4Transformer~\cite{cvpr_21_p4d_fanhehe}, (e) PST-Transformer~\cite{pami_22_psttransformer_fanhehe}, (f) PSTNet2~\cite{pami_21_pstnet2_fanhehe}, and (g) the ground truth on the \textit{Swing} sequence.  We also refer the readers to the \textcolor{magenta}{\textbf{video demo}} included in the \textit{Supplementary Material}
for more visual results.}
    \vspace{-0.3cm}
    \label{intpl_swing} 
\end{figure*}

\begin{table}[t]
    \centering
    \caption{Quantitative results of geometric fidelity of structured point cloud sequences by our method. The values of CD ($\times 10^{-4}$), HD ($\times 10^{-2}$), and mNUC ($\times 10^{-3}$) refer to the average CD, HD, and mNUC of all frames.}
    \label{quant_geo_animals}
    {\fontsize{10}{12}\selectfont
        \renewcommand{\arraystretch}{1.2}
        \begin{tabular}{l|c|c|c}
            \toprule[1.2pt]
            Sequence & CD & HD & mNUC \\
            \midrule
            Camel & 0.024 & 0.66 & 1.3 \\
            Elephant & 0.041 & 0.78 & 1.2 \\
            Horse & 0.032 & 0.68 & 1.1 \\
            Crane & 0.034 & 0.68 & 1.0 \\
            Swing & 0.029 & 0.65 & 1.0 \\
            Bouncing & 0.032 & 0.68 & 1.2 \\
            \bottomrule[1.2pt]
        \end{tabular}
    }
    \vspace{-0.5cm}
\end{table}

\subsubsection{Results of static 3D point clouds} 

We quantitatively evaluated the representation capability of our framework on static 3D point clouds by computing the spatial smoothness and geometric fidelity of resulting 2D image representations. Under this scenario, only the first stage of our framework, i.e., frame-wise structurization, is used. Specifically, we quantify the spatial smoothness by examining the proportion of pixels with a $k\times k$ window on the resulting 2D representation that are the $(k^2-1)$-nearest neighbors of the window's central pixel after reprojecting the pixels into 3D space. The higher the proportion, the better. We utilized CD, Hausdorff Distance (HD), and Mean Normalized Uniformity Coefficients (mNUC) \cite{yu2018pu} between structured and original point clouds to comprehensively assess the geometric fidelity. We employed six point clouds with various topological structures, as shown in the top of Fig. \ref{fig:quality_static}, and each point cloud contains $10^4$ points that are uniformly distributed.  We also bench-marked two recent methods for representing static point clouds as 2D images: Flattening-Net \cite{zhang2023flattening} and RegGeoNet \cite{zhang2022reggeonet}.

From Fig. \ref{reconstruction_smoothness}, it can be observed that our method achieves a smoothness ratio ranging roughly from 35\% to 100\% for window sizes from $3 \times 3$ to $12 \times 12$ among various types of 3D point clouds, which are consistently higher than those of the two compared methods. Such a superiority is also verified by the visual results of the resulting 2D representations shown in Fig. \ref{fig:quality_static}, where our results show smooth pixel distributions while the results of Flattening-Net and RegGeoNet show obvious block effects.

In terms of geometric fidelity, as compared in Table \ref{tab_reconstruction_result_representation}, our method significantly outperforms the competing state-of-the-art, achieving about ten times smaller CD and at least two times smaller HD and mNUC. The advantages reflected in these metrics demonstrate that our representation preserves the original 3D geometry more accurately and captures the point distribution more uniformly. As shown in Fig. \ref{fig:quality_static}, benefiting from our geometry accuracy and distribution uniformity, the 3D surfaces reconstructed from our structured representations are closer to the ground truths, while the results produced by Flattening-Net and RegGeoNet degrade to a large extent.

Overall, when evaluating the representation quality, the metrics of both spatial smoothness and geometry fidelity should be jointly taken into account. Although the spatial smoothness ratios of Flattening-Net and RegGeoNet can be relatively close to ours in some cases, e.g., Figs. \ref{reconstruction_smoothness} (\textcolor{red}{b}), (\textcolor{red}{e}), and (\textcolor{red}{f}), it does not mean their representations achieve sufficiently satisfactory quality because of their non-uniform point distribution, which can also be verified by their inferior surface reconstruction results as shown in Fig. \ref{fig:quality_static}.

\subsubsection{Results of dynamic 3D point cloud sequences}

We quantitatively evaluated the representation ability of our method on dynamic 3D point cloud sequences by computing the mean spatial smoothness and geometric fidelity over all frames of the resulting SPCV representations. Additionally, we defined the correspondence ratio to quantify the temporal consistency of SPCVs, i.e., for each pair of adjacent frames of the SPCV, we calculated the corresponding proportion of the $k$-nearest neighborhood of each point and proposed an algorithm to calculate the correspondence ratio in percentage, with higher values indicating better performance. This metric takes into account both reconstruction and correspondence accuracy. We refer readers to the \textit{Supplementary Material} for detailed descriptions of the calculation process.

Here, we employed six 3D point cloud sequences, including three human motion sequences \cite{MeshAnimationDataLink} and three animal motion sequences \cite{sumner2004deformation}. For the human sequences, we utilized 46 frames from the \textit{swing}, 31 frames from the \textit{bouncing} and \textit{crane}, while for the animal sequences, we employed 10 frames from the \textit{horse}, 48 frames from the \textit{camel} and \textit{elephant}. Each point cloud frame of the sequences contains 10K points.

Our method showcases superior representation quality, as evidenced in Fig. \ref{quant_temporal_consistency_smooth_regular} and Table \ref{quant_geo_animals}. Specifically, as observed in Fig. \ref{quant_temporal_consistency_smooth_regular}, our representation maintains over 60\% temporal consistency and a mean spatial smoothness ratio of over 35\%, indicating its capability to handle various types of sequences. Additionally, Table \ref{quant_geo_animals} reveals that our method consistently achieves robust geometric fidelity across different sequence types, as indicated by stable CD, HD, and mNUC metrics. 
Besides, Fig. \ref{hum_anim_seq_vis_ours} visualizes the SPCV representations on the human and animal sequences, further verifying the effectiveness of the proposed framework in terms of spatial smoothness, temporal consistency, and geometric fidelity. 
As shown in Fig.~\ref{kitti_3of10_crop_frames_1wpts},
we cropped a patch from the first ten frames of KITTI and generated the SPCV for better illustration. We selected 3 frames to visualize a typical situation where a tree appears in the first two frames but vanishes in the last frame. 
The tree is mapped to a similar pixel area, as shown in the red box on the 2D grid. In the last frame, these pixel positions are replaced by the coordinates of the ground.
These results demonstrate the robustness and versatility of the SPCV representation even for scenarios involving scanned LiDAR point cloud sequences.

Moreover, we also compared the learned temporal consistency by our framework with CorrNet3D \cite{zeng2021corrnet3d}, a representative unsupervised method for building dense correspondence between two point clouds, following its evaluation setups. As shown in Fig. \ref{quant_temporal_consistency_diff_methods}, our method outperforms CorrNet3D, achieving over 95\% correspondence ratios for varying neighborhood sizes (i.e., values of $K$). The visual results in Fig. \ref{correspondence_results_representation} demonstrate that our representation maintains more consistent patterns across SPCV frames and ensures more accurate correspondences between adjacent frames.

\subsection{Results of Action Recognition} \label{sec_app_act}

\begin{table}[t]
    \centering
    \caption{Quantitative comparisons of recognition accuracy (\%), GPU memory cost (MB), and forward pass time (seconds) on DeformingThings4D \cite{li20214dcomplete}.}
    \label{tab_action_recognition_result}
    \resizebox{0.9\columnwidth}{!}{
    \begin{tabular}{c|ccc}
        \toprule[1.2pt]
        Methods & Accuracy & Memory & Time\\
        \hline
        P4Transformer \cite{cvpr_21_p4d_fanhehe}  & $80.95$ &  $4020$ & $0.119$\\
        PSTNet2 \cite{pami_21_pstnet2_fanhehe}  & $60.90$ & $10816$ & $0.216$\\
        PSTTransformer \cite{pami_22_psttransformer_fanhehe}  & $76.19$ & $2044$ & $0.010$\\
        \hline
        Ours &  $90.48$ & $2026$ & $0.006$\\
        \bottomrule[1.2pt]
    \end{tabular}}
\end{table}

We compared our SPCV-based action recognition framework with existing state-of-the-art 3D sequence processing methods \cite{cvpr_21_p4d_fanhehe,pami_21_pstnet2_fanhehe,pami_22_psttransformer_fanhehe} on the challenging dataset of deformingThings4D \cite{li20214dcomplete}, which consists of 1,972 animation 3D point cloud sequences across 31 categories of humanoids and animals with large non-rigid deformations. Each frame of the point cloud sequences contains 2,048 points. We segmented the whole sequence into multiple fixed-frame clips, which are fed into all competing methods. To adapt our method, we converted all 3D point cloud sequences to SPCVs with the height and width of each frame equal to 64 and 32, respectively. Following \cite{cvpr_21_p4d_fanhehe}, during inference, the prediction result of the whole sequence is determined by averaging the probabilities of all clips.

As compared in Table \ref{tab_action_recognition_result}, our method outperforms existing state-of-the-art methods with the highest recognition accuracy, which reveals the enhanced feature extraction capability of our SPCV-based learning network. Moreover, our framework turns out to be much more efficient in terms of the GPU memory cost and inference speed, benefiting from the structured nature of SPCVs that can circumvent many expensive operations of indexing, sampling, and grouping.

\subsection{Results of Temporal Interpolation} \label{sec_app_temp}

\begin{table}[t]
    \centering
    \caption{\label{Tab_MITdyn_1} Quantitative comparison of temporal interpolation by different methods on the DHB dataset~\cite{Zeng_2022_CVPR}. The values of CD and EMD is scaled by $\times10^{3}$.}
    \resizebox{0.5\textwidth}{!}{
    \begin{tabular}{c|cc|cc|c|c}
	\toprule[1.2pt] 
	Methods & \multicolumn{2}{c|} {Swing} & \multicolumn{2}{c|} {Longdress} & \multicolumn{1}{c|}{Memory} & \multicolumn{1}{c}{Time} \\
        \cline{2-5}
        &\rule{0pt}{2.6ex}EMD & CD & EMD & CD & (MB) & (seconds) \\
        \midrule 
        \text { PointINet~\cite{lu2020pointinet} } & 15.03 & 1.70 & 10.09 & 0.95 & 6488 & 0.143 \\
        \text { PSTNet2~\cite{pami_21_pstnet2_fanhehe} } & 27.12 & 2.47 & 39.60 & 3.43 & 3009 & 0.142 \\
        \text { P4Transformer~\cite{cvpr_21_p4d_fanhehe} } & 38.35 & 3.24 & 105.68 & 8.88 & 3008 & 0.089 \\
        \text { PST-Transformer~\cite{pami_22_psttransformer_fanhehe} } & 32.43 & 3.02 & 116.26 & 13.83 & 3006 & 0.058 \\
        \text { IDEA-Net~\cite{Zeng_2022_CVPR} } & 7.07 & 1.24 & 5.92 & 0.88 & 11428 & 0.106 \\
        \hline
        \text { Ours } & $2.74$ & $0.86$ & $2.16$ & $0.85$ & $2708$ & $0.014$ \\
        \bottomrule[1.2pt]
    \end{tabular}}
\end{table}

Following the experimental protocols in \cite{Zeng_2022_CVPR}, we performed temporal interpolation on the DHB dataset and made comparisons with existing state-of-the-art 3D point cloud sequence processing approaches~\cite{lu2020pointinet,pami_21_pstnet2_fanhehe, cvpr_21_p4d_fanhehe, Zeng_2022_CVPR, pami_22_psttransformer_fanhehe}, where we adopted CD and EMD as quantitative metrics.

In our implementation, we bench-marked our model against the original learning framework of \cite{Zeng_2022_CVPR}, along with the point-level task models outlined in \cite{pami_21_pstnet2_fanhehe, cvpr_21_p4d_fanhehe, pami_22_psttransformer_fanhehe}. Similarly to \cite{Zeng_2022_CVPR}, we utilized EMD as the training loss function for the compared methods. All the competing methods share the same training and testing data as prepared in \cite{Zeng_2022_CVPR}. As depicted in Fig.~\ref{fig:SPCV_intpl_model}, we constructed an image-based learning framework to fully leverage the capabilities of our regular grid representation for temporal interpolation. Thanks to the structured nature of our SPCV representation, we can directly utilize the pixel-wise Frobenius norm as the loss function.

As shown in Table~\ref{Tab_MITdyn_1}, our approach demonstrates significant improvements over the existing state-of-the-art methods, as indicated by its lower CD and EMD metrics. Besides, our learning framework shows less GPU memory consumption and forward pass time cost. As visualized in Fig.~\ref{intpl_swing}, our method distinctly outperforms the others in preserving geometric fidelity within the interpolated frames. Compared with previous methods, our SPCV representation has explicit temporal consistency, which greatly reduces the difficulty of frame interpolation by the model, thereby achieving a lower geometric error. Thanks to the regular structure of SPCVs, our method can directly use sliding window convolution operations without the need for specially designed grouping and feature aggregation methods, thus improving the processing speed. Moreover, when extracting multi-scale features, due to the regularity of each frame, we do not need to pre-use FPS or other sampling methods to extract key points, which enhances the speed and reduces the GPU storage.

\begin{figure}[t]
    \centering
    \includegraphics[width=0.48\linewidth]{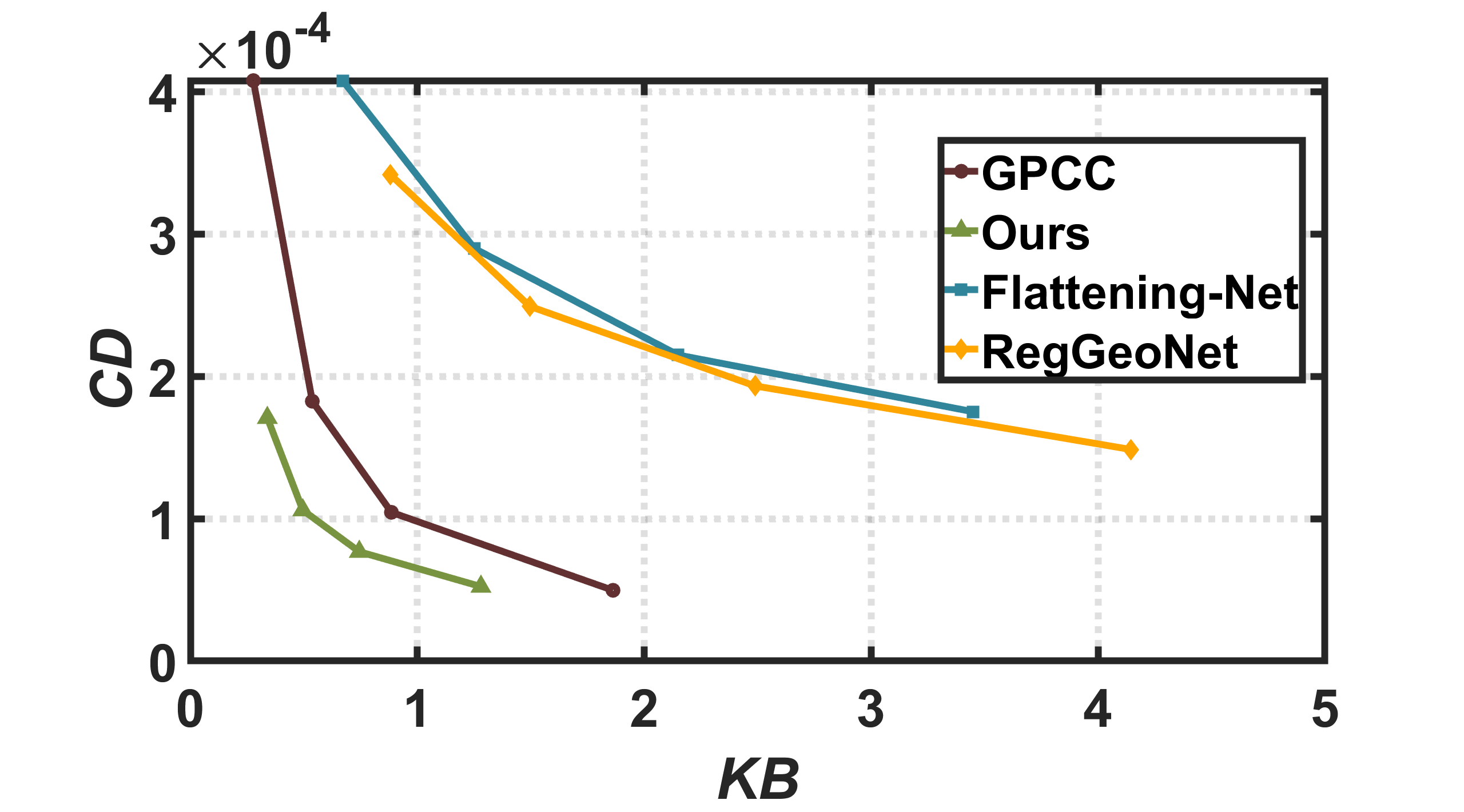}
    \includegraphics[width=0.48\linewidth]{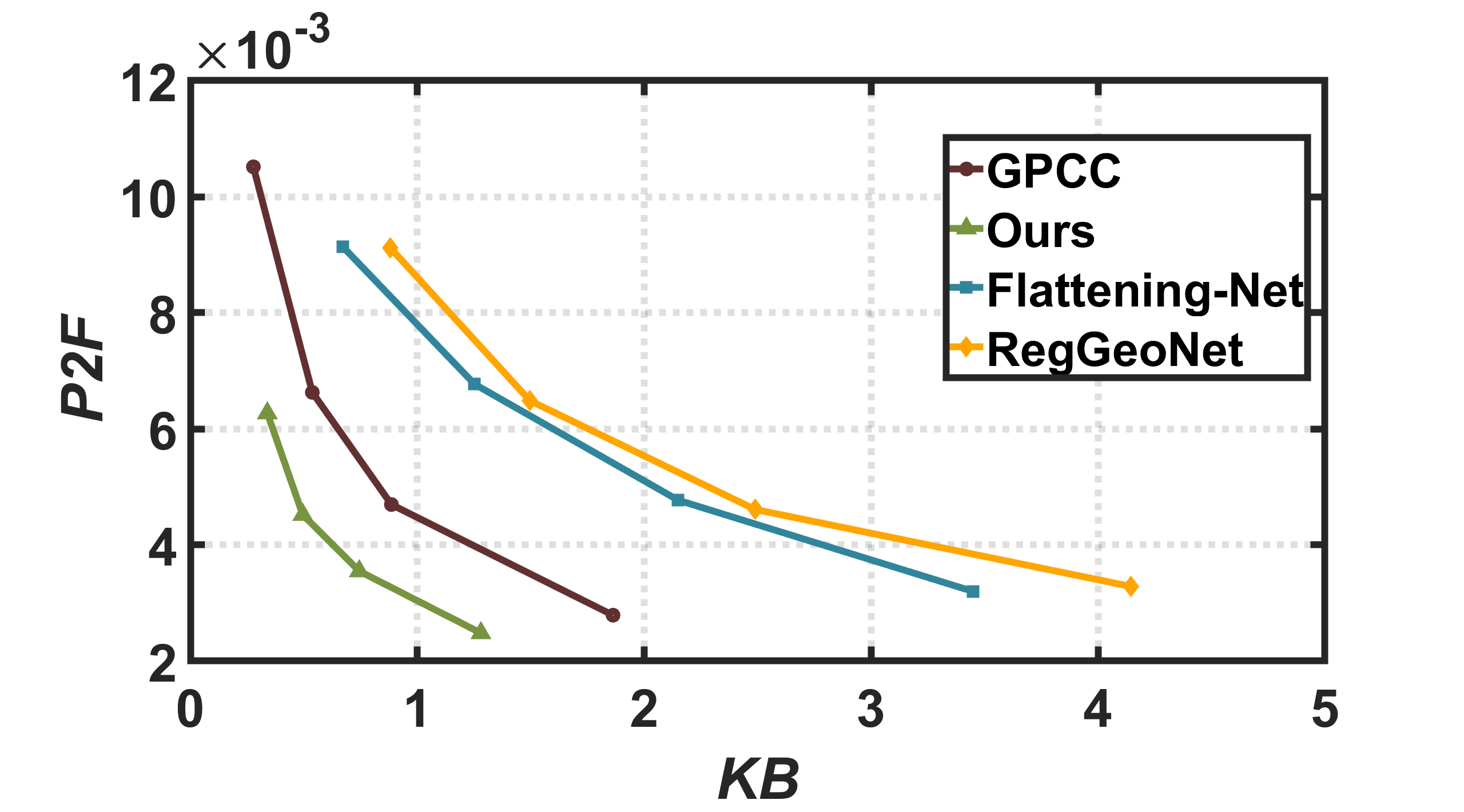}
    \caption{Comparison of different methods for compressing static point clouds. Here, the results refer to a typical frame of the sequence \textit{Exercise} from the MPEG dataset \cite{Owlii2017}.}
    \vspace{-0.3cm}
    \label{compression_results_gpcc} 
\end{figure}

\begin{figure}[t]
    \centering
    \includegraphics[width=0.5\textwidth]
    {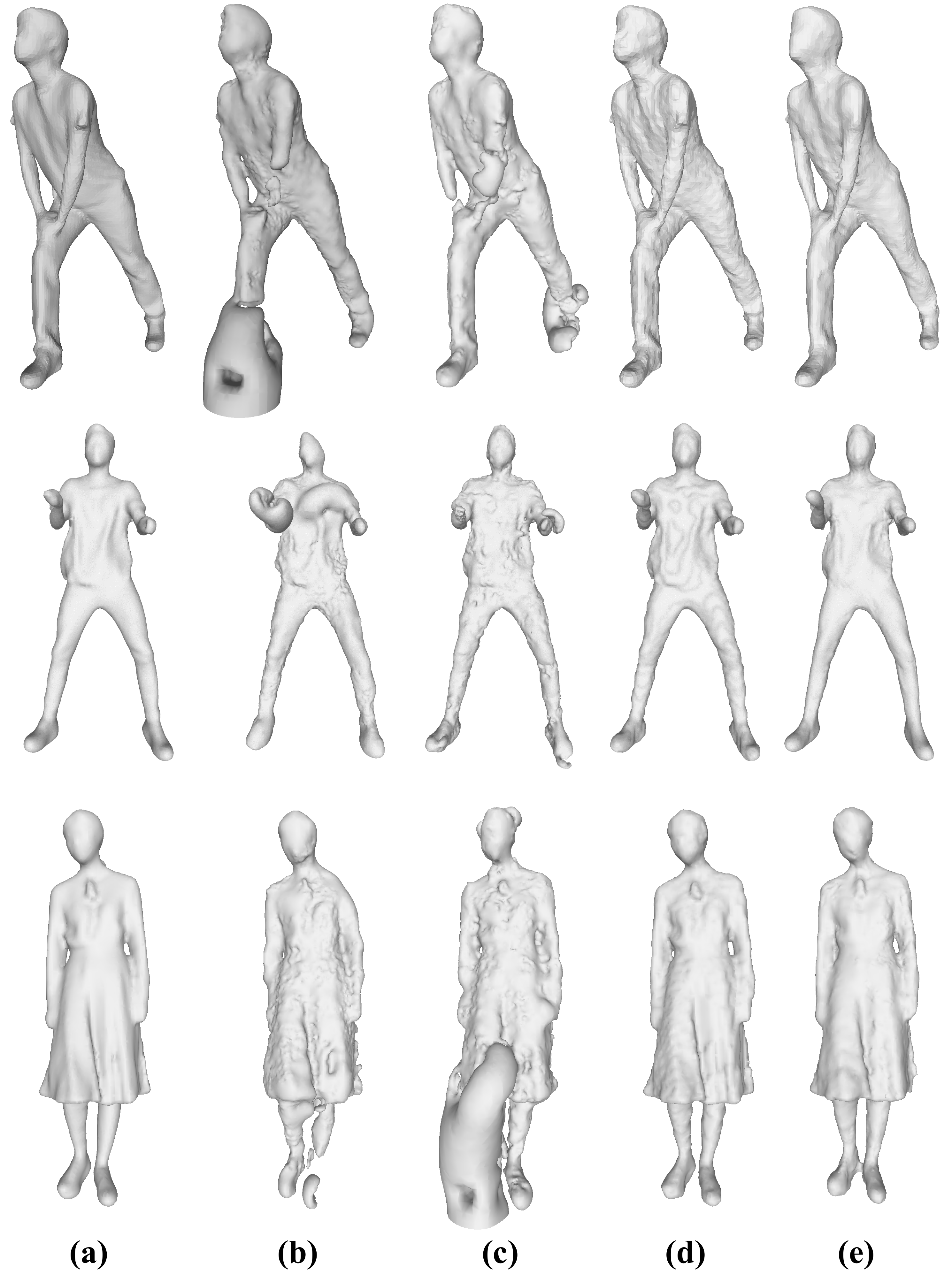}
    \caption{Visual comparisons of reconstructed surfaces from 3D point clouds compressed by various methods with the same compression ratio. (a) Original point clouds, (b) Flattening-Net \cite{zhang2023flattening}, (c) RegGeoNet \cite{zhang2022reggeonet}, (d) G-PCC, and (e) Our approach. The results refer to the shapes from the MPEG dataset \cite{Owlii2017}.}
    \vspace{-0.2cm}
    \label{gpcc_compression_surf_vis}
\end{figure}

\begin{figure}[t]
    \centering
    \includegraphics[width=0.48\linewidth]{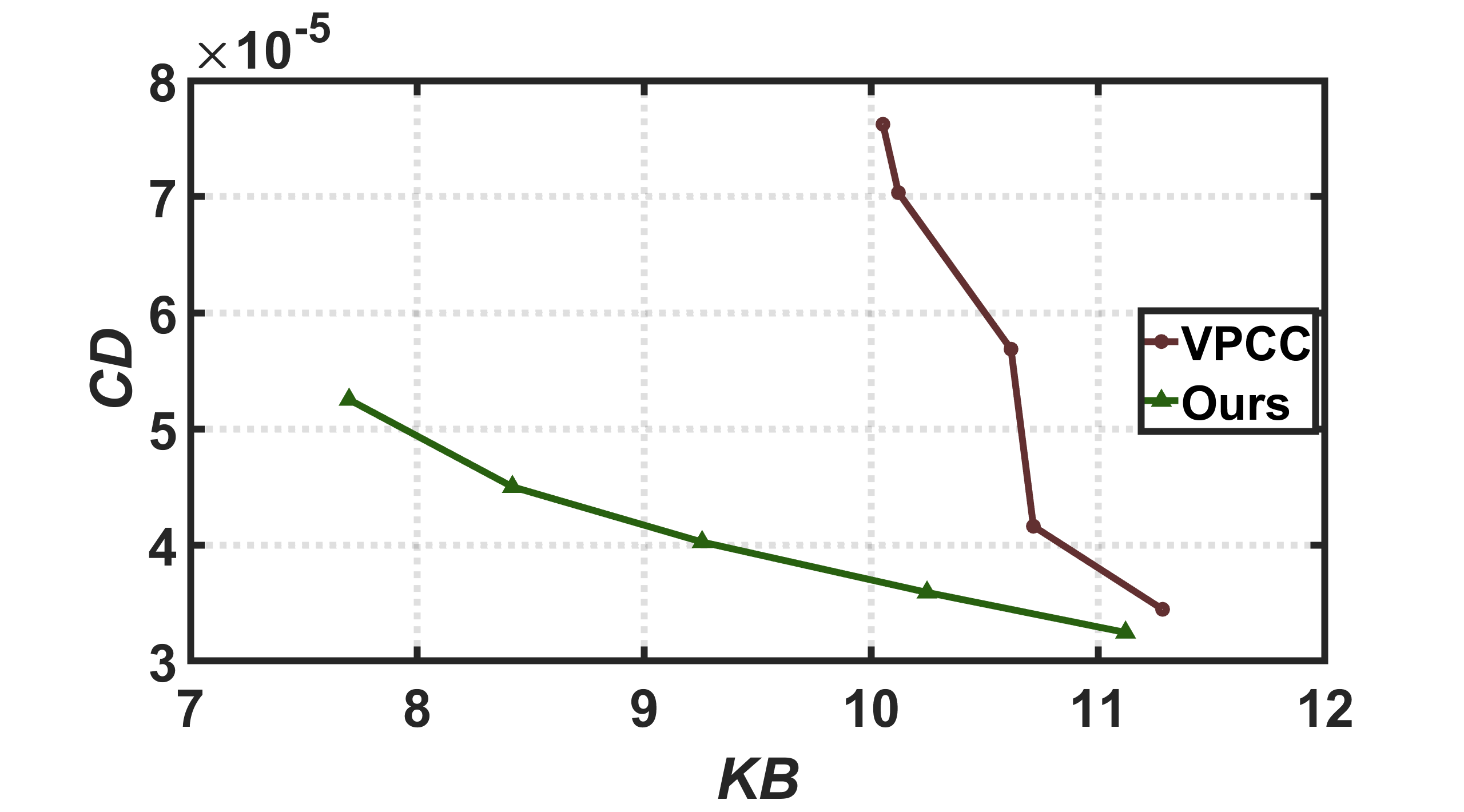} 
    \includegraphics[width=0.48\linewidth]{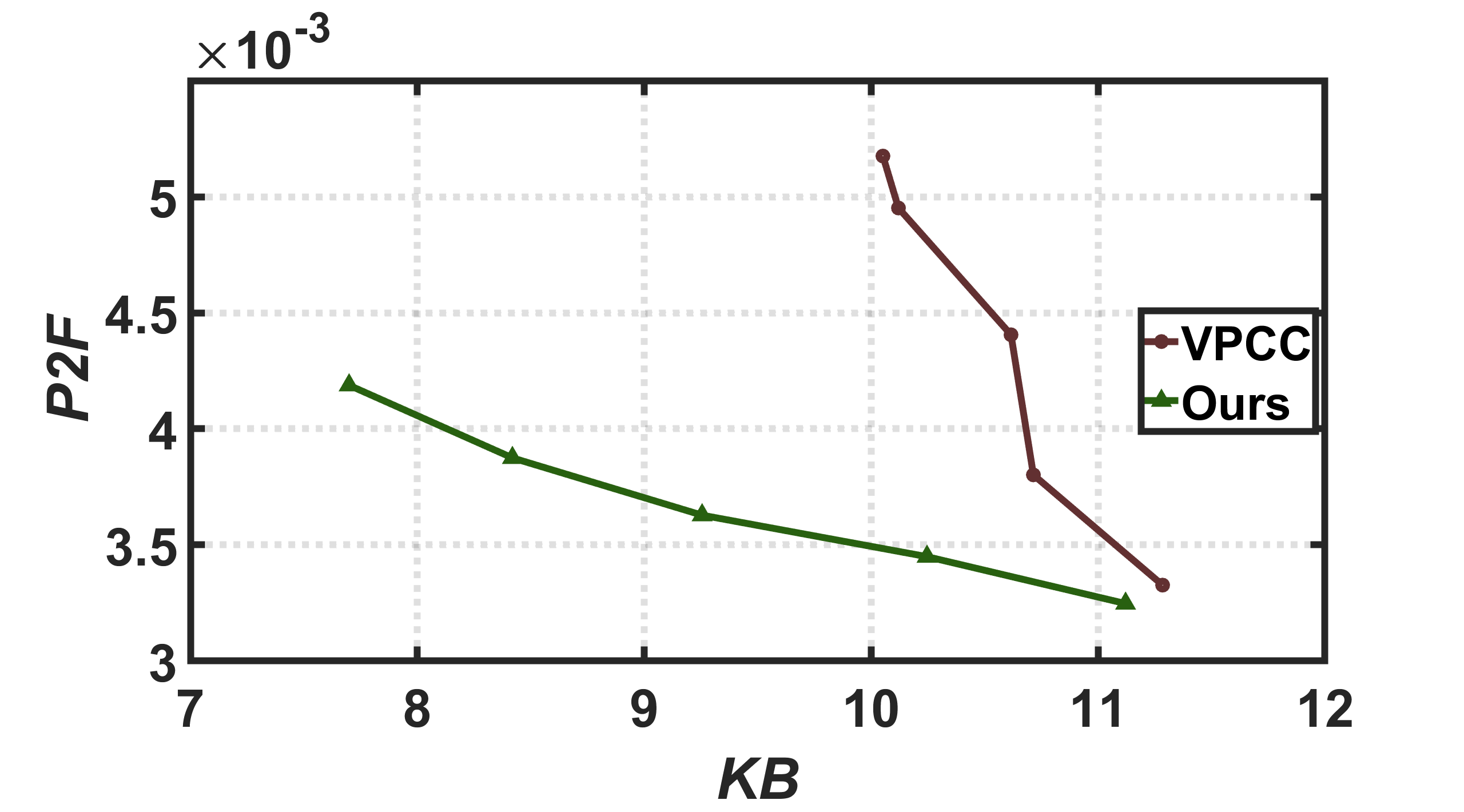}
    \caption{Comparison of our method and V-PCC for compressing point cloud sequences. Here, the results refer to a 5-frame sequence of mixed shapes from the MPEG dataset \cite{Owlii2017}.}
    \vspace{-0.5cm}
    \label{our_vpcc_comparisons} 
\end{figure}

\begin{figure*}[t]
    \centering
    \subfigure[]{
        \includegraphics[width=1.3cm]{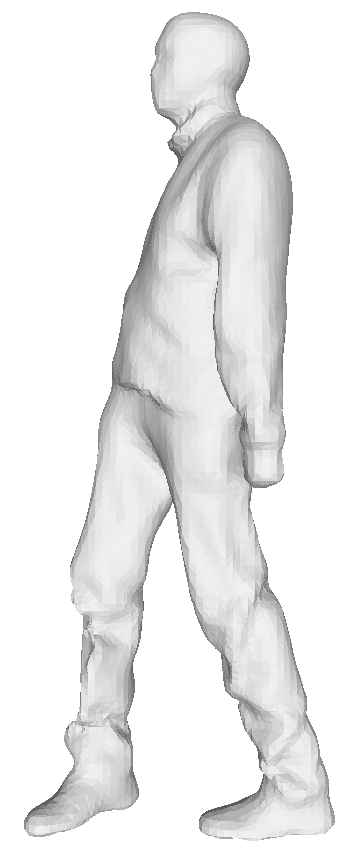}
        \includegraphics[width=2.0cm]{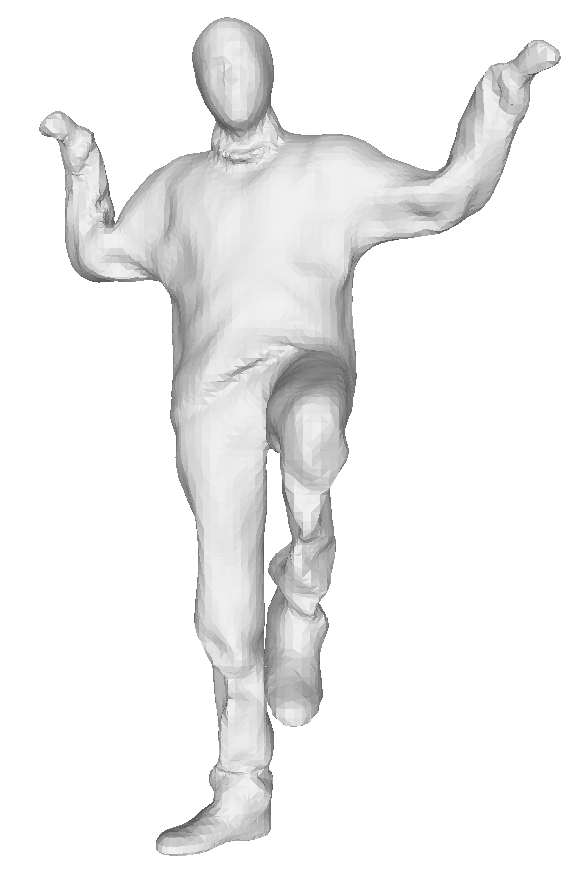}
        \includegraphics[width=2.0cm]{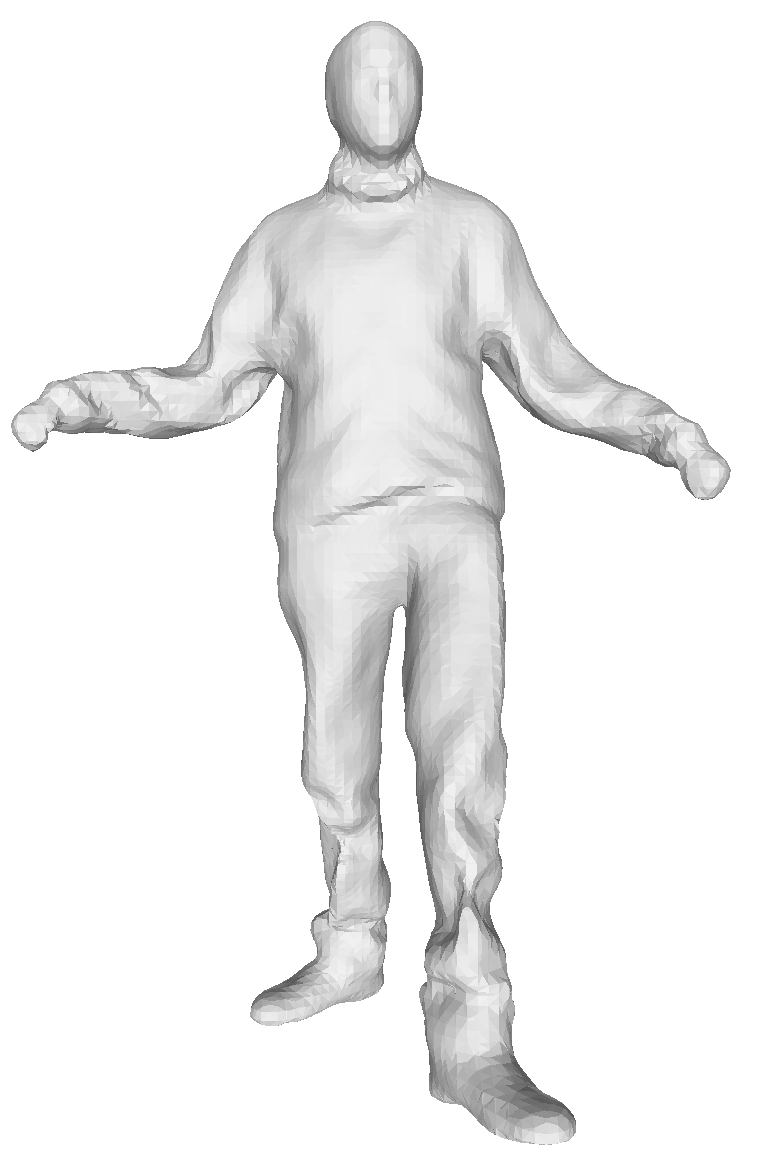}
    }
    \subfigure[]{
        \includegraphics[width=1.3cm]{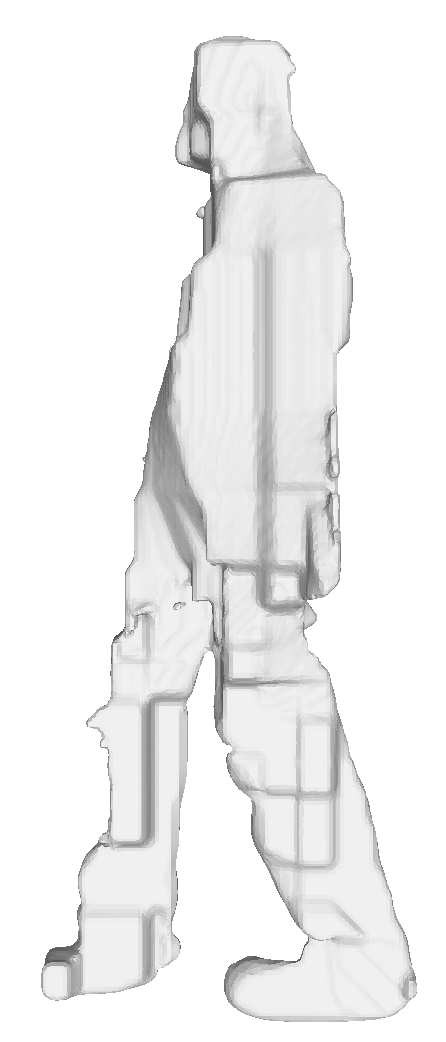}
        \includegraphics[width=2.0cm]{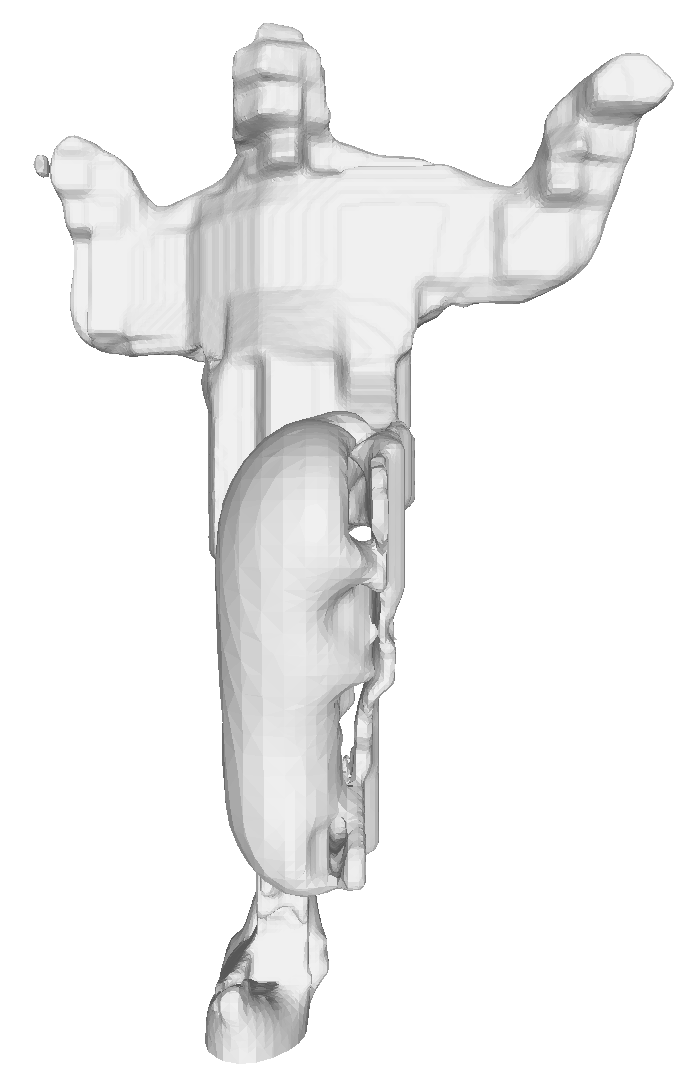}
        \includegraphics[width=2.0cm]{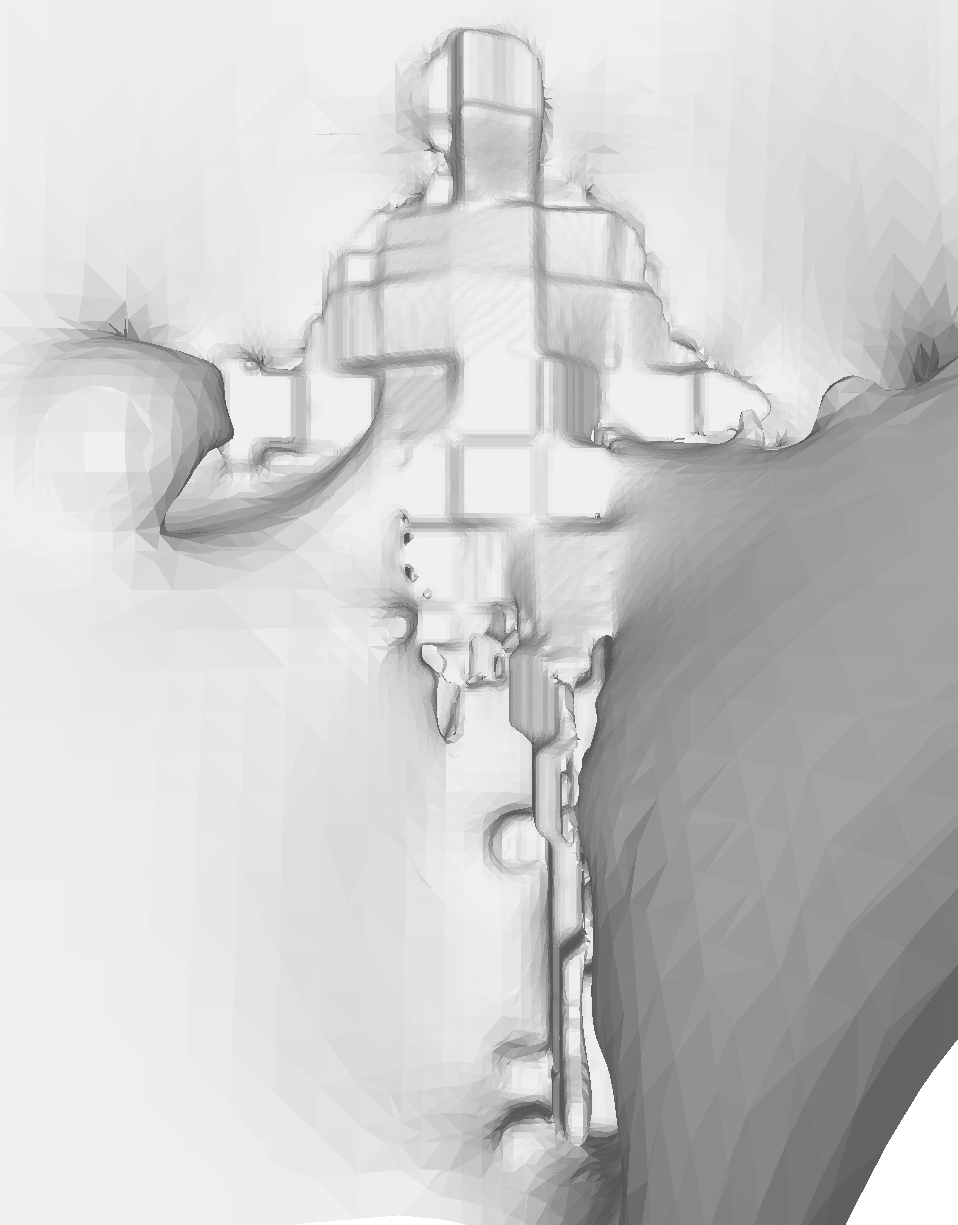}
    }
    \subfigure[]{
        \includegraphics[width=1.3cm]{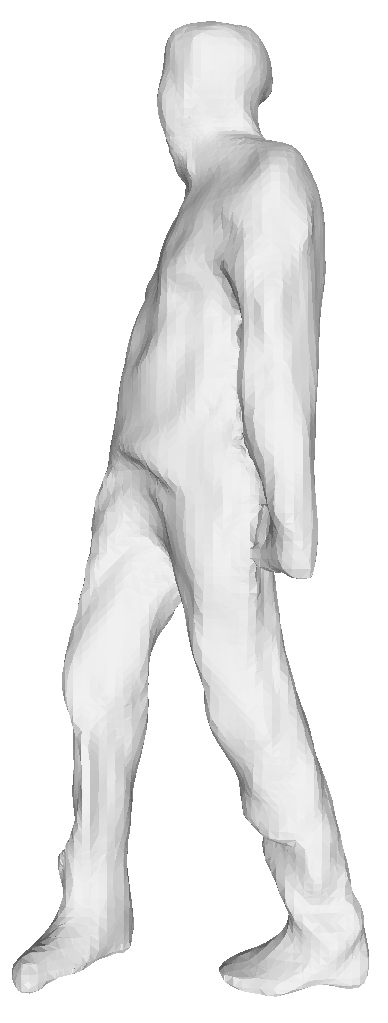}
        \includegraphics[width=2.0cm]{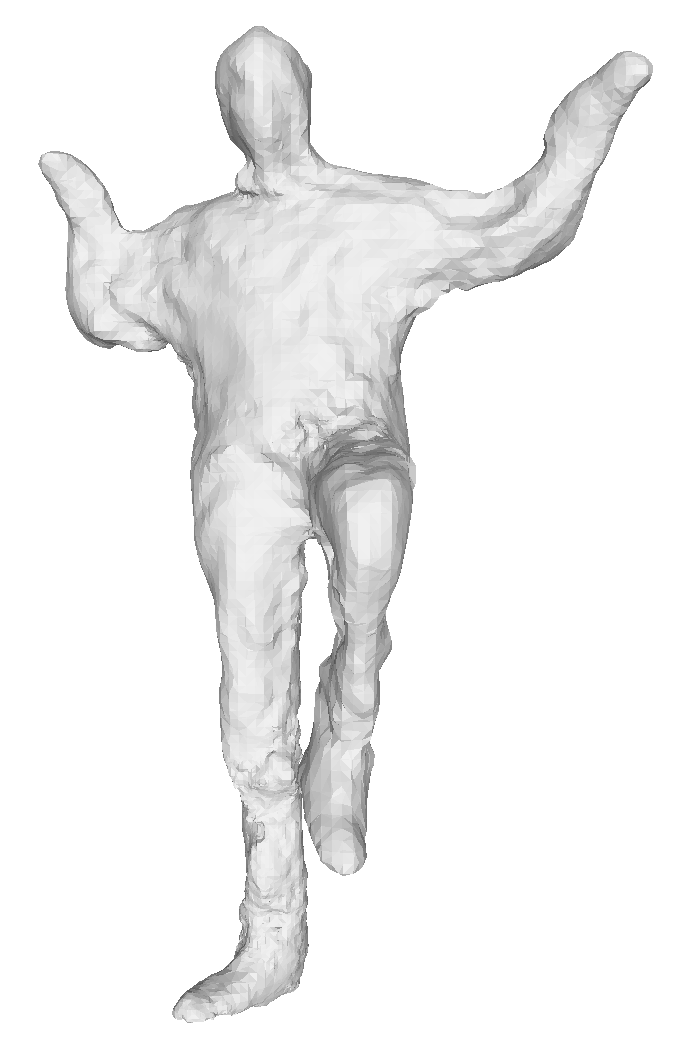}
        \includegraphics[width=2.0cm]{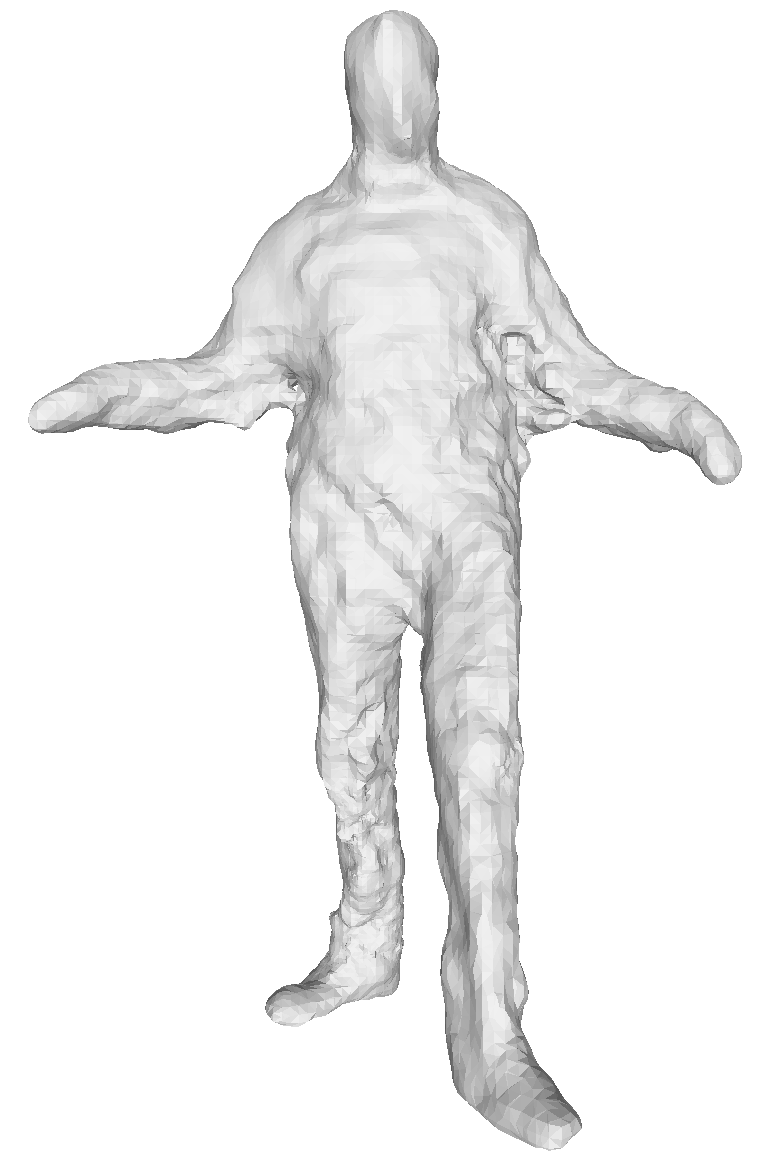}
    }
    \caption{Visual comparisons of surfaces reconstructed from compressed point cloud sequences by various methods at the same compression ratio. (a) Original point cloud sequence (b) V-PCC and (c) Our method. The results refer to the \textit{Crane} sequence.}
    \label{vpcc_vis}
\end{figure*}

\subsection{Results of Compression} \label{sec_app_compre}

We adopted the popular MPEG~\cite{Owlii2017} and MITAMA~\cite{Vlasic2008ArticulatedMA} datasets and bench-marked our SPCV-based point cloud compression framework for compressing both static and dynamic point clouds. For experiments on static point cloud compression where each point cloud uniformly contains 10K points, we made comparisons with the latest version of G-PCC (TMC13 v23.0-rc2). Besides, we also included Flattening-Net \cite{zhang2023flattening} and RegGeoNet \cite{zhang2022reggeonet} for comparison by feeding their generated 2D representations into the H.266/VVC video encoder. For experiments on dynamic point cloud compression where each sequence consists of 5 frames and each frame contains 250K points, we made comparisons with the latest version of V-PCC (TMC2 Release 18.0).

The quantitative results depicted in Fig. \ref{compression_results_gpcc} indicate that our SPCV-based compression framework outperforms both G-PCC and recent point cloud structurization methods. This is substantiated by achieving lower CD and Point-to-Face (P2F) metrics at the same compression ratio. The enhanced performance can be attributed to the structured nature of the SPCV representation that is conducive to the utilization of the advanced video compression techniques, and the intrinsic spatial smoothness of SPCV that further boosts the compression performance. Visual comparisons in Fig. \ref{gpcc_compression_surf_vis} show that our compression method more accurately preserves the geometric fidelity of the surfaces reconstructed from the decoded 3D point clouds.

For point cloud sequence compression, the results in Fig. \ref{our_vpcc_comparisons} demonstrate that our SPCV-based compression framework significantly outperforms V-PCC in compression efficiency, as evidenced by much lower CD and P2F metrics at the same compression ratio (or much smaller compressed file size at the same CD and P2F metrics). The SPCV's temporal consistency enhances its intra-frame compression capabilities, optimizing the overall performance. Visual comparisons in Fig. \ref{vpcc_vis} demonstrate that our SPCV more faithfully preserves the geometric fidelity in the reconstructed surfaces.

\subsection{Ablation Studies}\label{sec_abl_stu}

We conducted ablation studies to facilitate analyzing and understanding the proposed learning framework for structuring an arbitrary 3D point cloud sequence into a 2D video.

\begin{figure}[t]
    \centering
    \raisebox{-\height}{\includegraphics[width=0.45\linewidth]{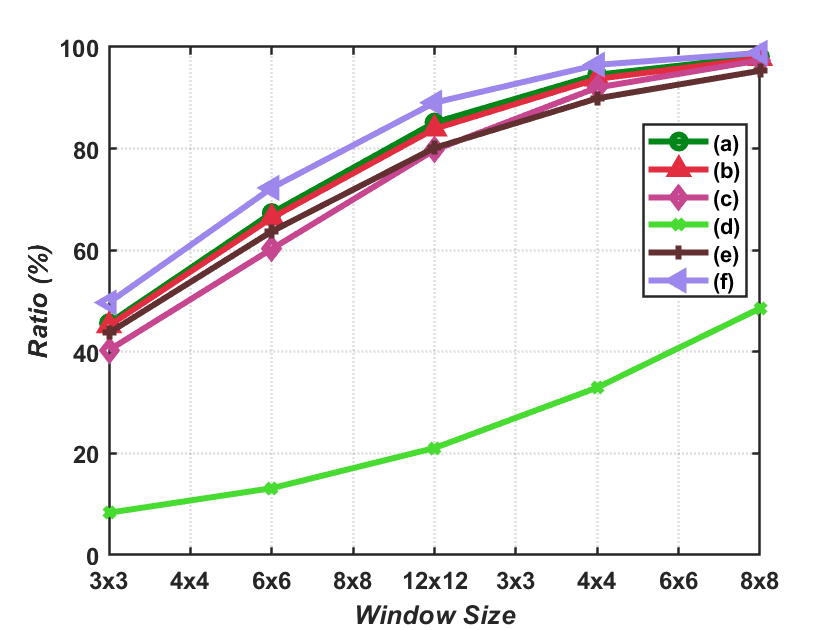}}
    \raisebox{-\height}{\includegraphics[width=0.45\linewidth]{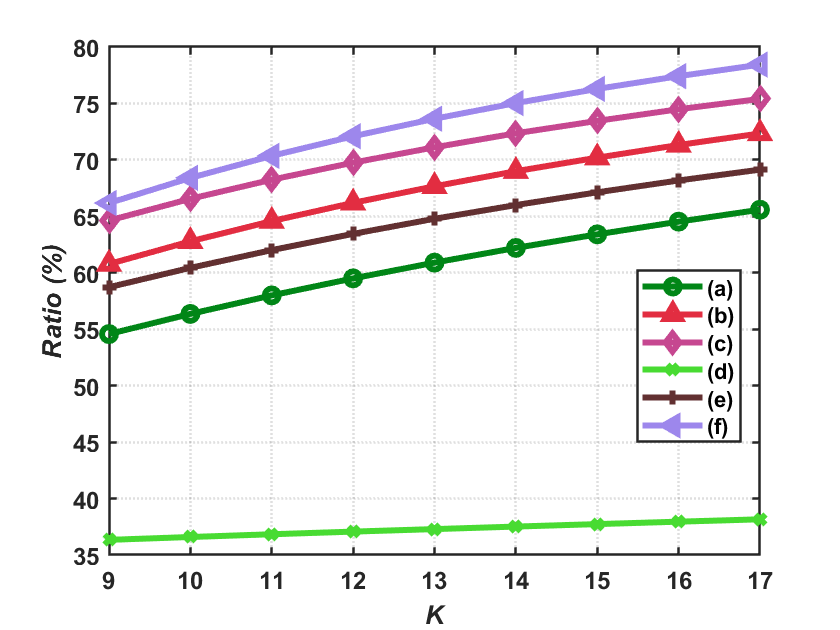}}
    \raisebox{-1.2\height}{
        \scalebox{0.9}{
            \begin{tabular}{c|c|c|c}
                \toprule[1.2pt]
                Variants & CD $\times 10^{-4}$ & HD $\times 10^{-2}$ & mNUC $\times 10^{-3}$ \\
                \hline
                (a) & 0.043 & 0.70 & 1.1 \\
                (b) & 0.037 & 0.70 & 1.2 \\
                (c) & 0.029 & 0.65 & 1.0 \\
                (d) & 0.059 & 0.74 & 1.3 \\
                (e) & 0.258 & 1.76 & 2.3 \\
                (f) & 0.029 & 0.65 & 1.0 \\
                \bottomrule[1.2pt]
            \end{tabular}
        }
    }
    \caption{Ablative analyses of spatial smoothness (top left), temporal consistency (top right), and geometric fidelity (bottom) for variants of our method. (a) without sequential structuring, (b) without normal constraints, (c) without spatial constraints, (d) with EMD, (e) with CD, and (f) full model.}
    \label{ablations}
    \vspace{-0.3cm}
\end{figure}

\subsubsection{Effectiveness of sequence-wise structurization}

In the framework depicted in Fig. \ref{fig:SPCV_twostages}, we introduced a sequence-wise stage representing the remaining point cloud frames. To validate the importance of this stage in maintaining temporal consistency, we removed the sequence-wise processing stage and instead processed each frame independently through the frame-wise stage to obtain the SPCV. As evident from Fig. \ref{ablations} (a), this setting ensures a considerable level of intra-frame smoothness but fails to maintain temporal consistency across the entire frame sequence.

\subsubsection{Effectiveness of geometric regularization in Eqs. (\ref{equ:frame_reg}) and (\ref{equ:seq_reg})} 

We explored the effects of the geometric regularization terms used in the SPCV framework. As shown in Figs. \ref{ablations} (\textcolor{red}{b}) and (\textcolor{red}{c}), the overall performance degrades when removing each individual constraint. Specifically, the removal of the spatial regularization, i.e., $\lambda_{\rm s} = 0$, impacts the model's ability to maintain spatial smoothness within individual frames. Besides, the absence of normal regularization, i.e., $\lambda_{\rm n} = 0$, leads to a decline in both the spatial smoothness and temporal consistency, highlighting its critical role in preserving the geometric integrity of the model. Collectively, these findings underscore the importance of geometric constraints in the overall efficacy and quality of the SPCV framework.

\subsubsection{Impact of distance metrics}

We evaluated the impact of alternative distance metrics (EMD and CD) on our framework's performance in comparison to standard settings. As shown in Figs. \ref{ablations} (\textcolor{red}{d}) and (\textcolor{red}{e}), both EMD and CD do not match the efficacy of our original configuration. EMD  prioritizes global shape alignment over maintaining point proximity during point cloud reconstruction; however, searching for such a global alignment cannot guarantee spatial smoothness and temporal consistency, even with the corresponding constraints. On the other hand, although CD  is more effective than EMD in preserving smoothness, the reconstructed shapes under its alignment are extremely inaccurate than EMD and our selected distance metric.

\section{Conclusion} \label{S_Con}

This paper presents a novel generic representation for structuring dynamic 3D point cloud sequences as 2D videos. We constructed a self-supervised learning framework together with geometrically meaningful constraints for achieving spatial smoothness and temporal consistency. The generated SPCVs show satisfactory representation quality. To demonstrate the practical value, we conducted a variety of downstream tasks, including action recognition, temporal interpolation, and compression, where our SPCV-based learning frameworks outperform existing state-of-the-art approaches. Through comprehensive experimental evaluations, we demonstrated the universality and potential of the proposed SPCV representations. We believe that this study opens up many new possibilities for research on dynamic point cloud processing and learning.

\bibliographystyle{IEEEtran}
\bibliography{Ref}

\vspace{-1cm}
\begin{IEEEbiography}[{\includegraphics[width=1in, height=1.25in, clip, keepaspectratio]{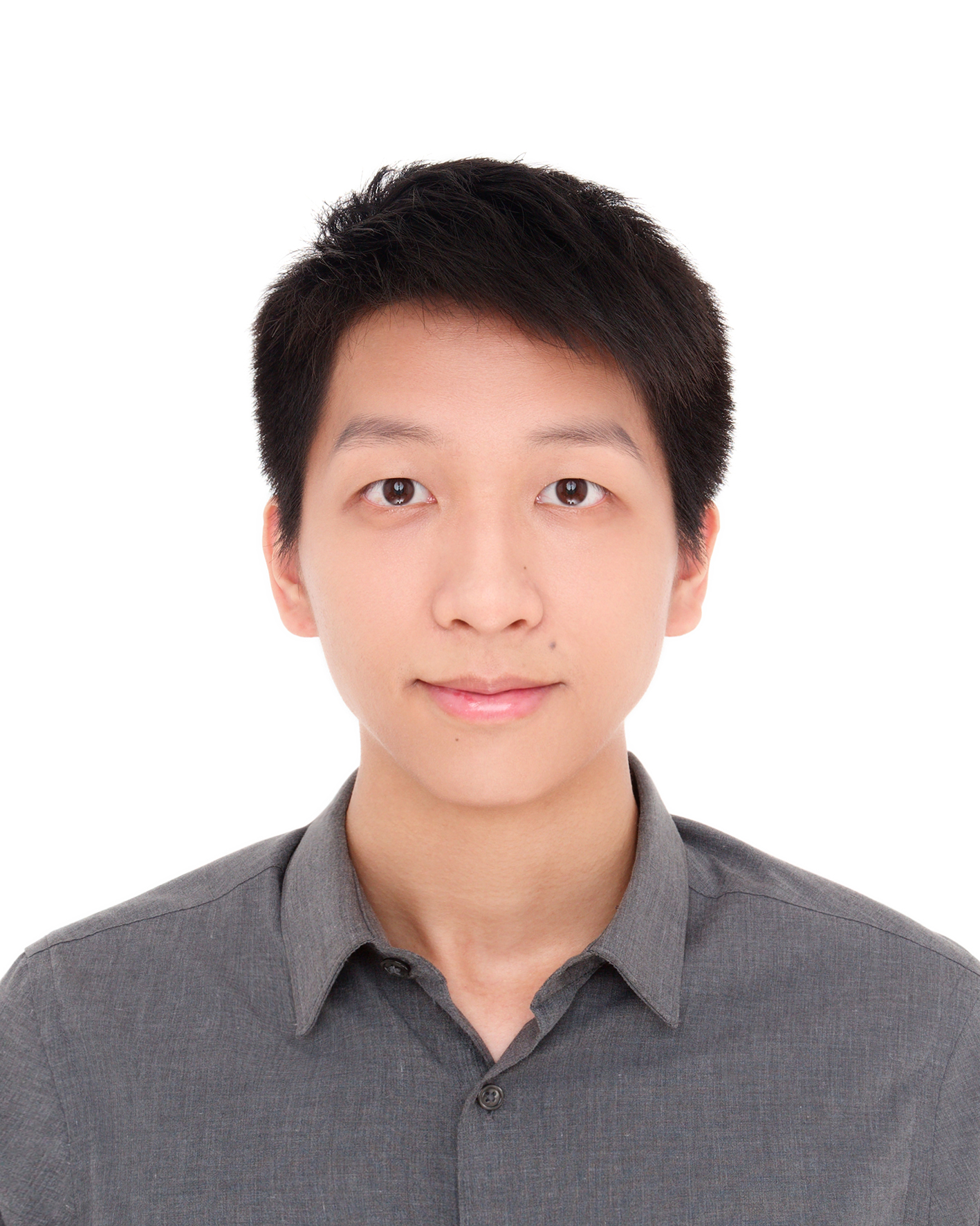}}]{Yiming Zeng} received his B.S. degree in Automation from the South China University of Technology, Guangzhou, China, in 2019, and his Ph.D. degree in Computer Science from the City University of Hong Kong in 2024. His research interests include deep learning processing and the development of geometrically meaningful representations for both static and dynamic 3D point clouds.
\end{IEEEbiography}

\vspace{-0.8cm}
\begin{IEEEbiography}[{\includegraphics[width=1in, height=1.25in, clip, keepaspectratio]{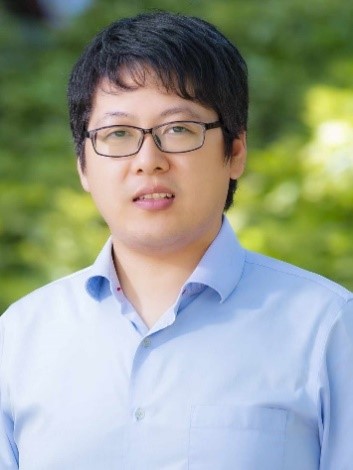}}]{Junhui Hou} (Senior Member, IEEE)  is an Associate Professor with the Department of Computer Science, City University of Hong Kong. He holds a B.Eng. degree in information engineering (Talented Students Program) from the South China University of Technology, Guangzhou, China (2009), an M.Eng. degree in signal and information processing from Northwestern Polytechnical University, Xi’an, China (2012), and a Ph.D. degree from the School of Electrical and Electronic Engineering, Nanyang Technological University, Singapore (2016). His research interests are multi-dimensional visual computing.

Dr. Hou received the Early Career Award (3/381) from the Hong Kong Research Grants Council in 2018. He is an elected member of IEEE MSATC, VSPC-TC, and MMSP-TC. He has served or is serving as an Associate Editor for IEEE Transactions on Visualization and Computer Graphics, IEEE Transactions on Image Processing, IEEE Transactions on Circuits and Systems for Video Technology, and Signal Processing: Image Communication, and The Visual Computer.
\end{IEEEbiography}

\vspace{-1cm}
\begin{IEEEbiography}[{\includegraphics[width=1in, height=1.25in, clip, keepaspectratio]{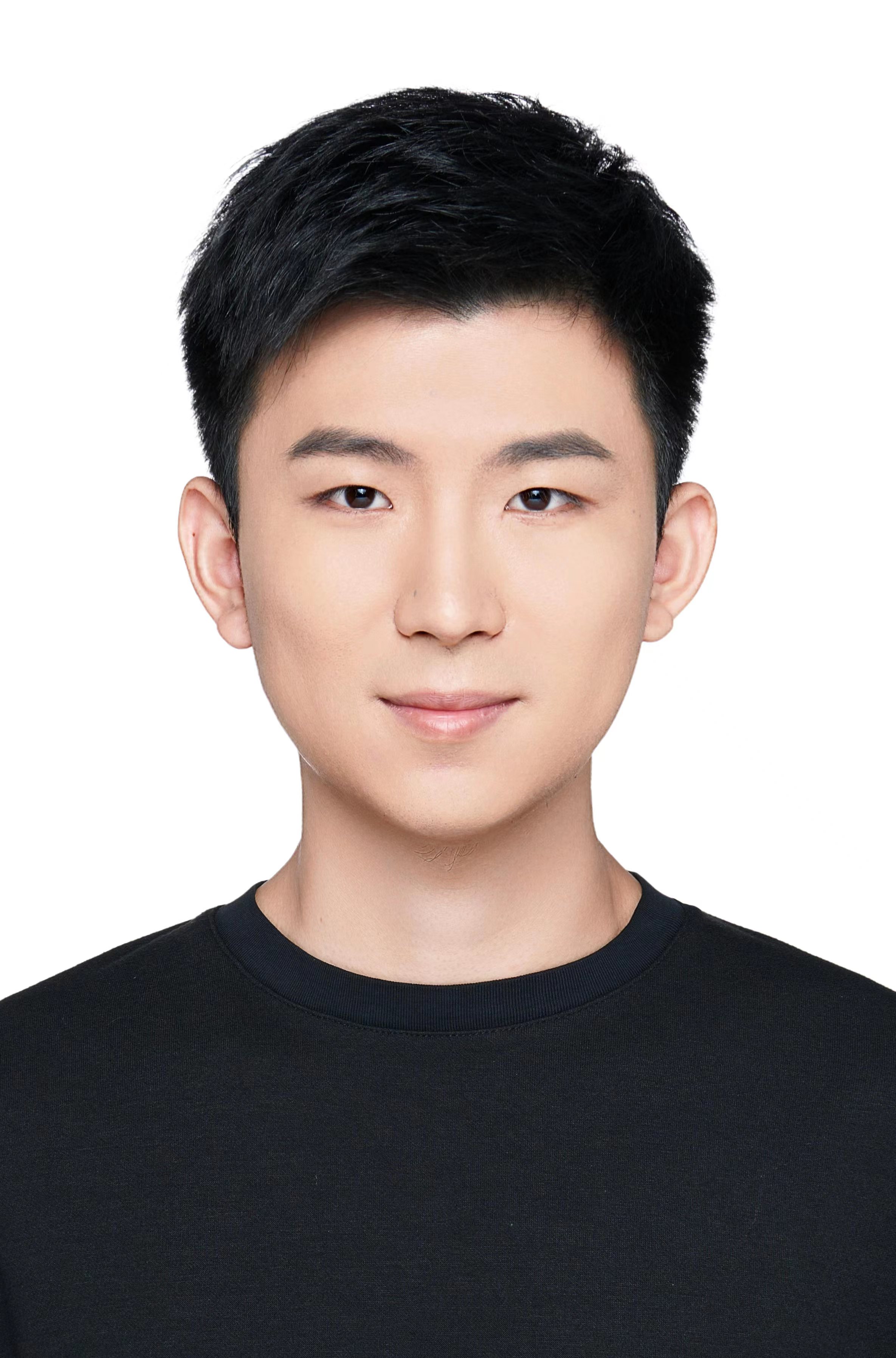}}]{Qijian Zhang}
received the B.S. degree in Electronic Information Science and Technology from Beijing Normal University, Beijing, China, in 2019. Currently, he is a Ph.D. student (2020-present) under the Department of Computer Science, City University of Hong Kong, HKSAR. His research interests include geometry processing, 3D computer vision, computer graphics, and cross-modal learning.
\end{IEEEbiography}

\vspace{-1cm}
\begin{IEEEbiography}[{\includegraphics[width=1in, height=1.25in, clip, keepaspectratio]{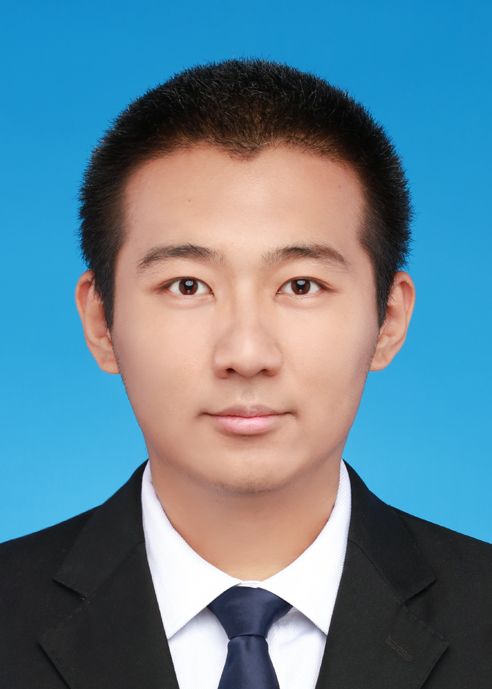}}]{Siyu Ren} received the B.S. degree in Optoelectronic Information Science and Engineering from Tianjin University, Tianjin, China, in 2018. He is currently pursuing the Ph.D. degree in Computer Science at the City University of Hong Kong and Optical Engineering at the Tianjin University. His research interests include deep learning and 3D point cloud processing.
\end{IEEEbiography}

\vspace{-1cm}
\begin{IEEEbiography}[{\includegraphics[width=1in, height=1.25in, clip, keepaspectratio]{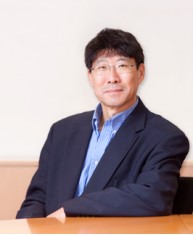}}]
{Wenping Wang} (Fellow, IEEE) received the Ph.D. degree in computer science from the University of Alberta in 1992. He is a Professor of computer science at Texas A\&M University. His research interests include computer graphics, computer visualization, computer vision, robotics, medical image processing, and geometric computing, and he has published over 180 journal papers in these fields. He is a journal associate editor of Computer Aided Geometric Design (CAGD), Computer Graphics Forum (CGF), and IEEE Transactions on Visualization and Computer Graphics. He has chaired a number of international conferences, including Pacific Graphics 2012, ACM Symposium on Physical and Solid Modeling (SPM) 2013, SIGGRAPH Asia 2013, and Geometry Summit 2019. Prof. Wang received the John Gregory Memorial Award for his contributions to geometric modeling. He is an ACM Fellow.
\end{IEEEbiography}

\end{document}